%% file: root.tex
\documentclass[journal]{IEEEtran}
\usepackage{amsmath,amsfonts}
\usepackage{algorithmic}
\usepackage{algorithm}
\usepackage{array}
\usepackage[font=small]{caption}
\usepackage{subfig}
\usepackage{multirow}

\usepackage{textcomp}
\usepackage{stfloats}
\usepackage{url}
\usepackage{verbatim}
\usepackage{graphicx}
\usepackage{cite}
\usepackage{siunitx}
\usepackage{makecell}
\usepackage{import}
\usepackage{hyperref}
\usepackage[usenames,svgnames,table,rgb,dvipsnames]{xcolor} %
\usepackage{transparent}
\usepackage{bm}
\DeclareUnicodeCharacter{2212}{\ensuremath{-}}

\usepackage{xcolor}         %

\renewcommand{\sec}[1]{Sec.~\ref{#1}}
\newcommand{\fig}[1]{Figure~\ref{#1}}
\newcommand{\figs}[1]{Fig.~\ref{#1}}

\newcommand{\tab}[1]{Table~\ref{#1}}%

\newcommand{\website}[0]{\url{https://sites.google.com/view/evetac}}

\let\ACMmaketitle=\maketitle
\renewcommand{\maketitle}{\begingroup\let\footnote=\thanks \ACMmaketitle\endgroup}

\makeatletter
\newcommand*\titleheader[1]{\begingroup\gdef\@titleheader{#1}\let\footnote=\thanks\endgroup}
\AtBeginDocument{%
  \let\st@red@title\@title
  \def\@title{%
  \begin{flushleft}
    \vspace{-2.15em}
    \bgroup\normalfont\small\@titleheader\par\egroup
    \vspace{-25pt}\par\noindent\rule{\textwidth}{0.1pt}
    \end{flushleft}
    \vskip0.5em\st@red@title
        }

}

\usepackage{fancyhdr}
\newcommand{\mytitle}{
\copyright 2024 IEEE. Personal use of this material is permitted.
Permission from IEEE must be obtained for all other uses, in any current or future media, including reprinting/republishing this material for advertising or promotional purposes, creating new collective works, for resale or redistribution to servers or lists, or reuse of any copyrighted component of this work in other works.} 
\input{misc/math_commands}

\makeatother

\title{Evetac: An Event-based Optical Tactile Sensor\\ for Robotic Manipulation}

\author{Niklas Funk$^{1}$, Erik Helmut$^{1}$, Georgia Chalvatzaki$^{1,2}$, Roberto Calandra$^{3,4}$, Jan Peters$^{1,2,5,6}$%
\thanks{Corresponding author: Niklas Funk.~$^{1}$Computer Science Department, Technical University of Darmstadt, Darmstadt, Germany {\tt\small \{niklas,georgia,jan\}@robot-learning.de}}%
\thanks{$^{2}$Hessian Centre for Artificial Intelligence}%
\thanks{$^{3}$LASR Lab, Technische Universit\"at Dresden, Germany
{\tt\small roberto.calandra@tu-dresden.de}}%
\thanks{$^{4}$The Centre for Tactile Internet with Human-in-the-Loop (CeTI), Germany}%
\thanks{$^{5}$German Research Center for AI (DFKI), Research Department: SAIROL}
\thanks{$^{6}$Centre for Cognitive Science, TU Darmstadt}
}

\titleheader{{\textbf{Accepted version.} To appear in \textit{IEEE Transactions on Robotics, Vol. 40, 2024.}.  DOI: 10.1109/TRO.2024.3428430}}

\begin{document}

\maketitle

\thispagestyle{fancy}

\begin{abstract}
Optical tactile sensors have recently become popular.
They provide high spatial resolution, but struggle to offer fine temporal resolutions.
To overcome this shortcoming, we study the idea of replacing the RGB camera with an event-based camera and introduce a new event-based optical tactile sensor called Evetac.
Along with hardware design, we develop touch processing algorithms to process its measurements online at 1000 Hz.
We devise an efficient algorithm to track the elastomer’s deformation through the imprinted markers despite the sensor’s sparse output.
Benchmarking experiments demonstrate Evetac’s capabilities of sensing vibrations up to 498 Hz, reconstructing shear forces, and significantly reducing data rates compared to RGB optical tactile sensors.
Moreover, Evetac’s output and the marker tracking provide meaningful features for learning data-driven slip detection and prediction models.
The learned models form the basis for a robust and adaptive closed-loop grasp controller capable of handling a wide range of objects.
We believe that fast and efficient event-based tactile sensors like Evetac will be essential for bringing human-like manipulation capabilities to robotics.
The sensor design and additional material is open-sourced at \website{}.

\end{abstract}


\section{Introduction}

\IEEEPARstart{M}{anipulation} of mechanical objects is essential for real-world robotic applications ranging from industrial assembly~\cite{nottensteiner2021towards} to household robots~\cite{ciocarlie2014towards, lach2022placing}.
Physical manipulation of objects includes making and breaking contact between the robotic manipulator and the object of interest and the application of sufficient contact forces.
For the goal of having reactive, adaptive, reliable, and efficient dexterous manipulation skills also in unstructured environments with little or no prior knowledge available, direct sensing of contacts, i.e., tactile sensing, is of crucial importance~\cite{li2020review}.

Due to their huge potential, there exists a long history in developing tactile sensors for robotics~\cite{dahiya2009tactile, kappassov2015tactile, chen2018tactile, li2020review}. %
While a wide range of sensing technologies have been proposed \cite{schmitz2010tactile, taunyazov2020event, buscher2015flexible, bhirangi2021reskin, fishel2012design}, recently, especially RGB optical tactile sensors have received increased attention~\cite{yamaguchi2019recent}.
They are also known as vision-based tactile sensors, as their functioning principle relies on an RGB camera capturing an elastomer’s deformation.
RGB optical tactile sensors are appealing due to their small form factors, compatibility with standard interfaces, low-cost, and high spatial resolution.
Yet, compared to other tactile sensing technologies and the human sense of touch \cite{johansson2009coding, dahiya2009tactile}, they typically lack temporal resolution.
\begin{figure}
    \centering
    \includegraphics[width=\columnwidth]{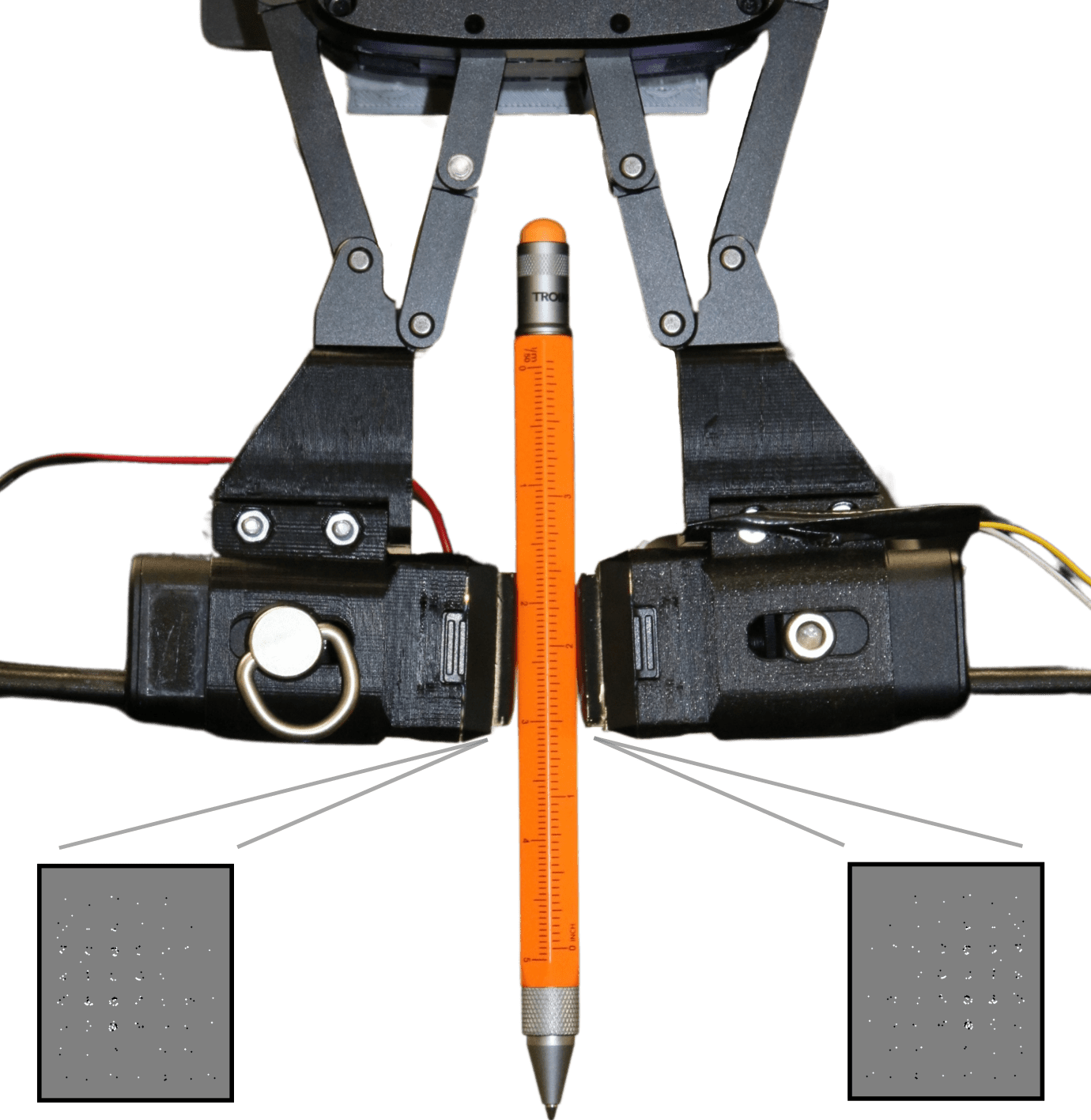}%
    \caption{Two Evetac sensors installed in the ROBOTIS \textrm{RH-P12-RN(A)} gripper holding a pen. In the bottom left and right, we depict a magnified version of the sensors' measurements.
    Evetac is an open-source event-based optical tactile sensor for robotic manipulation. Its main components are an illuminated, dotted, soft silicone gel that interacts with the environment. Changes in gel configuration are captured by an event-based camera inside the sensor as shown in the bottom left \& right.}
    \label{fig:evetac_parallel_grip}
\end{figure}
To overcome this shortcoming, in this paper, we propose a novel event-based optical tactile sensor called Evetac.
While event-based optical tactile sensors have been presented previously \cite{rigi2018novel, kumagai2019event, ward2020neurotac}, it still remains an underexplored area.

The proposed Evetac sensor is largely inspired by popular RGB optical tactile sensors such as GelSight~\cite{yuan2017gelsight}, TacTip~\cite{ward2018tactip}, and DIGIT~\cite{lambeta2020digit}.
The sensor's functioning principle thus also relies upon a camera capturing the deformation of a soft silicone gel.
The gel has imprinted markers which provide natural features for the interaction between sensor and object. 
What makes Evetac substantially different from common RGB optical tactile sensors is that we replace the RGB camera with an event-based camera. %
Event-based cameras recently gained lots of attraction due to their properties of high temporal resolution, high pixel bandwidth, and low energy consumption \cite{gallego2020event}.
This camera replacement fundamentally changes the sensor's properties as it allows for achieving a significantly increased temporal resolution, which can be beneficial for reliably detecting fast contact-related phenomena such as vibrations.
In our case, we obtain touch measurements at \SI{1000}{\hertz} while at the same time pertaining a spatial resolution of 640x480 pixels.
Moreover, the sparse signal returned from event-based cameras, which only return signal upon intensity changes at the pixel locations, also enables real-time signal processing despite the high readout rate. 
This opens the door to closing tactile control loops at high frequencies.

Evetac's design attempts to maximize the re-use of existing commercially available components to reduce the entry barrier into the field of event-based optical tactile sensing. %
Only Evetac's custom housing has to be 3D-printed. It is open-sourced, with the list of the other components and assembly instructions on \href{https://sites.google.com/view/evetac}{our website}.
In addition to the hardware design, we develop the necessary software interfaces for reading out the sensor in real-time, at \SI{1000}{\hertz}.
We also present a novel, gradient-based algorithm for real-time tracking of the dots imprinted in the gel.
The algorithm allows to keep track of the gel's global state despite the sensor's sparse outputs.
The dots' displacement can be used to reconstruct the shear forces acting on the sensor.
Evetac's raw sensor output and the information from the dot tracking provide the main features for developing slip detection and prediction models.
In particular, we integrate Evetac into a commercially available robotic parallel gripper (cf. \fig{fig:evetac_parallel_grip}) and present data-driven approaches for slip detection.
We train and compare different neural network architectures that benefit from the expressive, low-dimensional features.
The models can be evaluated online at \SI{1000}{\hertz} and form the basis for the design of a closed-loop grasp controller operating at \SI{500}{\hertz}, capable of stably grasping a wide range of objects with different masses and materials.

In summary, our contributions are the design of a novel, open-source event-based optical tactile sensor called Evetac.
The sensor design aims to maximize re-use of existing, commercially available components to mitigate the manufacturing barrier and incentivize reproducibility.
Besides sensor design, we demonstrate Evetac's high temporal resolution by sensing vibrations up to \SI{498}{\hertz} and showcase improved sensing efficiency w.r.t. data rate.
Despite Evetac's high sensing frequency, it generates significantly fewer data compared to RGB optical tactile sensors.
Moreover, we provide real-time touch processing algorithms.
We devise an algorithm for tracking the dots imprinted in the gel with \SI{1000}{\hertz} and show its effectiveness for reconstructing the shear forces acting upon Evetac's gel.
Lastly, we showcase Evetac's effectiveness for robotic grasping by training efficient data-driven neural networks for slip detection and prediction. 
The models allow integration into high-frequency feedback control loops for achieving robust and reliable grasping across a wide range of household objects.
Importantly, we are able to show generalization of the slip detectors across objects and the controller's adaptiveness w.r.t. object mass and reactiveness upon grasp perturbation.

\section{Related Works}

Tactile sensing \cite{kappassov2015tactile} has a huge potential for robotics.
Contact information is a crucial source of information to, e.g., recover object properties \cite{li2020review}, capture haptic information \cite{zhu2023visual}, or react to undesirable contact configurations for preventing slip and achieving stable grasping \cite{chen2018tactile, romeo2020methods}.
The following section mainly focuses on the latter aspect as progress in this direction holds the promise to improve the performance of every robotic manipulation system regarding reliability, robustness, and generalization to a wider range of objects.

\textbf{Vibration-based Tactile Sensing.} %
For stable grasping, humans make use of fast-adapting receptors to detect small localized slips that allow adaption of grasping force prior to gross slippage \cite{tremblay1992utilizing, johansson2018tactile}.
Inspired by these fast human mechanoreceptors, several tactile sensors have been developed, offering high temporal resolutions \cite{dahiya2009tactile, howe1989sensing, fishel2008robust, schurmann2012high}, including event-based tactile sensors~\cite{lee2015kilohertz, gupta2018neuromorphic, nakagawa2019bio, taunyazov2020event, taunyazov2021extended}.
The corresponding touch processing algorithms for slip detection investigate the energy of potential vibrations \cite{tremblay1992utilizing, fishel2012design, su2015force}, frequency-domain features in combination with neural networks \cite{schopfer2010using, schurmann2012high}, signal coherence analysis \cite{heyneman2013slip}, and data-driven approaches using the raw sensor data \cite{veiga2015stabilizing}.
Notably, \cite{taunyazov2020event} learn a fully asynchronous event-driven visual-tactile spiking neural network for slip detection.
Regarding slip timing, \cite{su2015force} showed that their approach can detect slip more than \SI{30}{\ms} before an IMU accelerometer picked it up.
While all these works present promising approaches to slip detection, almost all of them rely on special hardware, which is difficult to access and requires substantial manufacturing knowledge. 

\textbf{Optical Tactile Sensors} have recently gained increased attention. They offer small form factors, high spatial resolution, compatibility with standard interfaces, and have become relatively cheap to acquire, thereby, significantly reducing the entry barrier into the field.
As standard RGB cameras are typically significantly slower than the previously presented sensors, different slip detection criteria have been developed.
They include analysis of the marker displacement field \cite{yuan2015measurement, dong2017improved, li2018slip}, model-based criteria analyzing the inhomogeneity of the force field \cite{dong2019maintaining}, and data-driven slip detectors \cite{james2020biomimetic} eventually combined with closed-loop feedback control \cite{james2018slipsingle} and multi-fingered hands \cite{james2020slipmulti}.
Overall, these works on slip detection using RGB optical sensors rather focus on sensing displacements than high-frequency phenomena such as vibrations. 
Particularly, these RGB optical tactile sensors cannot offer the temporal resolutions found in human fast-adapting type II mechanoreceptors, which are sensitive to mechanical vibrations of at least up to \SI{400}{\hertz}\cite{johansson2009coding, dahiya2009tactile}.

\textbf{Event-based Optical Tactile Sensing.} Recent progress and commercialization in event-based cameras have led to the development of event-based optical tactile sensors that can eventually provide both high spatial and temporal resolution.
Ward et al.~\cite{ward2020neurotac} introduced the NeuroTac, an event-based optical tactile sensor based on TacTip~\cite{ward2018tactip}, showed its effectiveness for texture classification, and presented a miniaturized version~\cite{ward2020miniaturised}.
In similar efforts,~\cite{naeini2019novel, baghaei2020dynamic, naeini2022event} used an event-based camera behind a piece of silicone inside a parallel gripper, and investigated its effectiveness for force reconstruction and material classification for robotic sorting~\cite{huang2020neuromorphic}.
Compared to this paper, these works investigated different tasks.
In particular, they do not cover exploiting the sensors for high-frequency closed-loop feedback control for robotic manipulation.%

More closely related are \cite{kumagai2019event, rigi2018novel}.
Both investigate slip detection from event-based optical tactile readings, however, without considering closed-loop robotic manipulation.
The authors of~\cite{kumagai2019event} presented a marker-based tactile sensor that is read out using a temporal resolution of \SI{0.5}{\ms} and presented a proof of concept for slip detection using a hand-defined threshold.
Their proof of concept only included a single object, compared to 20 objects investigated herein.
Rigi et al.~\cite{rigi2018novel} placed an event-based camera behind a transparent silicone. 
They integrated the events for \SI{10}{\ms}, thereby operating at a 10 times reduced temporal resolution compared to this paper. 
They also employed a hand-defined threshold on the change in contact area between sensor and object to detect the onset of slip.  
They evaluated their approach on five different objects, which are all dark in color, as this benefits their approach.

Closest to this paper is the work of Muthusamy et al.~\cite{muthusamy2020neuromorphic}.
Their tactile sensor consists of an event-based camera placed behind transparent plexiglass, creating an event-based version of the Fingervision~\cite{yamaguchi2017implementing}.
They investigated two approaches for slip detection: one based on the raw spike count and another one on edge and corner features extracted from integrated images.
They also combined their model-based slip detectors with control.
Contrary to our proposed Evetac, their sensor is transparent and comes without any imprinted markers.
Therefore, their model-based slip detectors rely on the manipulated objects having sufficient texture, making them less general.
Additionally, the gel's transparency results in their sensor capturing not only contact-related phenomena but also events triggered, for instance, by moving background.
This might be disadvantageous as the background events are essentially noise when considering the task of slip detection.   

From a methodological point of view, all previous approaches for slip detection using event-based optical tactile sensors leveraged hand-designed, model-based criteria.
Herein, we take a different approach and use a model-free approach for learning slip detection and prediction models solely relying on labeled experimental data.
Our approach is thus not focused on pre-defined criteria, instead, during training, the neural network models are refined to automatically extract the most important information from the input features.
We propose efficient models suitable for real-time inference at \SI{1000}{\hertz} and demonstrate their integration into a real-time closed-loop grasp controller.
For automatic data labeling without compromising temporal resolution, we develop a new criterion based on optical flow.
Regarding input features, this work proposes a novel method for tracking the markers imprinted in the gel.
The marker tracking is capable of providing information about the gel's global deformation, complementing the raw sensor's sparse and local measurements. 
We showcase that this information benefits slip detection. 
Additionally, we contribute by open-sourcing Evetac's design.
In contrast, for the other event-based optical tactile sensors, the design files are not openly available.
Evetac's housing is the only custom component and can be recreated using an off-the-shelf 3D printer. All other components, including the soft silicone elastomer from GelSight Inc., are commercially available.

\section{Background - Event-based Cameras}
\label{sec:ebcameras}

Event-based cameras work fundamentally differently than standard cameras.
While in standard, frame-based cameras, every pixel is read out at a constant frequency, in event-based cameras, every pixel is independent and only reacts to brightness changes at its location, making the sensor asynchronous \cite{gallego2020event}.
Given the brightness of a pixel at position $x,y$, i.e., its log photocurrent $L(x,y,t) {=} \log(I(x,y,t))$ at time $t$, the pixel is sensitive to intensity changes $\Delta L(x,y,t) {=} L(x,y,t) {-} L(x,y,t_k)$ w.r.t. reference value $L(x,y,t_k)$.
If this intensity change reaches either the positive or negative threshold ($\pm C$) at time $t_{k+1}$, this pixel triggers a new event $e_{k+1}{=}(x,y,t_{k+1},p_{k+1})$, i.e., a tuple containing the event's location ($x,y$), timing ($t_{k+1}$), and polarity $p_{k+1} {\in} \{-1,1\}$, which signals whether the brightness increased or decreased.
Subsequently, the pixel's reference brightness value is adapted to $L(x,y,t_{k+1}) {=} L(x,y,t_k)+p_{k+1} C$, and from then on, the pixel is sensitive to changes w.r.t. the updated value.
Overall, the camera outputs this stream of events.
This asynchronous functioning principle offers many appealing properties, such as reduced power consumption, high dynamic range, lower latencies, and high temporal resolution \cite{gallego2020event}, which we, herein, aim to exploit in the context of touch sensing, processing, and robotic manipulation.

\section{The Evetac Sensor}

We now introduce our proposed Evetac sensor, a new event-based optical tactile sensor. 
The sensor consists of off-the-shelf components and a 3D printed case aiming to reduce the entry barrier into the field.
This section introduces Evetac's design, hardware components, and raw sensory output.

\begin{figure}[t]
    \centering
    \includegraphics[width=0.9\columnwidth]{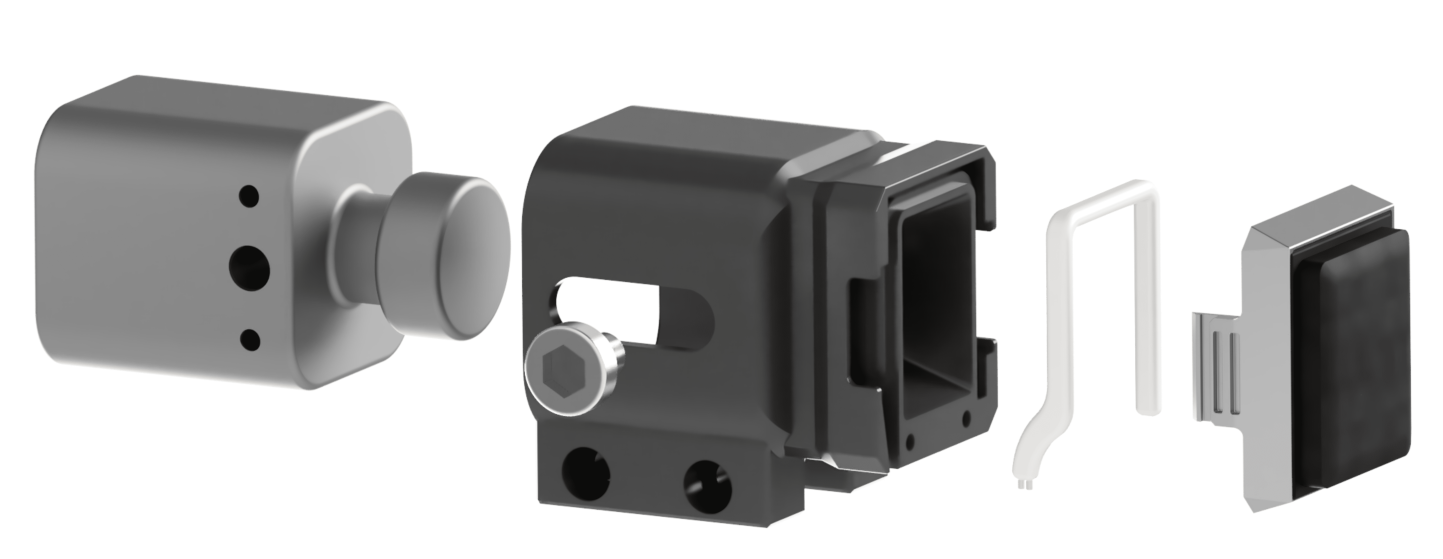}
    \caption{Exploded view of the proposed Evetac sensor. From left to right: A) DVXplorer Mini, event-based camera, B) 3D printed camera housing, C) LED stripe for illumination from the inside, and D) GelSight Mini dotted gel. The 3D printed housing allows adjusting the camera's distance from the gel to ensure that it is in focus. It also allows mounting Evetac to an external gripper (cf .~\fig{fig:evetac_parallel_grip}). For the components see Table \ref{table:hw_componets}. The total dimensions of the assembled sensor are $32 \mathrm{x}33 \mathrm{x} 65 \SI{}{\mm}$ (width x height x length).}
    \label{fig:evetac_design}
\end{figure}

\begin{table}[t]
\begin{center}
\caption{Evetac Hardware Components.}
\label{table:hw_componets}
\vspace{-0.2cm}
\scriptsize
\begin{tabular}{l|c}
Component & Specifications  \\
\hline
\hline
Camera   & DVXplorer Mini from Inivation, 640 x 480px resolution.  \\
\hline
Housing & Custom Design, 3D printed. \\
\hline
Camera Screws & 1/4", thread length 6.3mm \\
\hline
LED stripe & \makecell{LED COB band 4000K, \\ height 5mm, width 2.2mm, 12V.} \\
\hline
Gel & GelSight Mini Marker Gel. \\
\hline
\end{tabular}
\end{center}
\vspace{-0.5cm}
\end{table}

\subsection{Hardware}

\fig{fig:evetac_design} depicts Evetac's components.
It consists of an event-based camera capturing the deformation of a soft silicone gel.
The sensor is held together by a 3D printed housing and the illumination is provided by a white LED stripe surrounding the gel.
The housing is designed such that the distance between camera and gel can be adapted through two 1/4" camera screws. This allows to position the camera such that the gel is in focus.
The housing further offers a mounting mechanism on the bottom for integrating the sensor into parallel grippers or a single-finger (cf.~\fig{fig:evetac_parallel_grip} \& \fig{fig:force_reconstruct}).
We also designed a similar casing for the standard GelSight Mini such that we can mount one standard GelSight Mini and one Evetac inside a parallel gripper (cf.~\fig{fig:bandwidth_exp}).
All 3D printing files are open-sourced on~\href{https://sites.google.com/view/evetac}{our website}.
The two main design goals of Evetac were to ensure modularity and maximize reuse of existing components.
Apart from the 3D printed housing, all the components are commercially available.
The hardware components are also summarized in \tab{table:hw_componets}.
The event-based camera is a DVXplorer Mini from Inivation.
The soft silicone gel is the same dotted gel that is also used in the standard GelSight Mini.
The choice of this non-transparent, dotted gel (cf. \fig{fig:cut_gels}) ensures that the camera cannot see through the gel. It thus naturally focuses on capturing contact-related phenomena between sensor and object with minimum distraction.
This also implies that the sensor is agnostic w.r.t. the manipulated object being textured or not.
Moreover, the black dots that are imprinted in the gel can provide important information about the gel's current global deformation.
The dots provide natural features that can be captured by the camera and used in tasks such as shear force estimation or slip detection \cite{yuan2015measurement, dong2017improved, dong2019maintaining}, as we will later also show for Evetac in \sec{sec:evetac_characterization} \& \sec{sec:slip_detection}.

Besides the modular design, the event-based camera is the key component that differentiates Evetac from standard, classical optical tactile sensors such as GelSight \cite{johnson2011microgeometry}, DIGIT \cite{lambeta2020digit}, or TacTip \cite{ward2018tactip}.
The asynchronous functioning principle has many desirable properties. In this work, we particularly want to build upon the camera's sparse output, high temporal resolution, and low latency for high-frequency, real-time touch sensing, processing, and feedback control.

\subsection{Raw Sensor Output}
\label{sec:raw_sensor_output}

\begin{figure}[t]
\hfill%
\subfloat[Touching Evetac with the fingertip and moving the gel. \label{fig:finger_contact}]{\includegraphics[width=0.445\columnwidth]{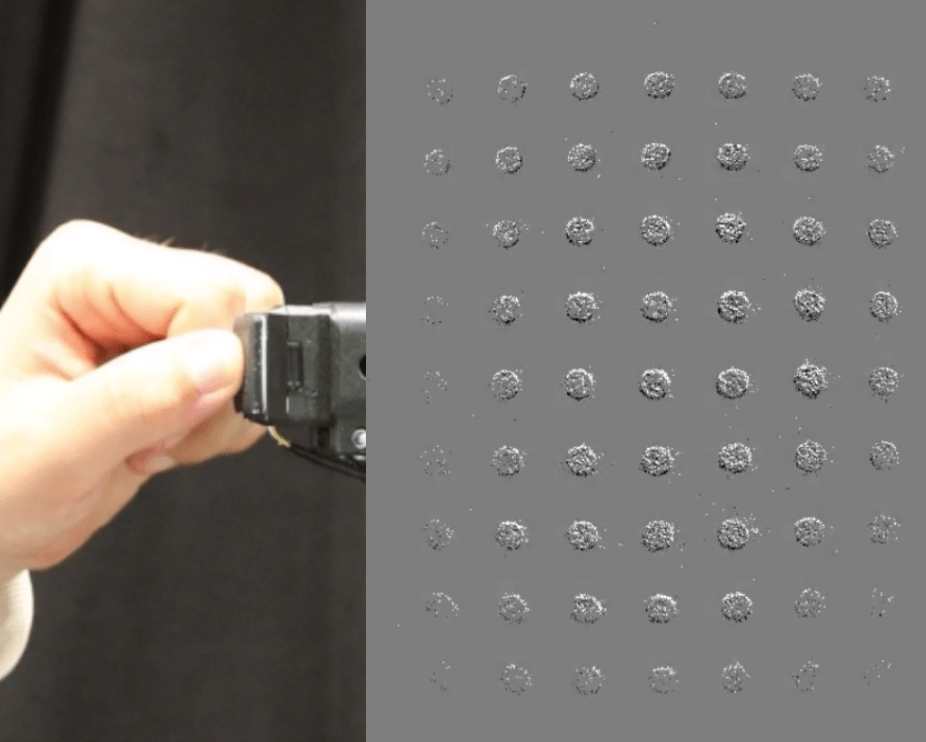}}%
\hfill%
\subfloat[Touching Evetac with an object and rubbing it fast over the gel. \label{fig:obj_contact}]{\includegraphics[width=0.445\columnwidth]{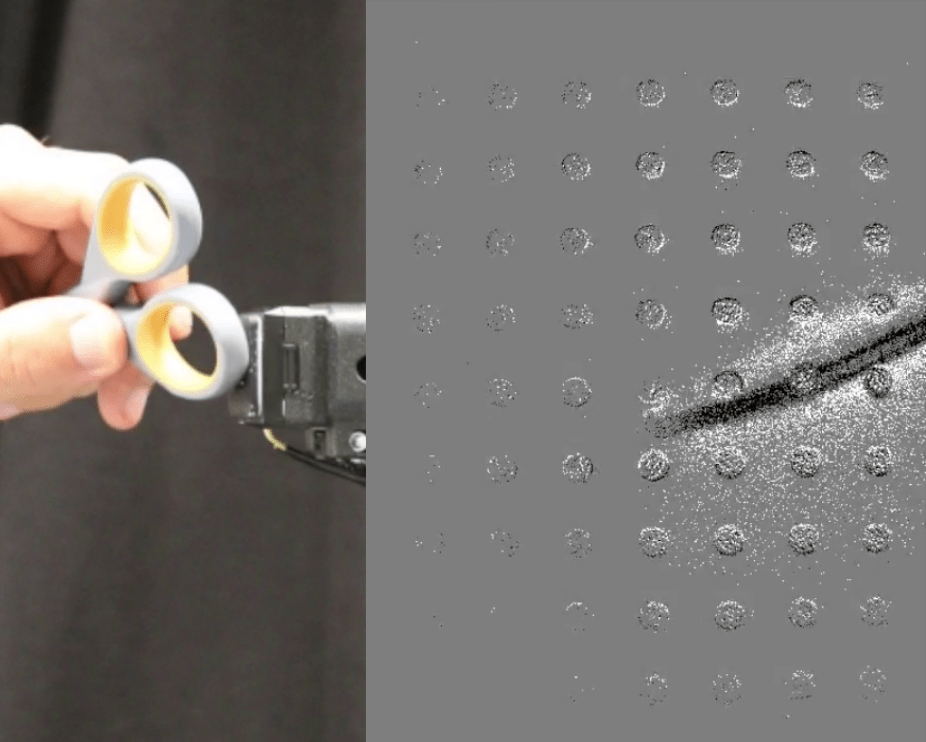}}%
\hfill%
\vspace{-0.35cm}
\begin{center}
\caption{Both pictures show the current contact configuration (left) \& Evetac's output in image form (right). 
As mentioned in \sec{sec:raw_sensor_output}, Evetac returns the raw events accumulated for \SI{1}{\ms}.
Since Evetac's raw output is extremely sparse, for visualization purposes, we actually illustrate the combination of the last 5 measurements, i.e., the events triggered within the last \SI{5}{\ms}.
In the pictures that show Evetac's raw output, all gray pixels correspond to locations where no events have been triggered, while the white \& black pixels illustrate locations of on \& off-events, respectively.} 
\end{center}
\vspace{-1.2cm}
\end{figure}

As mentioned in \sec{sec:ebcameras}, on the lowest level, event-based cameras return single events characterized by their location $x,y$, timing $t$, as well as polarity $p$, i.e., $e{=}(x,y,t,p)$.
Yet, when reading out event-based cameras with standard computers, the events are typically accumulated on the camera before they are sent via USB.
Herein, we configure the event-based camera such that the events are accumulated for \SI{1}{\ms} before they are sent to the computer.
Therefore, every millisecond we receive the set of $N_\mathrm{E}$ events $\mathcal{S}_\mathrm {E}(t_i) {=} \{e_k, k {\in} N_\mathrm{E}\}$, that have been created within the previous millisecond, i.e. $\forall e_k {=} (x_k,y_k,t_k,p_k) {\in} \mathcal{S}_\mathrm {E}(t_i): t_i {-} \SI{1}{\ms} {=} t_{i-1} {\leq} t_k {<} t_i$.
Evetac's current measurements, i.e., the received set of events, can also be visualized in image form, e.g., as shown in \fig{fig:finger_contact}.
For the image visualization, all pixel locations where no events occurred are colored in gray. The white pixels correspond to locations where on-events have been triggered (i.e., the pixel intensity increased), and the pixels in black correspond to locations of off-events.
As the sensor's output is very sparse, for visualization purposes, we actually show the combination of the last five measurements in all figures throughout the paper.
Thus, the images depict the events triggered within the past \SI{5}{\ms}.
The images have the same resolution as provided by the camera (640x480 pixels).
We want to point out that the information received from Evetac and its event-based camera is highly time-dependent.
When the gel's state does not change, no events are triggered, and no information is received.
If the gel is moved (e.g., as shown in \fig{fig:finger_contact} \& \fig{fig:obj_contact}), many events are triggered.
While it would be possible to configure Evetac's event-based camera to return event sets even more frequently, we believe that our choice of grouping the events for \SI{1}{\ms}, thereby receiving measurements at \SI{1000}{\hertz}, provides a good tradeoff between temporal resolution and computational feasibility of additional signal processing algorithms.
In terms of software implementation, we build on top of the dv-processing library provided by Inivation \cite{inivation_dv_processing}. We extend their code with our touch processing methods presented in the next section and also implement a ROS interface, facilitating data recording and visualization.

\section{Evetac Touch Processing}

In its standard configuration, Evetac returns event sets at \SI{1000}{\hertz} (cf. \sec{sec:raw_sensor_output}).
While it would be possible to directly exploit Evetac's raw sensory output for solving tasks such as slip detection, this section introduces additional touch processing algorithms operating on top of Evetac's output.
These algorithms and their output aim to provide meaningful intermediate representations.
In particular, they also further compress the dimensionality of Evetac's raw output, which benefits meeting computational requirements and ultimately achieving real-time feedback control.
Moreover, the proposed dot tracking algorithm is designed for keeping track of Evetac's global configuration, i.e., its gel configuration, despite the sensor's sparse measurements triggered by local changes.%

\subsection{Signal Processing - Dot Tracking}
\label{sec:dot_tracking}

\begin{figure}[t]
	\centering
	\def\svgwidth{0.95\columnwidth} %
        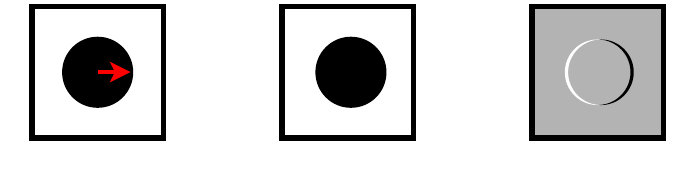 
	\caption{Illustrating how the movement of a black dot in front of a white, bright background triggers events. Left: At time $t_i$, the dot is moving to the right. Middle: This results in a slightly shifted position at time $t_{i+1}$. Right: The dot movement between $t_i$ and $t_{i+1}$ causes events $\mathcal{S}_\mathrm {E}(t_{i+1})$ that are visualized in pictorial form. The pixels colored in white correspond to the locations where on-events were triggered. The brightness of all these pixels changed from being occupied by the black dot at $t_i$ to being occupied by the white background at $t_{i+1}$. Due to this intensity change, events have been triggered at these locations. For the pixels colored in black, the opposite happened, i.e., the brightness changed from the white background to the black dot. At all the remaining pixel locations, no events have been triggered. They are thus colored in neutral grey.}%
	\label{fig:dot_track_visual}
 \vspace{-0.5cm}
\end{figure}

Evetac's raw sensory output solely contains local, relative information.
This is fundamentally different from the raw output of RGB optical tactile sensors, which, at every measurement, return a complete image of the gel.
While the former might be more efficient from the perspective of only retrieving information when changes occur, it comes at the disadvantage of not being able to reconstruct the gel's global configuration given a single measurement. When no changes are happening across the gel surface, Evetac's measurements will not contain any events, and, therefore, provide no information.
As mentioned earlier, Evetac's gel contains imprinted markers, i.e., dots (cf. \fig{fig:cut_gels}), to visualize gel movements.
Their positioning provides information about the current gel deformation.
Since this global information might be important for tasks such as shear force reconstruction or slip detection, we next provide an algorithm for tracking the dots' positions.

Our proposed dot tracking algorithm builds upon the work of \cite{ni2015visual, ward2020neurotac}, who presented a model-based tracker for event-based cameras. 
The algorithm is based on the assumption that all triggered events are caused by the movement of the object that is to be tracked.
Since event-based cameras register changes in lighting intensity, it is the edges of uniformly colored objects that trigger events.
The two leftmost frames of \fig{fig:dot_track_visual} illustrate the scenario of a black dot (i.e., the object) moving to the right between times $t_i$ and $t_{i+1}$. As shown in the rightmost frame of \fig{fig:dot_track_visual}, in the direction of movement, pixels change from the white, bright background to the black color of the moving dot. 
This change, i.e., decrease, in brightness triggers negative off events. 
Opposite to the dot's moving direction, pixels change from the dot's black color to the bright background, yielding positive on events.
Exploiting this insight that the edges of a uniformly colored object trigger events upon movement, the model-based tracking algorithm consists of two main steps.
First, finding correspondences between the triggered events and the object's edge points.
Second, updating the object pose estimate given the established correspondences.

In the following, we present the algorithm for tracking the markers, i.e., the dots that are imprinted in the gel.
Without loss of generality, for the derivation, we focus on tracking one of the dots.
The dot's pose is parameterized by rotation $R(\theta_i) {\in} SO(2)$ and translation vector $\vc_i{=}[c_{x_i}, c_{y_i}]^T {\in} \mathbb{R}^2$.
Given that we want to track a dot, which on average has a radius of 15 pixels, and that the dot's edge triggers events, the tracker first prefilters the raw events.
Only events within a ring of 10 pixels inner radius and 20 pixels outer radius are considered for updating the dot's pose, i.e., $10{<}{\norm{\vx_k {-} \vc_i}_2^2}{<}20$.
The area of this receptive field has been determined empirically and takes into account that the size of the dot might vary depending on the gel deformation and that the current dot pose estimate is not granted to be perfectly accurate.
Assuming that Evetac registered a single event $e_k(x_k,y_k,t_k,p_k) {=} (\vx_k, t_k, p_k)$ within this region, in the first step of correspondence matching, we find the closest point on the edge of the dot $\vx^{m}_j {\in} \mathbb{R}^2$ that could have caused this event.
We create an assignment between event and dot edge point $k {\rightarrow} j$ minimizing $d(k,j){=}{\norm{\vx_k{-}\vx^{m}_j}_2^2}$.
Given the correspondence, we update the dot's pose through rotation matrix $\tilde{R}(\theta) {\in} SO(2)$ and translation vector $\vct{=}[c_x, c_y]^T$, attempting to correct the dot pose estimate such that it explains the observation.
This is done by minimizing objective function $f {=} {\norm{\vx_k {-} (\tilde{R}(\theta) \vx^{m}_{j {\rightarrow} k} {+} \vct) }}_2^2$.
While it would be possible to find the unique optimal solution to minimize $f$, the authors of \cite{ni2015visual} proposed to apply a gradient-based update, i.e., $\theta_{i+1} {=} \theta_i {-} \alpha \nabla_\theta f |_{\theta=0, \vct=[0,0]^T}$, and $\vc_{i+1} {=} \vc_i {-} \alpha \nabla_{\vct} f|_{\theta=0, \vct=[0,0]^T}$.
One motivation for the gradient-based update is that the event data is inherently sparse.
Also, in case of multiple events, the gradients can simply be accumulated.
Therefore, the gradient magnitude changes with the number of events.
Exemplary, in scenarios with only a few events, the gradient-based update only slightly adjusts the dot's pose, instead of greedily converging to the best pose.
This is advantageous since not all events are triggered by dot movements.
Especially in low event scenarios it might happen that the majority of events come from sensor noise.

For our particular case of dot tracking, we can further simplify the update rule and correspondence matching.
First, due to the rotational symmetry of the dot, i.e., the object that is to be tracked, we can omit optimizing its orientation.
Second, while \cite{ni2015visual} define the object's edges through a discrete set of points, for tracking dots, we can analytically calculate the closest object edge point $\vx^{m}_{j \rightarrow k}$ for every event at location $\vx_k$.
Using geometry, the closest dot edge point has to lie at dot radius $r$ away from the dot's center point, in the direction of the event, i.e., $\vx^{m}_{j \rightarrow k} {=} r \vx_k / {{\norm{\vx_k}}}_2$. 
Note that this assumes that the event's coordinate $\vx_k$ is already given with respect to the dot's current center location $\vc_i$.
Subsequently, the objective for adapting the dot's location equates to
\begin{equation}
\begin{split}
    f &= {\norm{\vx_k {-} (\vx^{m}_{j \rightarrow k} {+} \vct) }}_2^2 = {\norm{\vx_k {-} (r \vx_k / {{\norm{\vx_k}}}_2 {+} \vct) }}_2^2 \mathrm{ .}
\end{split}
\end{equation}
The gradient can be obtained in closed form as
\begin{equation}
    \nabla_\vct f |_{\vct=[0,0]^T} = -2(\vx_k-r \vx_k / {{\norm{\vx_k}}}_2) \mathrm{ .}
\end{equation}
In conclusion, we obtain a closed-form solution for updating the dot's center coordinate, given an event at location $\vx_k$.
If more than one event is associated to a certain dot, i.e., when having to deal with a set of events $\mathcal{S}_\mathrm {E}(t_{i+1},\vc_{i})$, the resulting update is the sum of the updates for every individual event 

\begin{equation}
 \nabla_\vct f(\mathcal{S}_\mathrm {E}(t_{i+1},\vc_{i})) |_{\vct=[0,0]^T} = {\sum_{l=1}^{|\mathcal{S}_\mathrm {E}|}} {-}2(\vx_l{-} r \vx_l / {{\norm{\vx_l}}}_2) \mathrm{ .}
\end{equation}

Initial testing of this gradient-based tracking scheme showed that it worked well in conditions where the dots' movement caused the majority of the events.
Yet, the algorithm is prone to losing track when an external object moves quickly over the gel. 
In such situations, the stream of events caused by the edges of the moving object can outweigh the events triggered by the dots' movement, as shown in \fig{fig:obj_contact}.
This might result in our tracking algorithm following the stream of events caused by the external object, thereby losing track. %
This issue is especially problematic when considering that Evetac itself only returns sparse, relative measurements.
It is, therefore, very difficult to recover the dots' poses once track has been lost since single Evetac measurements are not sufficient.
To counteract the issue and attempt to prevent losing track at first place, we propose the following addition to the algorithm. 
The addition is a regularizing term, emerging from the idea that despite the gel deforming, the dots should still roughly stay within their original grid-like structure.
In other words, the dots' positions relative to each other should not change too drastically.
Without loss of generality, we now consider a pair of neighboring dots at locations $\vc^1_i$ and $\vc^2_i$ which have been at an initial distance $d_{1,2}={\norm{\vc^1_0-\vc^2_0}_2^2}$ at the start of the tracking for which we assume that nothing is pressing against Evetac. 
Exemplarily, if we want to regularize the position update $\vct$ of dot 1 based on its initial distance to dot 2, the updated objective with weighting factor $w_{\textrm{dist}}$ equates to
\begin{equation}
    f_\mathrm{reg} = f + w_{\textrm{dist}} f_{\textrm{dist}} {=} f + w_{\textrm{dist}} ({\norm{(\vc_i^1+\vct)-\vc_i^2}}_2^2 - d_{1,2})^2 \mathrm{ .}
\end{equation}
The additional term basically penalizes moving too far or too close to the other dot, taking the initial distance as reference.
Calculating the gradient for the regularizing term results in
\begin{equation}
    \nabla_\vct f_{dist} |_{\vct=[0,0]^T} = 4 (\vc_i^1 -\vc_i^2) ({\norm{\vc_i^1 -\vc_i^2}}_2^2 - d_{1,2}) \mathrm{ .}
\end{equation}
While this example only considered a pair of dots, the same regularization term is added for all direct neighbors of a dot. 
A dot can have at maximum 8 neighbors (i.e., along the horizontal, vertical, and the two diagonals), and at minimum three for the dots in the corner.
Due to these differences in the number of neighbors, $w_{\textrm{dist}}$ is scaled by 8 divided by the dot's actual number of neighbors such that the regularizing term is of similar magnitude for all dots. 
Lastly, we want to point out that a dot's location is only updated if more than 10 events are triggered at its location.
This reflects the fact that an actual dot movement should always trigger multiple events at once and thus aims to improve the tracker's robustness w.r.t. sensor noise.
Moreover, without this additional constraint, the regularizing term might cause a dot to move without any events being present at its location.

\subsection{Evetac Touch Features}

Before proceeding with the next sections, in which we will use Evetac to infer contact-related phenomena, we provide a short overview of the features that we consider in this work.
While the sensor's raw measurement (cf. \sec{sec:raw_sensor_output}) provides a set of events $\mathcal{S}_\mathrm {E}(t_{i})$ at current time $t_i$ at a spatial resolution of 640x480 pixels, the next sections will mainly exploit lower dimensional features.
Building on top of these lower dimensional features facilitates meeting Evetac's real-time requirements, as any further processing also has to be able to handle the data at a frequency of \SI{1000}{\hertz}.
In the remainder of the paper, we will consider the following features:
\begin{itemize}
    \item Overall, raw number of events, $N_\mathrm{E}(t_i) = |\mathcal{S}_\mathrm {E}(t_{i})|$.
    \item Raw number of events per dot, i.e., for dot at location $\vc_i$, $N_\mathrm{E}(t_i, \vc_i) = |\mathcal{S}_\mathrm {E}(t_{i}, \vc_i)|$. This quantity relies on the dot tracking algorithm estimating the dot's center location $\vc_i$. An event is associated with a dot if the event's location is less than 20 pixels away from the dot's center.
    \item Number of events for all the $N_\mathrm{D}$ dots $E_\mathrm{C}(t_i) = \{ N_\mathrm{E}(t_i, \vc^l_i), l \in N_\mathrm{D} \}$   
    \item Positions of all of the $N_\mathrm{D}$ dots $P_\mathrm{C}(t_i) = \{ \vc^l_i, l \in N_\mathrm{D} \}$
    \item Displacement of the dot at current location $\vc_i$ w.r.t. its initial location $\vc_0$, i.e., $d_{\vc_i} = {\norm{\vc_i-\vc_0}}_2$
    \item Displacements of all of the dots $D_\mathrm{C}(t_i) = \{ d_{\vc_i}^{l}, l \in N_\mathrm{D} \}$
    \item Set of events in image form $\mathcal{I}(t_i)=\mathcal{S}_\mathrm {E}(t_{i})$ (see e.g. \fig{fig:finger_contact}-right)
\end{itemize}
Our reasoning for this choice of features is that the raw number of events might be beneficial for resolving very fast phenomena, while the displacement features will provide more information about the gel's current deformation.
Additionally, we will exploit the set of events in image form for training baseline models.
We nevertheless want to point out that this list only covers a very small portion of possible features that could be extracted from Evetac's raw output.
It might well be that there exist more powerful and informative features that would improve Evetac's performance on downstream tasks.

\section{Benchmarking Evetac}
\label{sec:evetac_characterization}

In this section, we present four experiments to showcase Evetac's properties.
First, we use Evetac to sense vibrations up to \SI{498}{\hertz} and validate its high temporal resolution.
Second, we compare the data rate of Evetac and a RGB optical tactile sensor in a grasping and slipping experiment, highlighting the advantages of Evetac's sparse output.
Third, we investigate the effectiveness of the regularizing term for dot tracking.
Lastly, we evaluate the quality of the proposed regularized dot tracker through the task of shear force reconstruction based on the dot's positions.
Additional videos are available on \href{https://sites.google.com/view/evetac}{our website}.

\subsection{Sensing Vibrations}
\label{sec:sensing_vibrations}

\begin{figure}[t]
\vspace{-0.25cm}
\hfill
\subfloat[\label{fig:vibration_sensing}]{\includegraphics[width=0.45\columnwidth]{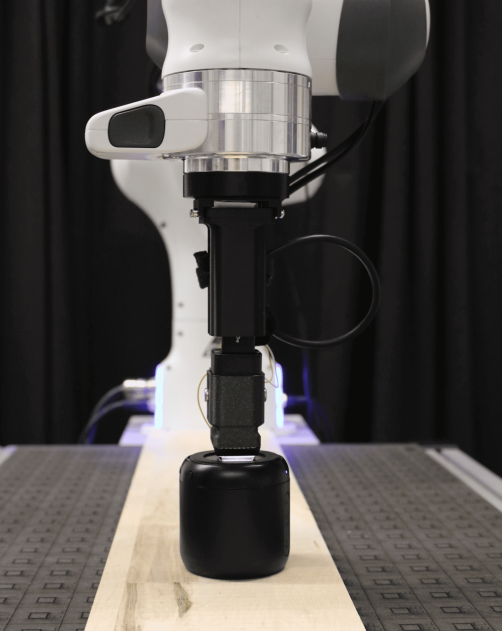}}
\hfill
\subfloat[\label{fig:force_reconstruct}]{\includegraphics[width=0.45\columnwidth]{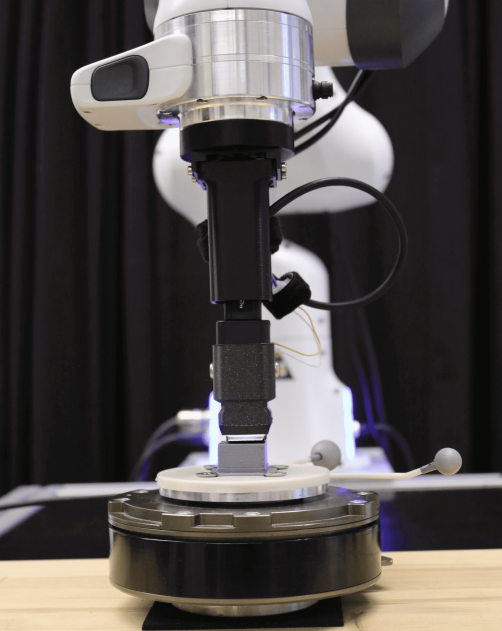}}
\hfill
\vspace{-0.35cm}
\begin{center}
\caption{Experimental setups for the Evetac benchmarking experiments of sensing vibrations (cf. \sec{sec:sensing_vibrations}, \textbf{(a)}) and shear force reconstruction (cf. \sec{sec:shear_force_reconstruction}, \textbf{(b)}). For both experiments, Evetac is mounted in the end-effector of a Franka Panda robot. 
In \figs{fig:vibration_sensing}, Evetac presses against a speaker which is set to generate a tone with a desired frequency. Through making contact with the speaker, Evetac can perceive the vibrations of the speaker and reconstruct the vibration frequency (cf.~Tab.~\ref{table:vibration_sensing}).
In \figs{fig:force_reconstruct}, Evetac presses against an object mounted on top of a F/T sensor. Through moving the robot, we shear the gel. By combining our proposed dot tracking algorithm with a model, we attempt to recover the shear forces (cf.~\figs{fig:force_recon}). 
\label{fig:evetac_char_exps}
} 
\end{center}
\vspace{-.5cm}
\end{figure}

This experiment aims to verify Evetac's high temporal resolution of receiving sensor feedback at \SI{1000}{\hertz} by measuring vibrations generated by a speaker.
According to Nyquist–Shannon sampling theorem \cite{shannon1949communication}, given sampling frequency $f_s$, perfect signal reconstruction is possible for bandwidth $B < f_s /2$, i.e., $B < \SI{500}{\hertz}$ for our case.
The experimental setup is shown in \fig{fig:vibration_sensing}.
We attach Evetac to a 3D-printed mount that can be screwed to the mounting flange of the Franka Panda 7DoF robot.
The robot presses Evetac against a commercially available Bluetooth speaker (Anker Soundcore mini) with a predefined normal force of \SI{2.5}{\N}.
The tone frequency played by the speaker $f_d$ is controlled through the mobile phone app ``Tone Generator`` which generates pure sine wave tones at the desired frequency.
The speaker is placed upside down and set to maximum volume. Evetac presses against the bottom of the speaker. We ensure that Evetac directly touches the speaker's metal housing and not the rubbery ring on the edge, as the rubber will dampen the vibrations. On the other side, the speaker's housing is in direct contact with a piece of wood.
This experimental setup ensures that there is no dampening material between the metal housing of the speaker and the piece of wood / Evetac. Therefore, the speaker will vibrate when generating the sound.

Since Evetac is touching the speaker, the vibrations will be transmitted to the gel and make the black dots within the gel move.
Evetac's event-based camera will capture this movement of the dots, and events will be generated proportional to the velocity of the dots or changes in their size.
We, therefore, expect the number of events to oscillate with the same frequencies that are triggered by the speaker's vibrations.
The most excited frequency component should coincide with the tone frequency $f_d$ that the speaker is tasked to play.
For recovering $f_d$ from the sensor readings of Evetac, we propose to apply a Fourier transform to the absolute number of events ($N_\mathrm{E}(t_i)$) over a time series of length $T_w$.
Given the spectrum in the frequency domain, we remove all frequency components smaller than \SI{25}{\hertz}, as we are only interested in sensing higher frequency vibrations.
Subsequently, we select the frequency component that exhibits maximum amplitude.
If this frequency is within \SI{\pm1}{\hertz} of the tone frequency that we set the speaker to, we label this as a correct detection of the vibration.

\begin{figure}[t]
	\centering
	\def\svgwidth{\columnwidth} %
        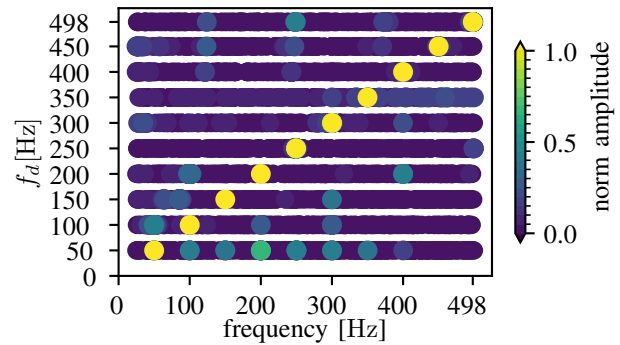 
	\caption{Frequency spectrum of the vibration sensing experiment. The plot shows the normalized frequency spectrums (i.e., amplitudes normalized to [0,1]) measured by Evetac, when pressing against a speaker which is set to different frequencies $f_d$ (cf. \fig{fig:vibration_sensing}). The spectrums are obtained by performing a Fourier transform on the total number of events measured per millisecond for a duration of $T_w=\SI{10}{\s}$. As shown, Evetac can recover the main vibration frequency of the speaker up to frequencies of \SI{498}{\hertz}. This confirms the sensor's high temporal resolution and its readout frequency of \SI{1000}{\hertz}.}%
	\label{fig:freq_analysis}
\end{figure}

\begin{table}
\begin{minipage}{\columnwidth}
\centering
\caption{Numerical results for the vibration sensing experiment (cf. \figs{fig:vibration_sensing}). The table reports the success percentage and successful detections of the speaker's frequency $f_d$ considering different time windows $T_w$. The results underline Evetac's high temporal resolution. Vibrations up to \SI{498}{\hertz} can be detected reliably.}
\label{table:vibration_sensing}
\begin{tabular}[t]{l|ccc}
Speaker  & \multicolumn{3}{c}{Detected Frequency} \\
Frequency & $T_w = \SI{10}{\s}$ & $T_w = \SI{2}{\s}$ & $T_w = \SI{1}{\s}$ \\
\hline
\hline
\SI{100}{\hertz} & 100\% (10/10) & 98\% (49/50)  & 96\% (96/100)  \\
\SI{200}{\hertz}  & 100\% (10/10) & 92\% (46/50) & 69\% (69/100)  \\
\SI{300}{\hertz}  & 100\% (10/10) & 100\% (50/50) & 99\% (99/100) \\
\SI{400}{\hertz}  & 100\% (10/10) & 100\% (50/50)  & 98\% (98/100) \\
\SI{498}{\hertz}  & 100\% (10/10) & 100\% (50/50) & 100\% (100/100) \\
\hline
Avg. &  100\% (50/50) & 98\% (245/250) & 92.4\% (462/500) \\
\hline
\end{tabular}

\end{minipage}
\end{table}

For evaluation, we set the speaker to play tones with 5 different frequencies ranging from \SI{100}{\hertz} to \SI{498}{\hertz}.
For each frequency, we create a recording for \SI{100}{\s}.
We divide the entire recording into pieces of length~$T_w$ and apply our procedure to recover the tone frequency played by the speaker ($f_d$) to each sequence individually.
The results are presented in \fig{fig:freq_analysis}) and \tab{table:vibration_sensing}.
Overall, the sensor and processing pipeline are capable of sensing the speaker's tone through the vibrations with high success rates.
Considering longer time series, i.e., $T_w{=}\SI{10}{\s}$, it is possible to perfectly recover the vibration frequency. For $T_w{=}\SI{2}{\s}$, i.e., dividing the trajectory into 50 segments of length 2 seconds each, only in 5 out of 250 segments, the vibration frequency of the speaker ($f_d$) was only the second most excited frequency component, while half of the tone frequency ($f_d/2$) was the most excited one. 
Considering shorter time windows such as $T_w{=}\SI{1}{\s}$, the performance of identifying $f_d$ through the most excited frequency component again slightly decreases.
In the shorter time windows, it happens more frequently that either half or double of the tone frequency $f_d$ are the most excited components. 
Nevertheless, the speaker's vibration frequency is still always amongst the three most excited components.
Overall, we conclude that Evetac is able to accurately sense high-frequency vibrations. We thereby also validate Evetac's high temporal resolution. The highest frequency that was able to be recovered was \SI{498}{\hertz}, which is very close to the Nyquist frequency of our setup, i.e., \SI{500}{\hertz}.

\subsection{Data Rate Experiments}
\label{sec:bandwidth_experiments}

\begin{figure}
	\centering
	\def\svgwidth{\columnwidth} %
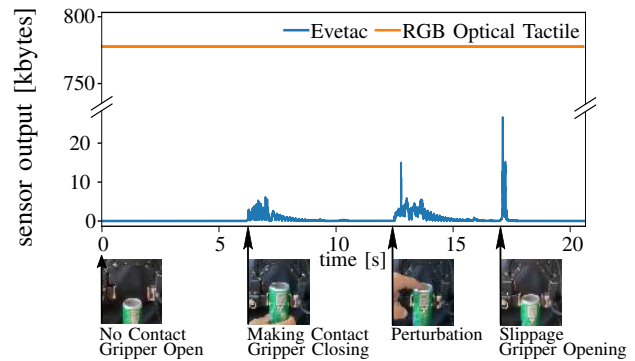 
	\caption{Data rate comparison between Evetac and a RGB optical tactile sensor (GelSight Mini), considering a maneuver of grasping an object, perturbing the object, and finally forcing object slippage. Evetac's output size correlates with changes in contact configuration, and, per measurement, is of significantly smaller size. 
 Considering the whole trajectory, Evetac only produces 1.7\% of the data of the RGB optical tactile sensor.
 For more numerical results, see \tab{table:bandwidth_comparison_numbers}.}%
	\label{fig:bandwidth_exp}
 \vspace{-0.25cm}
\end{figure}

Next, we compare the sensor data rate of our proposed Evetac and a standard RGB optical tactile sensor (GelSight Mini \cite{gelsightGelSightMiniStore}), considering the sensors' raw outputs.
Since a reduced sensor data rate correlates with less information that needs to be processed, depending on the application, the sensor data rate might be a crucial property.
For instance, if multiple tactile sensors should cover a larger area, e.g., an entire robot's surface.
As experimental setup, we equip a parallel gripper with one Evetac and one GelSight Mini (cf. \fig{fig:bandwidth_exp}).
As shown in \fig{fig:bandwidth_exp} and the supplementary videos, for data rate comparison, we start with the gripper open, close it to make contact with the object, and establish a predefined grasping force for stable grasping.
Thereafter, we mimic perturbations that might occur during manipulation by pressing onto the object.
Since the sensors' gel is elastic, this will cause the object to oscillate within the fingers without loosing contact.
Finally, we open the gripper to cause object slippage.

For the data rate comparison, we make the following assumption.
Since we used a modified version of the gels, which will later be introduced in more detail (cf. \sec{sec:slip_detection}), we only consider a sensing area of 540 by 480 pixels for both sensors.
Yet, we want to point out that the reduced sensing area has no effect on the results.
The RGB optical tactile sensor, i.e., GelSight Mini, returns 3 color values per pixel. 
Each of them is in the range of 0-255 and can be represented by 1 byte.
Thus, one reading from the RGB optical tactile sensor has $540{*}480{*}3{=}777600$ bytes~${=}777.6$ kbytes.
Evetac returns a set of events with varying size every millisecond (cf. \sec{sec:raw_sensor_output}).
Every event can be represented by 5 bytes - we require 2 bytes to encode the event's x coordinate, 2 bytes for the y coordinate, and 1 byte for its polarity.

\fig{fig:bandwidth_exp} illustrates the two sensors' outputs in bytes for one of the exemplary trajectories.
As can be seen, while the synchronous RGB optical tactile sensor always returns a fixed-size output, Evetac is more selective.
Evetac's output exhibits clear spikes whenever something happens at the contact location, as its output size correlates with the number of events.
At all other times, almost no sensor output is generated.
Numerical results are presented in \tab{table:bandwidth_comparison_numbers}. The table reports the ratio between data rate (i.e., bytes/\SI{}{\s}) used by Evetac w.r.t. data rate generated by the RGB optical tactile sensor for different parts of the trajectory.
Although Evetac is read out at a 40 times higher frequency (\SI{1000}{\hertz} vs \SI{25}{\hertz}), it only produces 1.7\% of the RGB optical tactile sensor's data rate considering the entire trajectories.
Even when only considering a \SI{0.5}{\s} interval around the moment of slippage, which is the point in time when Evetac returns most information, it still only generates around 11.9\% of the RGB optical tactile sensor's data.

This experiment underlines Evetac's sensing principle of only returning information upon changes in contact configuration. Despite Evetac's substantially increased sensing frequency, this results in producing considerably less data compared to the raw output of a standard RGB optical tactile sensor that always returns fixed-size measurements.
Nevertheless, Evetac's reduced output comes at the cost of only returning information upon changes in contact configuration.
Compared to RGB optical tactile sensors, it is, therefore, impossible to recover the gel's global configuration given only a single measurement.

\begin{table}
\centering
\captionof{table}{Comparing the relative data rate of Evetac w.r.t. the RGB optical tactile sensor averaged over 5 trajectories (reporting mean \& standard deviation). See~\fig{fig:bandwidth_exp} for one of them. Since Evetac's output is correlated with changes in contact configuration, different time intervals are considered, i.e., 1) the entire trajectories, 2) the timespan between making contact and object slippage, and 3) a \SI{0.5}{\s} interval around the moment of slippage.} 
\label{table:bandwidth_comparison_numbers}
\scriptsize
\begin{tabular}[t]{c|ccc}
data rate ratio  & entire trajectory & \makecell{making - \\ breaking contact} &  \makecell{slip \\ only} \\
\hline
\hline
\makecell{Evetac / RGB Optical \\ Tactile Sensor}   & \textbf{0.017} (0.004) & 0.028 (0.008) & 0.119 (0.006) \\
\hline
\end{tabular}
\end{table}

\subsection{Effectiveness of Regularization in Dot Tracking}

This experiment investigates the effectiveness of the regularized dot tracking algorithm (cf. \sec{sec:dot_tracking}).
As mentioned previously, the dot tracking algorithm assumes that dot movements trigger all events.
While this holds true for some tactile interactions (e.g., see \fig{fig:finger_contact}), in others, the edges of an object sliding over the gel might trigger the majority of the events (cf. \fig{fig:obj_contact}).
These additional events, which are not triggered by dot movements, contradict the tracker's assumptions and might cause it to lose track.
To counteract the dot predictions erroneously following the events triggered by the moving object, we introduced a regularization term, regularizing the dots' positions relative to each other based on their initial distance.

This experiment compares the regularized and the unregularized version of the dot tracking algorithm.
We consider 10 trajectories in which we either make contact with Evetac using a finger or the handle of a scissor.
Snapshots from two trajectories are shown in \fig{fig:finger_contact} and \fig{fig:obj_contact}.
After making contact, we performed several movements to shear the gel.
Most importantly, we also ensure that there is slippage, i.e., relative motion between the sensor and object, during which also the object's edges and shape will trigger events.
All the trajectories start and end with nothing touching the gel.
Thus, the dots should be at their equilibrium position at the start and end.
Yet, if track is lost during the trajectory, this will not be the case.
For comparing the two dot tracking versions, we thus investigate end-point consistency.
In particular, we report the percentage of trajectories for which all the dots end up close to their initial position, i.e., within a radius of 20 pixels from their initial position.
Additionally, for the unsuccessful trajectories, we report the mean number of dots for which track was lost ($N_{lt}$).

\tab{table:dot_tracking} shows the results and underlines the effectiveness of the regularized dot tracking.
The regularizer helps in increasing the number of successful trajectories by a factor of 4, from two to eight.
Moreover, even in the unsuccessful cases, the number of dots for which track was lost is substantially decreased.
We also provide supplementary videos comparing the two trackers across the trajectories.
The videos underline that the regularized tracker is substantially less prone to loosing track of the dots.
They also show that qualitatively, the regularized tracker does not compromise tracking quality throughout the trajectories and follows well the deformations of the gel.
We nevertheless want to point out that the trajectories in this experiment rather covered extreme cases. As we show on our website, when an object slips, normally, most of the events are still triggered by the dots.
We still think that a robust dot tracker is important as Evetac and the touch processing algorithms should be capable of dealing with a wide range of scenarios.
Thus, throughout all of the following experiments, we use the regularized version of the tracker.

\begin{table}[t]
\centering
\caption{Evaluating the effectiveness of the regularized dot tracking.
We report on how many out of 10 trajectories the respective tracking algorithm successfully completed the dot tracking.
Since all trajectories start and end with the gel in its equilibrium configuration, successful tracking is defined through end-point consistency, i.e., all dots ending up close to their initial position.
We also report the average number of dots for which track was lost during the unsuccessful trajectories ($N_{lt}$).} 
\label{table:dot_tracking}
\scriptsize
\begin{tabular}[t]{c|cc}
Tracking Algorithm Version  & Success Rate & $N_{lt}$ \\
\hline
\hline
Unregularized  & 20\% (2/10) & 9.5 \\ \hline
Regularized  & \textbf{80\%} (8/10) & \textbf{1.5} \\
\hline
\end{tabular}
\end{table}

\subsection{Dot-based Shear Force Reconstruction}
\label{sec:shear_force_reconstruction}

\begin{table}
\begin{minipage}{\columnwidth}
	\centering
	\def\svgwidth{\columnwidth} %
        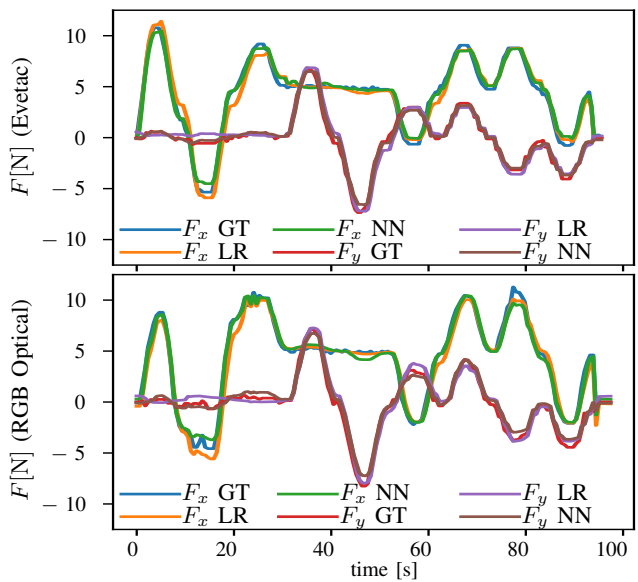 
	\captionof{figure}{Shear force reconstruction from tracking the dots of Evetac (top) and a RGB optical tactile sensor (bottom). For the experimental setup, see \figs{fig:force_reconstruct}. GT corresponds to ground truth measurements of a F/T sensor, LR to the linear regression model, and NN to the neural network model. Qualitatively, the shear force reconstruction leads to similar quality despite the different sensor types. Since the reconstruction is solely based on the dots' displacement, we conclude that our presented regularized dot tracking tracks the dots with good accuracies, despite the sparse and relative output of Evetac. Numerical results are presented in \tab{table:shear_force}.}%
	\label{fig:force_recon}
 \end{minipage}
 
\vspace{-.25cm}
\end{table}

\begin{table}
\centering
\caption{\small
Numerical Results for Shear Force Reconstruction experiment (cf. \figs{fig:force_recon}). \label{table:shear_force}
}
\vspace{-0.3cm}
\scalebox{0.9}{
\begin{tabular}{c|c|cc}
    &  & \multicolumn{2}{c}{MAE [N]} \\
Method                     & Sensor               & $F_x$   & $F_y$ \\ \hline
\multirow{2}{*}{Linear Regression}   & Evetac & 0.53 (0.41) & 0.41 (0.29) \\ \cline{2-4}
 & RGB Optical & 0.53 (0.49) & 0.44 (0.29) \\ \hline
 \multirow{2}{*}{Neural Network}   & Evetac & 0.33 (0.25) & 0.22 (0.18) \\ \cline{2-4}
 & RGB Optical & 0.35 (0.30) & 0.37 (0.26) \\ \hline
\end{tabular}}
\vspace{-0.3cm}
\end{table}

Due to their sensing principle, event-based cameras naturally only return relative, sparse information.
While this property is beneficial with respect to sensor data rate, it might be a limitation in scenarios where knowledge about the gel's global deformation
is important.
This section, therefore, evaluates the proposed regularized dot tracking algorithm (cf. \sec{sec:dot_tracking}) through the task of shear force reconstruction solely using the dots' displacements $D_\mathrm{C}(t_i)$ as input.
We will compare against the quality of shear force estimation based on tracking the dots using a standard RGB optical tactile sensor (GelSight Mini).

For conducting the experiment, we mount the two tactile sensors on the end effector of a 7DoF Franka Panda robot. 
\fig{fig:force_reconstruct} shows the setup for Evetac.
We designed a similar mount allowing to attach GelSight Mini to the robot in the same pose. 
As shown, the robot presses the respective sensor against a flat, 3D-printed object, which in turn is mounted on top of a \textit{SCHUNK FTCL-50-40} force torque sensor (F/T sensor).
Upon the robot establishing contact between tactile sensor and the 3D printed object, it executes a trajectory of predefined waypoints that result in shearing the gel.
During execution, we track the position of the dots and record the readings of the F/T sensor.
The trajectories cover shear forces in the range of $\pm \SI{10}{\N}$.
The tactile sensors are aligned such that the direction in which they have 9 dots, i.e., the longer side, aligns with the x-axis of the F/T sensor.
For Evetac, we use the previously proposed regularized dot tracker \sec{sec:dot_tracking}.
For the RGB optical tactile sensor, we use the optical-flow-based tracker from the GelSight repo \cite{Shure_GelSight_Robotics_Software_2021}.
We end up with one training dataset per sensor, which consists of 4 trajectories and one separate, previously unseen trajectory for evaluation. 
The trajectories differ in that the sequence of the waypoints is randomized.
The relative translation in x- and y-direction between sensor and 3D printed object is also randomized within $\pm \SI{1}{\cm}$.
Each trajectory contains 200 datapoints.
Each datapoint contains the displacement of all of the 63 dots of the gel ($D_\mathrm{C}(t_i)$) and the shear force readings of the F/T sensor.

To evaluate the dot-tracking quality, we attempt to reconstruct the shear forces based on the displacement of the tracked dots.
For this experiment, we investigate two models.

\textbf{Linear Regression.} Inspired by \cite{yuan2017gelsight}, we fit a linear model, mapping from the overall displacement of the dots in the x and y direction to the shear forces.
We fit the model to the training trajectories via least squares minimization.

\textbf{Neural Network.} The second model is a fully connected neural network that takes the displacement of every individual dot in x- and y-direction as input. Given we have 63 dots, this forms a 126 dimensional input. The network consists of 2 fully connected layers with 128 neurons, each, 25\% dropout, and using the ReLU activation function.
The last layer maps to two outputs, i.e., $F_x$ and $F_y$.
For training, the same trajectories as for the linear model are used.
However, we additionally divide these 4 trajectories in 3, which will be used for training, and 1 for testing. We train the model for 50 epochs and select the model with the lowest loss on the test dataset.

The results when evaluating the trained models on the previously unseen evaluation trajectory are presented in \fig{fig:force_recon} and \tab{table:shear_force}.
As shown in \fig{fig:force_recon}, qualitatively, both of the models are able to reconstruct the shear force throughout the trajectory.
Quantitatively (cf. \tab{table:shear_force}), considering the Linear Regression models, the force reconstruction through the dot tracking from Evetac, or the RGB optical tactile sensor yield similar results.
The fact that the error for the x-direction is slightly higher might be related to the fact that there are more measurements at higher forces for the x-direction.
Using the more powerful neural network model helps to improve the results, reducing the error in estimating $F_x$ by around 37\% for both sensors, and for the y-direction by around 46\% and 16\% for Evetac and the RGB optical tactile sensor, respectively.
Overall, we conclude that reconstructing the shear forces from tracking the dots' locations is possible with good accuracies.
Moreover, the shear force reconstruction is of similar quality for both sensors.
Small variations in the results might be related to the fact that the trajectories for recording the data with the different sensors might be slightly different.
Nevertheless, these results are in line with our expectations, and they confirm that our proposed regularized dot tracking from Evetac's raw output is of similar quality than performing dot tracking from images of the RGB optical tactile sensors.

\section{Slip Detection using Evetac}
\label{sec:slip_detection}

After successful validation of Evetac's basic properties, we focus on a more practically relevant task - slip detection.
Reliable slip detection is a crucial task in robotics as slippage is related with unstable contacts between finger, i.e., sensor, and object.
For achieving stable grasping, any slippage requires quick corrective actions to prevent dropping the grasped object.
Next, we introduce our data-driven, model-free approach for slip detection.
We particularly want to learn the slip detector from data as we want to avoid pre-specifying any slip criteria.
We want the neural network models to automatically learn and focus on the most important information based on labeled training data.
In the following section, we first describe the experimental setup and our procedure for labeled data collection. 
Second, we present the classifier for data labeling.
Third, we provide the training procedure and the neural network architectures for slip detection.
Last, we evaluate the trained models on previously unseen data, considering the training objects and novel objects.
One section later, we also investigate the effectiveness of the slip detection models for stable object grasping, by integrating them into a real-time, reactive feedback control loop.

\subsection{Experimental Setup and Data Collection Procedure}
\label{sec:data_collect}

\newlength{\unittable}
\setlength{\unittable}{6mm}
\begin{table*}[t]
\begin{center}
\caption{\small
Overview of the training and testing objects considered for the slip detection and grasp control experiments. In the materials column, G represents glass, M metal, P plastic, and Pa paper. For cylindric \& sphere objects, only their diameter is provided (as width). \label{table:obj_overview}
} 
\scalebox{0.785}{
    \begin{tabular}{|p{16mm}|p{\unittable}|p{\unittable}|p{\unittable}|p{\unittable}|p{\unittable}|p{\unittable}|p{\unittable}|p{\unittable}|p{\unittable}|p{\unittable}|p{\unittable}|p{\unittable}|p{\unittable}|p{\unittable}|p{\unittable}|p{\unittable}|p{\unittable}|p{\unittable}|p{\unittable}|p{\unittable}|}
        \hline
        & \multicolumn{9}{|c|}{\textbf{Training Objects.}} & \multicolumn{11}{|c|}{\textbf{Testing Objects.}} \\ \cline{2-21}
        & \includegraphics[width=\unittable]{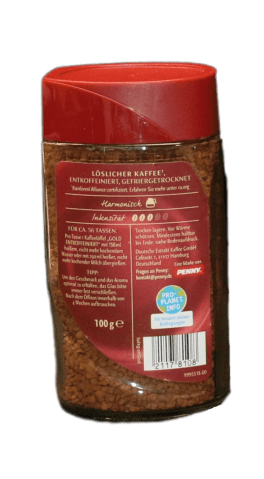} 
        & \includegraphics[width=\unittable]{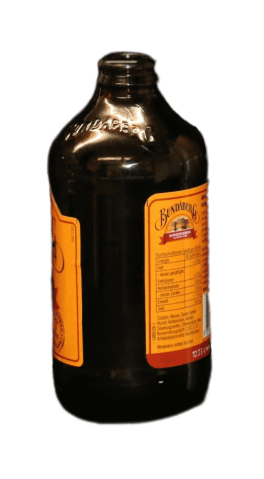}
        & \includegraphics[width=\unittable]{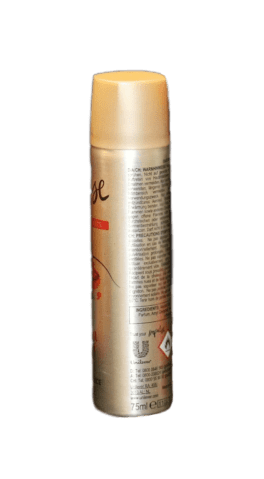}
        & \includegraphics[width=\unittable]{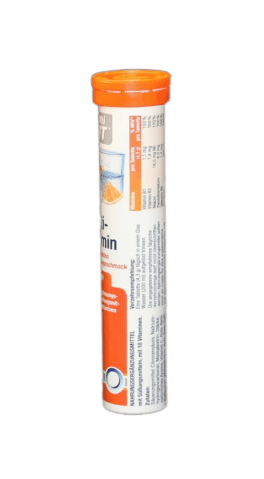}
        & \includegraphics[width=\unittable]{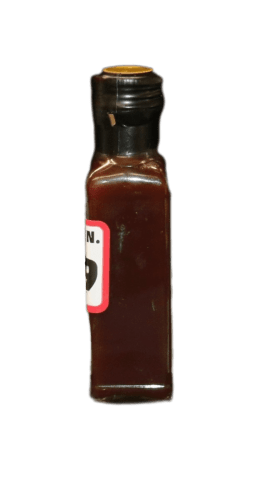}
        & \includegraphics[width=\unittable]{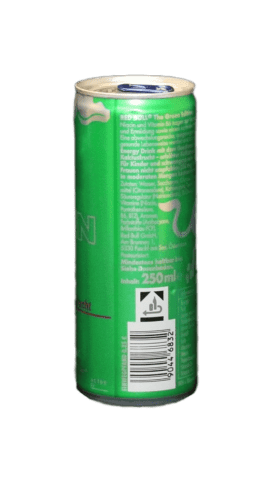}
        & \includegraphics[width=\unittable]{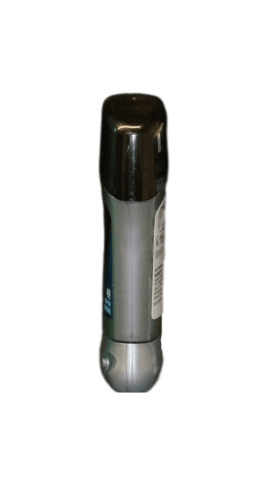}
        & \includegraphics[width=\unittable]{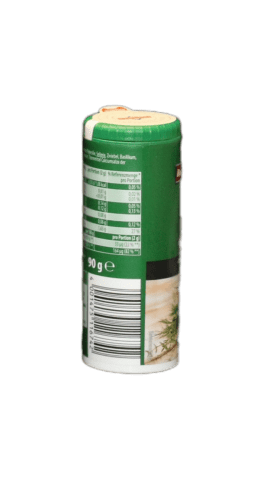}
        & \includegraphics[width=\unittable]{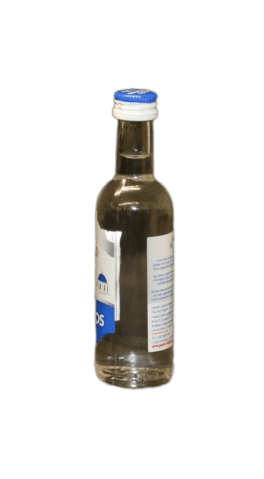}
        & \includegraphics[width=\unittable]{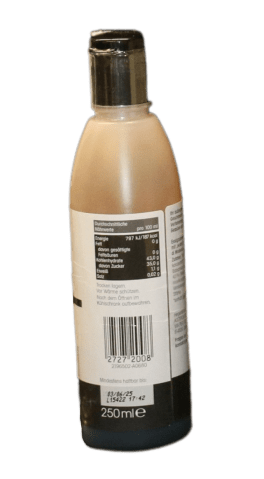}
        & \includegraphics[width=\unittable]{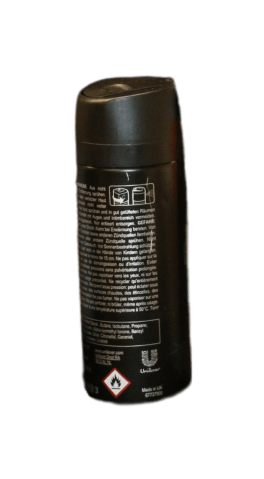}
        & \includegraphics[width=\unittable]{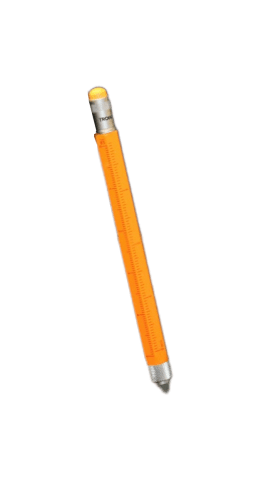}
        & \includegraphics[width=\unittable]{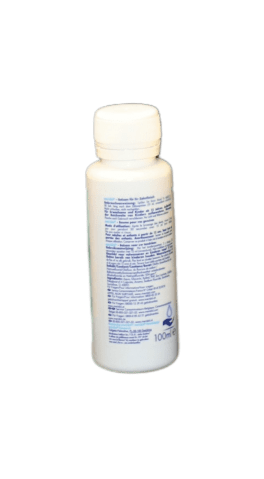}
        & \includegraphics[width=\unittable]{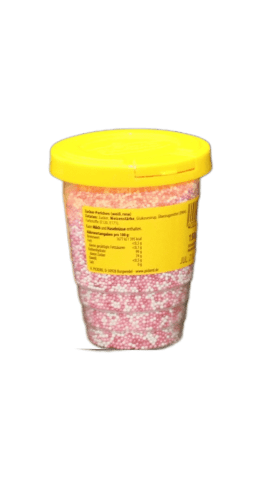}
        & \includegraphics[width=\unittable]{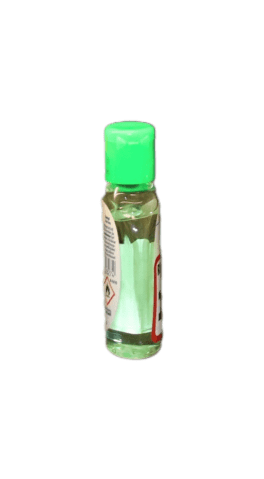}
        & \includegraphics[width=\unittable]{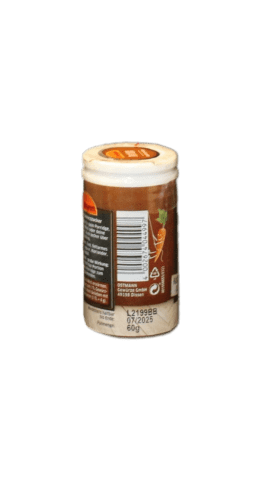}
        & \includegraphics[width=\unittable]{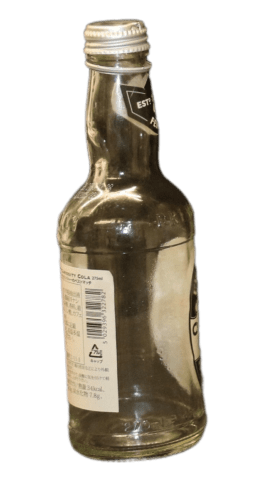}
        & \includegraphics[width=\unittable]{pictures/all_objs/transparent_bottle.png}
        & \includegraphics[width=\unittable]{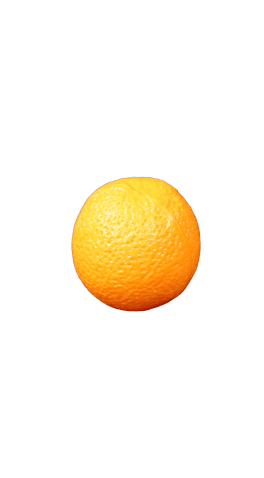}
        & \includegraphics[width=\unittable]{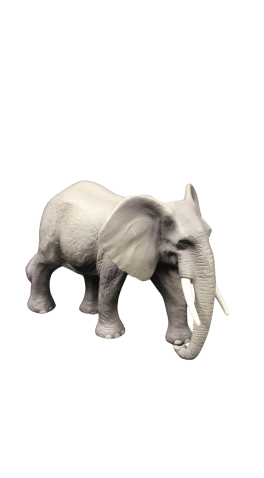}
        \\ \hline
        Obj ID & 1 & 2 & 3 & 4 & 5 & 6 & 7 & 8 & 9 & 10 & 11 & 12 & 13 & 14 & 15 & 16 & 17 & 18 & 19 & 20 \\ \hline
        Mass \SI{}{\g} & 341 & 325 & 70 & 102 & 236 & 273 & 100 & 114 & 119 & 72 & 108 & 38 & 116 & 166 & 56 & 74 & 248 & 550 & 96 & 374 \\ \hline
        Height \SI{}{\mm}$ \updownarrow$ & 145 & 145 & 145 & 145 & 140 & 135 & 115 & 105 & 125 & 185 & 140 & 150 & 118 & 85 & 102 & 80 & 190 & 190 &  & 100 \\ \hline
        Width \SI{}{\mm}$ \leftrightarrow$ & 74 & 70 & 35 & 27 & 36 & 52 & 26 & 41 & 33 & 50 & 48 & 10 & 40 & 71 & 25 & 42 & 60 & 60 & 55 & 151 \\\hline
        Depth \SI{}{\mm} & 61 & & & & 36 & & 63 & & & & & & & & 35 & & & & & 53\\\hline
        Material & G & G & M & P & G & M & P & Pa & G & P & M & M & P & P & P & Pa & G & G & P & P \\\hline
        Comment & &  &  &  &  &  &  &  &  &  &  &  &  &  &  &  & empty & filled & & \\\hline
    \end{tabular}}
\end{center}
\vspace{-0.75cm}
\end{table*}%

\begin{figure}[t]
	\centering
	\def\svgwidth{\columnwidth} %
        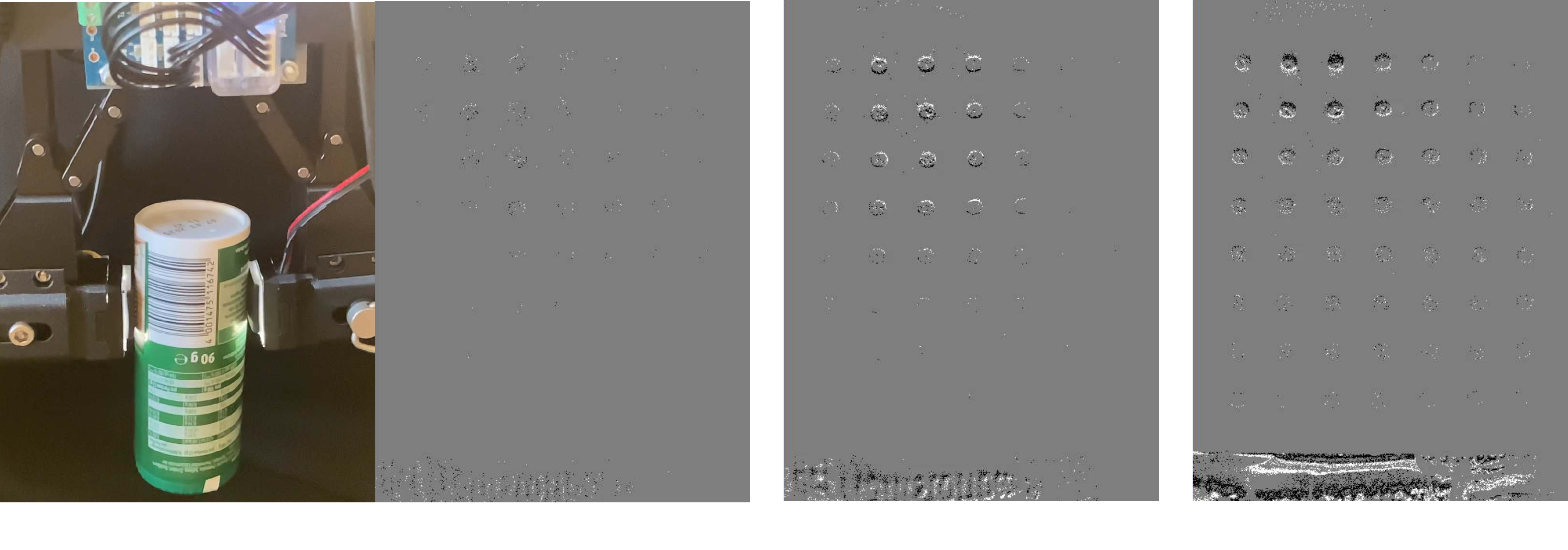 
	\caption{Series of pictures illustrating Evetac's measurements during data collection. Left, we see Object 8 held in the gripper, shortly (\SI{30}{\ms}) before it is going to slip. Upon further opening the gripper, the object starts to slip (middle). Note how, especially in the lower, window region of the gel, we can now clearly see events related to the texture moving. The number of events triggered by the moving texture increases further upon the object accelerating (right).}%
	\label{fig:slip_onset_visual}
 \vspace{-0.5cm}
\end{figure}

\begin{figure}
    \centering
    \includegraphics[width=0.85\columnwidth]{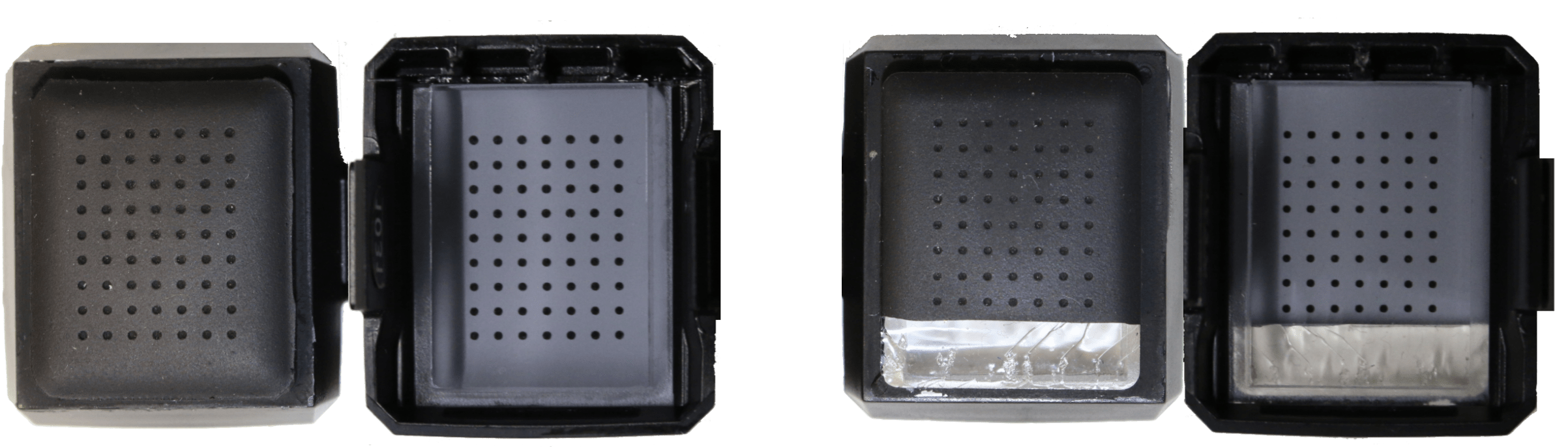}
    \caption{Illustrating the original gel (left) in comparison with the modified cut version (right), providing the view from the outside and the inside. The modified gel enables seeing the object's movement from within the sensor (cf. \figs{fig:slip_onset_visual}). This allows running the slip classifier on the same image, which also contains the tactile measurements and circumvents any additional delays (cf. \sec{sec:slip_classifier_section}). The part of the gel was removed using a simple box cutter, cutting vertically until reaching the plexiglass and subsequently scraping off the small part.}
    \label{fig:cut_gels}
\end{figure}

For labeled data collection, we use the setup shown in~\fig{fig:slip_onset_visual} - left.
It consists of 2 Evetac sensors installed in a ROBOTIS \textrm{RH-P12-RN(A)} gripper, which is mounted in a static configuration.
The two Evetac sensors are synchronized before data collection, and both read out at \SI{1000}{\hertz}.
The gripper is controlled at \SI{500}{\hertz}.
The training objects and their properties are presented in \tab{table:obj_overview}.
As shown in the videos, for collecting the data, the objects will be grasped by the gripper.
Subsequently, we force slip by opening the gripper using current control.
Starting from the currently applied current, when the object is held stably, we adapt the target current every timestep by randomly sampling a value from within $[-0.025, -0.0025]\SI{}{\mA}$, i.e., considering one second, the target current is adapted between $[-12.5, -1.25]\SI{}{\mA}$.
This way, we collect 40 trajectories per object and 360 in total.
We split them into 315 trajectories used to train the model and 45 trajectories that form the test dataset during training.

\subsection{Data Labelling - Slip Classifier}
\label{sec:slip_classifier_section}

One of the most important components for learning a slip detector from labeled data is the classifier for actually labeling the data.
While previous works rely on different sensors for data labeling, such as external cameras tracking Aruco markers \cite{james2020slipmulti}, OptiTrack markers \cite{taunyazov2020event}, or using IMUs attached to the object \cite{su2015force}, herein, we take a different approach.
We aim to recover the moment of slip from the readings of Evetac.
For this purpose, we cut the original gel and removed a small piece using a box cutter as shown in \fig{fig:cut_gels}.
Due to the transparent plexiglass remaining in the cut region, we can essentially see through the sensor at this location.
Now, looking through the transparent part of the sensor, i.e., the region without any gel remaining, we can determine when the object starts moving using an optical flow based criteria.
While this choice comes at the disadvantage of losing part of the contact area of the tactile sensor (roughly 15\%), it comes with the big advantage that the signal that is used to determine the moment of onset of object slippage is inherently aligned with the raw tactile measurements.
Any errors due to synchronization can thus be eliminated, and the classifier naturally provides the same temporal resolution as the tactile readings.
We nevertheless want to point out that we had to run the slip classification procedure offline, i.e., after data collection, thereby not having to meet any real-time requirements.
The slip classifier relies on representing Evetac's raw measurements as images and the objects exhibiting texture in the window region (cf.~\fig{fig:slip_onset_visual}).
We label Evetac's raw measurements as slip or non-slip samples by computing the optical flow between the current measurement and the measurements \SI{4}{\ms} ahead.
We motivate this forward-oriented flow calculation (i.e., calculating the flow of the current measurement w.r.t. a measurement in the future) with the fact that for events to be triggered, the object has to be in motion.
Thus, if we have optical flow between consecutive Evetac measurements, we know that the object changed its position, but also that the motion started already in the initial frame.
Moreover, this choice aims to mitigate any delays in slip classification and is enabled by the fact that we label the measurements offline after data collection.
For computing the optical flow, we use the OpenCV implementation of the Gunnar Farneback \cite{farneback2003two} method.
We compute the flows for both regions of the image separately, i.e., the flow for the region with the markers and the window region.
If the relative flow between the window and the marker region exceeds a certain threshold, then we mark the current measurement as belonging to the slip class.
We consider the relative flow between the two regions since in situations where the flow in the tactile region and the transparent window region are equal, but both nonzero, we still want to obtain a non-slip label, as the elastic gel and object are moving in accordance and not relative to each other.
Thus, there is no slip.
In general, we found this analysis of optical flow more robust compared to a simpler approach that would only analyze the number of events in the respective regions, since optical flow is more invariant to the object's specific texture.
While the previously described procedure for determining the onset of slip requires that the object has texture, we want to point out that textured objects are only needed during data collection as they enable automatic data labeling.
The trained models will only operate on the tactile data without having access to the cut, window region. 
Thus, during deployment, object texture is not necessary.

\subsection{Model Architectures}

\begin{figure*}[t]
	\centering
	\def\svgwidth{\textwidth} %
        {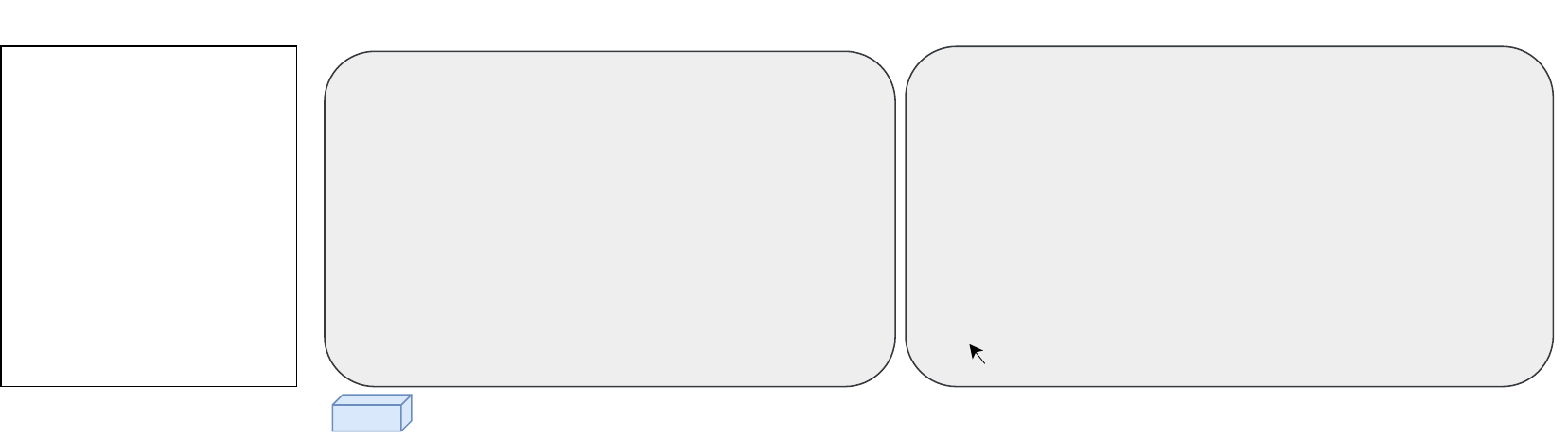} 
	\caption{General neural network architecture used for slip detection. First, per dot, a feature vector containing the current dot displacement $\mathcal{F}_D$ (eventually combined with a sequence of past measurements) is extracted. The same is repeated for the number of events per dot, yielding $\mathcal{F}_E$. Second, the feature vectors are concatenated and embedded using two fully connected layers, sharing weights between the dots. Third, the embedded features are spatially combined, in the same way as the dots are placed relative to each other, and processed through two convolutional layers. Finally, the result is flattened, processed using two last fully connected layers, and yields the slip predictions.}%
	\label{fig:netw_arch}
\hfill
\vspace{-0.35cm}
\centering
\captionof{table}{Parameters of the different model configurations used for slip detection. \figs{fig:netw_arch} shows the general architecture.}
\label{table:model_parameters}
\scalebox{0.725}{
\begin{tabular}[t]{c|c|c|c|c|c|c|c|c|c}
Name  & $\mathcal{F}_D$ & $\mathcal{F}_E$ & $l_i$ & $l_{\textrm{fc1}}$ & $l_{\textrm{fc2}}$ & $h_1, w_1, l_{\textrm{c1}}$ & $h_2, w_2, l_{\textrm{c2}}$ & $l_{\textrm{fc3}}$ & $l_{\textrm{fc4}}$ \\
\hline
\hline
no hist  & $[d_{\vc}(t_i)]$ & $[N_\mathrm{E}(t_i, \vc)]$ & 2 & 10 & 4 & 7,6,16 & 3,3,32 & 32 & 10 \\ \hline
hist 10  & $[d_{\vc}(t_i), d_{\vc}(t_{i-1}), ..., d_{\vc}(t_{i-9})]$ & $[N_\mathrm{E}(t_i, \vc), N_\mathrm{E}(t_{i-1}, \vc), ..., N_\mathrm{E}(t_{i-9}, \vc)]$ & 20 & 12 & 4 & 7,6,16 & 3,3,32 & 32 & 10 \\ \hline
\makecell{events only \\ hist 10} & & $[N_\mathrm{E}(t_i, \vc), N_\mathrm{E}(t_{i-1}, \vc), ..., N_\mathrm{E}(t_{i-9}, \vc)]$ & 10 & 8 & 4 & 7,6,16 & 3,3,32 & 32 & 10 \\ \hline
\makecell{disp only \\ hist 10}  & $[d_{\vc}(t_i), d_{\vc}(t_{i-1}), ..., d_{\vc}(t_{i-9})]$ & & 10 & 8 & 4 & 7,6,16& 3,3,32 & 32 & 10 \\ \hline
hist 20  & $[d_{\vc}(t_i), d_{\vc}(t_{i-1}), ..., d_{\vc}(t_{i-19})]$ & $[N_\mathrm{E}(t_i, \vc), N_\mathrm{E}(t_{i-1}, \vc), ..., N_\mathrm{E}(t_{i-19}, \vc)]$ & 40 & 20 & 8 & 7,6,16 & 3,3,32 & 32 & 10 \\ \hline
\makecell{hist 50 \\ down 5} & \makecell{ $[\sum_{l=0}^{4} 0.2 d_{\vc}(t_{i-l}), \sum_{l=0}^{4} 0.2 d_{\vc}(t_{i-5-l}),$ \\ $... , \sum_{l=0}^{4} 0.2 d_{\vc}(t_{i-45-l})]$} & \makecell{ $[\sum_{l=0}^{4} 0.2 N_\mathrm{E}(t_{i-l}, \vc), \sum_{l=0}^{4} 0.2 N_\mathrm{E}(t_{i-5-l}, \vc),$ \\ $... , \sum_{l=0}^{4} 0.2 N_\mathrm{E}(t_{i-45-l}, \vc)]$} & 20 & 12 & 4 &7,6,16 & 3,3,32 & 32 & 10 \\ \hline
\makecell{fast slow \\ hist 50} & \makecell{ $[d_{\vc}(t_i), d_{\vc}(t_{i-1}), ..., d_{\vc}(t_{i-9}),$ \\ $\sum_{l=0}^{9} 0.1 d_{\vc}(t_{i-l}), \sum_{l=0}^{9} 0.1 d_{\vc}(t_{i-10-l}),$ \\ $... , \sum_{l=0}^{9} 0.1 d_{\vc}(t_{i-40-l})]$} & \makecell{ $[N_\mathrm{E}(t_i, \vc), N_\mathrm{E}(t_{i-1}, \vc), ..., N_\mathrm{E}(t_{i-9}, \vc)$ \\ $\sum_{l=0}^{9} 0.1 N_\mathrm{E}(t_{i-l}, \vc), \sum_{l=0}^{9} 0.1 N_\mathrm{E}(t_{i-10-l}, \vc),$ \\ $... , \sum_{l=0}^{9} 0.1 N_\mathrm{E}(t_{i-40-l}, \vc)]$} & 30 & 15 & 8 & 7,6,16 & 3,3,32 & 32 & 10 \\ \hline
\end{tabular}
}
\vspace{-0.5cm}
\end{figure*}

For training the models for slip detection, similar as in \cite{li2018slip}, we make use of Neural Networks.
Apart from optimally fitting the data, we want to keep the network inference times low, such that we can later evaluate the models online in real time with \SI{1000}{\hertz}.
As features, we will, therefore, mainly focus on two quantities.
First, we make use of the dots' current displacement $d_{\vc_i}$, i.e., their distance to their initial location.
As shown in \sec{sec:shear_force_reconstruction}, this information can be used to reconstruct shear forces acting on the sensor and thus provide information about the global gel configuration.
Second, we consider the number of events per dot $N_\mathrm{E}(t_i, \vc_i)$, i.e., the current number of events triggered in the vicinity of each dot.
From \sec{sec:sensing_vibrations}, we know that the number of events is effective for resolving high frequency phenomena and sensing vibrations.
We want to point out that information from the transparent cut region is not available to the classifiers.
They only have the information from the remaining $7{*}8{=}56$ dots.
While we collected data for both Evetac sensors in the parallel gripper, herein, we focus on slip detection and grasp control using a single sensor only.
We leave slip detection and grasp control using multiple Evetacs for future work.

The general architecture of the slip detection models is depicted in \fig{fig:netw_arch}.
Later sections investigate different configurations of this architecture w.r.t. the input features that are available (cf. \tab{table:model_parameters}).
As input, the models either receive the dots' displacements, the number of events per dot, or both.
In case of both, we first concatenate the dot displacement $\mathcal{F}_D$ and dot event features $\mathcal{F}_E$.
Almost all architectures consider a time series of measurements, i.e., a history of previous measurements in addition to the current measurement.
Given the resulting input vector per dot with dimension $1xl_i$, we first pass it through a two-layered fully connected neural network for encoding.
This procedure is repeated for all of the dots, and we use the same weights for all dots.
Next, we spatially combine these initial per-dot embeddings, in the same way as the dots are placed relative to each other in the gel.
This allows us to subsequently use two convolutional layers, taking the spatial information and topology of the gel into account.
We use convolutions, as slip is a local phenomenon, i.e., it is likely that an entire region of the gel is slipping.
Thus, if an object is slipping, the slip signal should typically be sensed at multiple locations of the gel.
Lastly, we flatten the features and pass them through two fully connected layers to receive the output, which is a scalar between $[0,1]$ and can be interpreted as a slip probability.
To evaluate the effectiveness of our proposed architecture, we compare it against a baseline model architecture, which receives as input the events in image form ($\mathcal{I}(t_i)$). After removing the window region from the measurement, it is processed with 4 convolutional layers with (5,5) kernel size, and 15, 20, 25, and 32 output channels, respectively.
After every layer, we apply max-pooling using kernel sizes of (3,3), (3,3), (4,4), (9,7). This results in a 32-dimensional output that is converted to the slip probability through two fully connected layers of sizes (32,10) and (10,1).  

\subsection{Model Training \& Data Selection}

Apart from the model architecture, careful selection of the training data is also important.
If we consider a single trajectory that is recorded as described in \sec{sec:data_collect}, it first contains many measurements where the object is held stably, then follows object slippage, and slightly after, the object has slipped completely and lost contact with the sensor.
As also discussed in \cite{james2020slipmulti}, it is important to determine, how many measurements to consider after the classifier detected slip for the first time.
This choice is crucial since slightly after the onset of object slippage, it is likely that there is no contact between sensor and object. Thus, it does not make sense to classify such a measurement as belonging to the slip class.
We also want to stress the importance of the models correctly detecting the moment of onset of slip, or even predicting it, as this greatly influences the timing of the corrective actions to prevent and minimize slippage.
Empirically, we found best performance, when cutting the trajectories \SI{15}{\ms} after the onset of object slippage (i.e., after the classifier detected slip for the first time) for all models with a history of 10 or less, cutting after \SI{20}{\ms} for models with histories of 20, and cutting after \SI{50}{\ms} for the remaining architectures.

Cutting the trajectories after few samples of slip, however, also introduces data imbalance.
We have way more datapoints of non-slip data. 
To counteract model bias, we use a modified version of the dataloader.
In particular, we sample from three dataloaders at the same time. One of them solely contains slip data, and the other two non-slip data.
The dataloaders for the non-slip data differ in that one contains measurements where the number of events exceeds a certain threshold ($25$), and the other the remaining non-slip measurements.
The distinction across the non-slip frames attempts to separate measurements in which really nothing is happening, from measurements that do contain some events and information.
The latter measurements are especially important to label correctly, as the event patterns have to be differentiated from the actual slip data.
We train using a batch size of 72.
32 samples are drawn from the slipping frames, 32 from the non-slipping frames above the threshold, and 8 from the non-slipping frames below the threshold.
One episode consists of fully looping through the dataset with the non-slipping frames above the threshold.
To additionally improve the models' robustness, we add data augmentation.
Since all the training data was captured with the objects being grasped from top, with probability 50\%, we rotate the input features within $[90,180,\SI{270}{\degree}]$ to mimic different grasp poses. For the cases of $90$ and $\SI{270}{\degree}$, we crop the measurement and pad it such that it is compatible with the networks' usual input size.

We train each model for 70 epochs using stochastic gradient descent, a learning rate of $0.001$, and the binary cross entropy loss.
We log the models every 10 epochs.

\subsection{Evaluation Procedure \& Metrics}

\tab{table:model_parameters} shows the different model configurations we considered herein.
For model evaluation, we will mainly consider 2 metrics.
First, we investigate the point in time when the model first detects slip $t^m_{s0}$.
This point in time is very important, as in a control task, action needs to be taken whenever slip is detected.
It is thus highly undesirable if this point in time is not aligned well with the object really starting to slip, as determined by the classifier ($t^c_{s0}$).
In the following, we will refer to this as the slip timing criterion, which we define as follows.
If the first instance of slip detected by the models $t^m_{s0}$ is within the interval of $t^c_{s0} {-} \SI{50}{\ms} {\leq} t^m_{s0} {\leq}  t^c_{s0} {+} \SI{20}{\ms}$, then we label this trajectory as one in which the slip timing was identified correctly, i.e., ``slip corr``. 

\begin{figure*}[t]
	\centering
	\def\svgwidth{\textwidth} %
        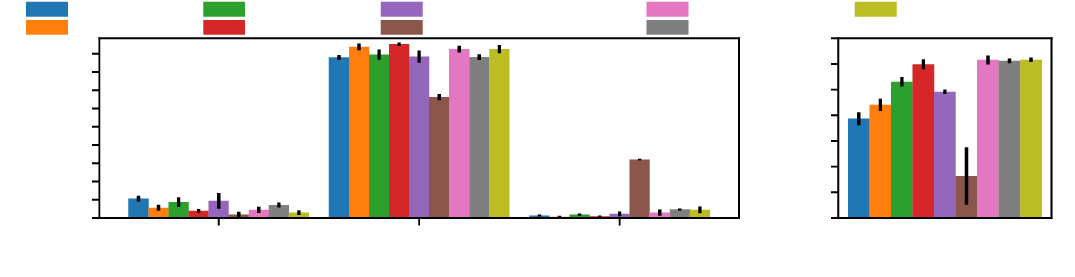 
 \caption{Evaluating different slip detection models on previously unseen trajectories (90 in total) using the training objects. The results show the mean and standard deviation averaged across all objects and five seeds per model configuration. The figure shows performance w.r.t. slip timing criterion (left) and F1 score (right).}%
	\label{fig:slip_off_train}
\vspace{-0.5cm}
\end{figure*}

In other words, the models are allowed to identify slippage at most \SI{50}{\ms} prior to the first instance of slip detected by the classifier, and at latest \SI{20}{\ms} afterwards.
We choose this rather long period before slip occurs, as it might be that the classifier learns to detect features related to incipient slip, which we do not want to label as a wrong detection \footnote{Later experimental results (cf. \sec{sec:slip_timing}) show that across all objects, there are consistent slip detections within $t^c_{s0} {-} \SI{50}{\ms} {\leq} t^m_{s0} {\leq}  t^c_{s0} {+} \SI{20}{\ms}$, while earlier \& later slip detections are clearly separated from this interval. These findings support our empirically chosen interval.}.
We refer to the remaining two cases as ``slip detected too early``, i.e., $t^m_{s0} {<} t^c_{s0} {-} \SI{50}{\ms}$, or as ``too late``, i.e., $t^m_{s0} {>} t^c_{s0} {+} \SI{20}{\ms}$.
Note that the ``too late`` slip detection includes trajectories where no slip is detected at all.
As second metric, we consider the F1 score, which is the geometric mean of recall and precision.
Recall is the ratio of true positives (TP) w.r.t. the sum of true positives and false negatives (FN), i.e., $\textrm{recall} {=} \frac{\textrm{TP}}{\textrm{TP}+\textrm{FN}}$.
Precision is the ratio of true positives w.r.t. the sum of true positives and false positives (FP), i.e., $\textrm{precision} {=} \frac{\textrm{TP}}{\textrm{TP}+\textrm{FP}}$.
Recall gets lower whenever positive samples are erroneously predicted as negative by the models, i.e., slip is not detected.
Precision drops when non-slipping samples are erroneously labeled as slip.
While this metric is related to the previous one, it has a clearer focus on temporal consistency, i.e., only detecting the moment in time of onset of slip correctly, will still result in a bad F1 score, as the recall would be low. 
Thus, the F1 score considers the entire trajectory.
In line with our previous definition regarding slip timing, we cut the trajectories \SI{20}{\ms} after the classifier detected slip for the first time when calculating the F1 scores.

Given that the models' output are continuous and can be interpreted as slip probability, but our metrics require binary labels, we have to convert the models' output into a binary signal.
This is done through thresholding.
The threshold is determined through a grid search from $0$ and $1$ with increments of $0.025$ on the test trajectories using the last three last checkpoints per model. The best combination of model checkpoint and threshold value, w.r.t. maximizing the combination of slip timing criterion and F1 score is selected.

\subsection{Evaluation on Training Objects}

\fig{fig:slip_off_train} shows the results when evaluating the trained models on new, previously unseen trajectories using the training objects.
We recorded 10 trajectories per object.
\fig{fig:slip_off_train} shows the results when evaluating the trained models on new, previously unseen trajectories using the training objects.
We recorded 10 trajectories per object.
We trained 5 models using different seeds per configuration (cf. \tab{table:model_parameters}).
The black bars depict the standard deviation.
As mentioned in \sec{sec:slip_classifier_section}, for all experiments, the slip detection models only have the tactile data from the non-cut gel region available.
They do not receive any information from the cut, window region.

The models without any history (i.e., no hist) already yield quite good results regarding slip timing. The onset of slip is detected correctly (i.e., ``slip corr``) in 87\% of all trajectories.
However, they only achieve an F1 score of 0.59.
The comparison with the baseline model that considers the whole image (i.e., no hist img) reveals that exploiting the full spatial information improves the performance as it achieves a correct slip detection rate of 93\% and an F1 score of 0.64.
Increasing the history length to 10 for our proposed model (hist 10) slightly improves the rate of correct initial slip detections to 89.5\% and the F1 score significantly to 0.73.
For this model configuration with histories of 10, we also show an ablation, comparing models that either only have the number of events (events only) or the dots displacement (disp only) as features available.
The models with access to the number of events per dot outperform the models with access only to the displacements.
Overall, they nevertheless still perform slightly worse than the models that have access to both, especially when considering the F1 score.
We therefore conclude that both features are crucial.
Additionally, we find that the baseline model that operates on Evetac's output in image form (hist 10 img) outperforms the other model configurations, achieving a correct slip detection rate of 95\% and 0.79 F1 score. 
Increasing the history length of our proposed model, i.e., to \SI{20}{\ms} (hist 20), further improves the rate of correct initial slip detection to 92.5\%, the F1 score to 0.815, thereby catching up with the baseline image model.
Using an even longer history, at the cost of downsampling the signal by a factor of 5 (i.e., the down 5 model), which mimics a sensor running at a 5 times reduced frequency, i.e., \SI{200}{\hertz} performs slightly worse in the correct initial slip detection (88.2\%). 
Lastly, we also investigate an architecture, which on the one hand has direct access to the last 10 measurements, i.e., the same input as the hist 10 model, but additionally also access to the last 50 measurements, which are downsampled by a factor of 10. 
We call this model fast slow, as it has direct, unfiltered access to the recent history, as well as knowledge about the signal evolution over a longer horizon. 
Due to the downsampling of the longer history, it overall still has a smaller input size than the model with a history of 20 (cf. \tab{table:model_parameters}). 
These fast slow models perform on par with the hist 20 model, having a correct initial slip detection rate of 92.6\% and an average F1 score of 0.816.

\begin{figure}[t]
	\centering
	\def\svgwidth{\columnwidth} %
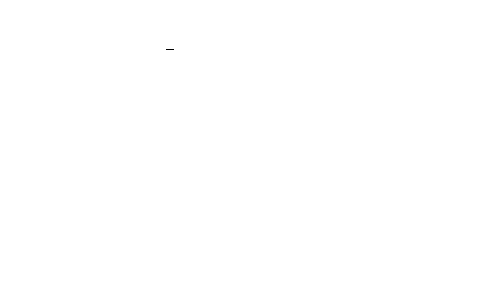
	\caption{Evaluating the different models on the task of slip detection considering a single previously unseen trajectory with Object 3. s corresponds to slip, and ns to no slip.}%
	\label{fig:slip_offline_time_evolution_deo}
 \vspace{-0.5cm}
\end{figure}

These findings are underlined in \fig{fig:slip_offline_time_evolution_deo}, showcasing the models' predictions over time for Object 3.
The model without any history correctly detects the moment of initial slip, however, has difficulties classifying slip states later on, which explains the lower F1 scores.
The model with a history of 10 achieves better temporal consistency, which explains the higher F1 scores.
For this trajectory, the events only model performs similarly, while the model that only has the displacements available exhibits a delayed slip detection, which might be due to a delay between initial events signaling slip and significant dot movement. %
The models with longer histories capture the timely evolution substantially better.
We can also see that the downsampled model (down 5) is slightly delayed, considering initial slip detection. This can be explained by the fact that it can only make predictions every \SI{5}{\ms}, thus, this phenomenon is due to temporal discretization.

\subsection{Evaluating on unseen Testing Objects \& Slip Prediction}

\begin{figure*}[t]
	\centering
	\def\svgwidth{\textwidth} %
        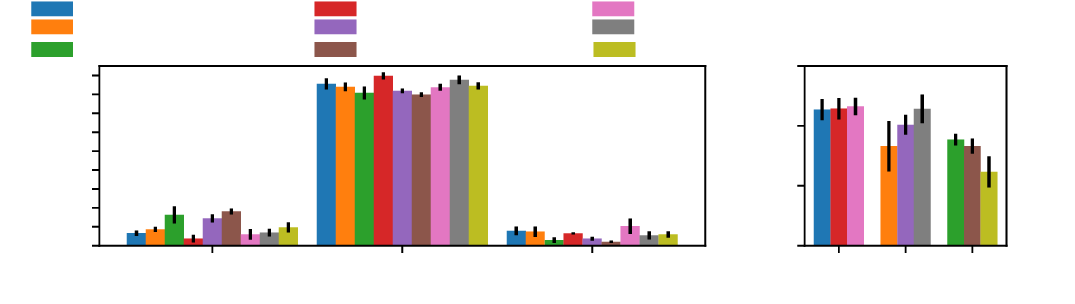 
	\caption{Evaluating three model configurations on the task of slip detection and prediction for $\Delta T_{\textrm{pred}} {=} \SI{10}{\ms}$ or \SI{20}{\ms} in advance using the previously unseen testing objects.
 On the left, we show the slip timing criterion, and on the right, the F1 scores. Mean and standard deviation are reported, averaging across 80 trajectories and five seeds per model configuration.}%
 \label{fig:slip_offline_test_objs}
 \vspace{-0.5cm}
\end{figure*}

Using the previously best performing models, i.e., the models with a history of 20, the fast slow models, and the baseline image model with a history of 10, we next present an evaluation on 8 previously unseen testing objects (Objects 10-17, cf. \tab{table:obj_overview}).
Again, for every object, we record 10 trajectories, yielding 80 trajectories.
We now also investigate the effectiveness of training the models on the task of slip prediction. 
In particular, we train the model configurations on data where we shift the classifier signal by $\Delta T_{\textrm{pred}} {=} \SI{10}{\ms}$ and $\Delta T_{\textrm{pred}} {=} \SI{20}{\ms}$ forward in time.
The respective models are thus tasked to detect slip \SI{10}{\ms} or \SI{20}{\ms} before the classifier detected the onset of slip.
Predicting slip is beneficial, as it increases the time window to react and counteract the slippage.
For consistency, the F1 scores for the slip prediction models are computed using the shifted labels.
The slip prediction models are trained on the same data as the other models, and we again consider five different seeds.%

The results are shown in \fig{fig:slip_offline_test_objs}.
Considering the prediction qualities w.r.t. slip timing criterion, we observe that for the models with a history of 20 and the image model, the percentage of correct initial slip detections (slip corr) continuously decreases (i.e., 86\%, 84\%, 81\%, and 89\%, 82\%, 80\%, respectively) when training the models for the task of slip prediction.
In line with this decrease, the percentage of trajectories where slip is detected too early increases.
Thus, when training the models for slip prediction, it seems that they adjust their features, which, however, also results in detecting slip too early more frequently.
For the fast slow models, the results are slightly different.
Here, the models that have been trained on the task of detecting slip \SI{10}{\ms} in advance yield the highest percentage of correct slip detections (88\%), while the models trained on the original signal and on the \SI{20}{\ms} shifted one perform equally (around 84\%), with the ones tending towards detecting slip too late, while the others again rather detect slip too early, respectively.
It thus seems that the fast slow models which can access the evolution of Evetac's measurements over a longer horizon, however, downsampled, as well as the most recent measurements, are capable of extracting discriminative slip prediction features while avoiding an increase in detecting slip too early for the task of predicting slip \SI{10}{\ms} ahead.
Considering F1 scores, the models for just detecting slip $\Delta T_{\textrm{pred}} {=} \SI{0}{\ms}$ perform comparable. 
For the task of predicting slip \SI{10}{\ms} ahead in time, the fast slow architecture achieves similar scores compared to the baseline slip detection models, while the hist 20 models perform worse on average.
For the task of predicting slip \SI{20}{\ms} in advance, the hist 20 models perform best, however, as analyzed previously, these models come at the cost of significantly lower performance concerning the timing of the first slip detection.

In this experiment, we found a tradeoff between attempting to predict slip, and the models detecting slip way too early.
We nevertheless find that the fast slow architecture offers slightly beneficial performance in that the tendency to detect slip too early is not as prominent while attempting to detect slip \SI{10}{\ms} ahead. 
In fact, this model still performs on par with baseline image model that does not predict slip considering F1 score.
These results hint at the fact that for our datasets \& experimental setup, models that have access to lower dimensional input features but to a longer series of measurements can perform on par with models that have access to the full raw measurements but consider shorter measurements.
There is a tradeoff between the dimensionality of a single measurement and the length of the considered time series w.r.t. computational efficiency.
Computational efficiency is important since the slip detection models should later be used online and integrated into a real-time grasp control loop.
While the proposed fast slow and hist 20 models can be evaluated on average within \SI{0.1}{\ms} on a PC with 128 GB RAM, \textit{NVIDIA RTX 3090}, and \textit{AMD Ryzen 9 5950X 16-Core}, the baseline image model with the history of 10 requires on average \SI{2.6}{\ms}. 
We will thus use the fast slow model architecture with predicting slip \SI{10}{\ms} ahead in the following experiments.

\subsection{Evaluating Slip Timing}
\label{sec:slip_timing}

We analyze the timing of the first slip detection $t^m_{s0}$ of the selected fast slow models predicting slip \SI{10}{\ms} ahead. We compare the timing with the classifier detecting slip for the first time ($t^c_{s0}$) on the previously unseen testing objects.

\begin{figure}
	\centering
	\def\svgwidth{0.95\columnwidth} %
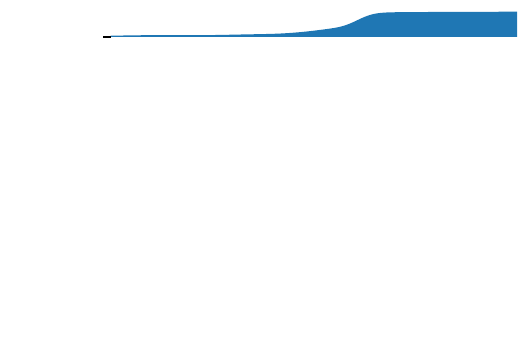
	\caption{Evaluating the timing of the fast slow models' initial slip detection $t^m_{s0}$ w.r.t. the first detection by the classifier ($t^c_{s0}$) on the previously unseen testing objects. We consider five seeds that have been trained for predicting slip \SI{10}{\ms} in advance. We illustrate the slip detections of all seeds across the objects with the black vertical marks on the x-axis. Additionally, we performed a non-parametric density estimation with a Gaussian kernel on the slip timing data and illustrate the resulting cumulative distribution function (CDF).}%
	\label{fig:slip_timing}
\end{figure}

\fig{fig:slip_timing} shows the results, i.e., the timing of the slip detections (black marks), and the CDF after performing a non-parametric density estimation with a Gaussian kernel on the data, considering five seeds.
In agreement with the training configuration, most of the initial slip detections, i.e., 57\%, are within the interval of \SI{10}{\ms} ahead to the actual moment of onset of slip according to the classifier (i.e., within $\left[-10,0\right]$).
The second most frequent interval is between \SI{30}{} to \SI{10}{\ms} ahead in time.
Averaged across all objects, the success percentages of detecting slip correctly, i.e., the interval between \SI{50}{\ms} prior to \SI{20}{\ms} after the classifier detects slip for the first time, is high, with 88\%.
However, for Objects 12 and 15, slip is often predicted too early or too late (26\% and 32\%, respectively).
Potential explanations for these findings could be that Objects 12 and 15 are lighter and smaller compared to the training objects~(cf.~\tab{table:obj_overview}), thereby deforming the gel differently, and eventually resulting in slightly reduced performance. 

\begin{figure}[t]
\vspace{-0.05cm}
\centering
\hfill
\subfloat[\label{fig:rr_initial}]{\includegraphics[width=0.3\columnwidth]{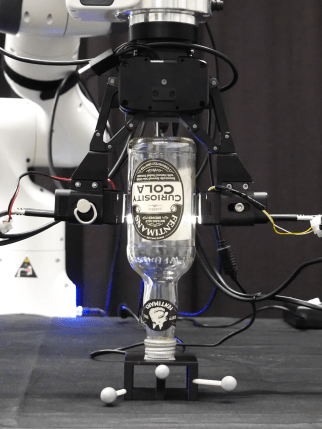}}
\hfill
\subfloat[\label{fig:rr_end_of_lift}]{\includegraphics[width=0.3\columnwidth]{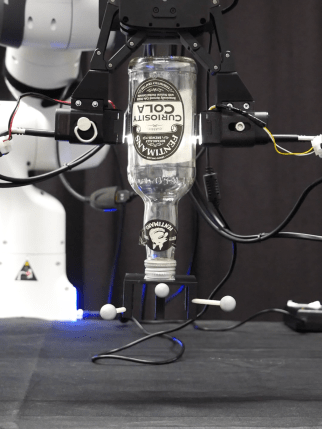}}
\hfill
\subfloat[\label{fig:rr_end_of_balance}]{\includegraphics[width=0.3\columnwidth]{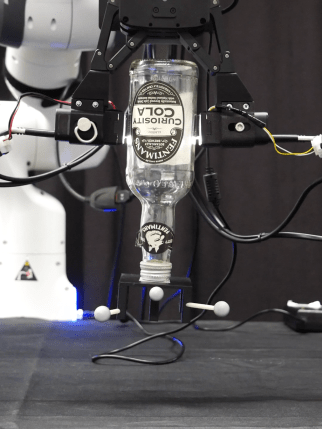}}
\hfill
\vspace{-0.3cm}
\begin{center}
\caption{Closed-loop grasp control experiments. We mount the gripper equipped with Evetacs (cf. \figs{fig:evetac_parallel_grip}) on a Franka Panda 7 DoF robot, for stably grasping and lifting previously unseen objects.
OptiTrack markers are attached to the object to determine how much the object moved relative to the gripper.
\textbf{(a)} depicts the initial situation in which the gripper establishes light contact with Object 17 (cf. \tab{table:obj_overview}), which is insufficient for lifting. Upon the robot starting to lift, object slip is detected by the slip detector, resulting in the closed-loop grasp controller adjusting the grasping force. \textbf{(b)} shows the end of the successful lifting phase. After object lift, we aim to minimize the applied grasping force, by identifying the gripper opening width that is just sufficient for grasping, and at the boundary to the object slipping. This so-called balancing phase typically includes small object slippages, and \textbf{(c)} depicts the object pose at the end of this phase.
The successful completion of this experiment, also demonstrates that our slip detector does not require any object texture, as the object is transparent at the grasping locations.
\label{fig:rr_exp_evolution}
} 
\end{center}
\vspace{-1.cm}
\end{figure}

\section{Closed-Loop Grasp Control using Evetac}
\label{sec:closed_loop_control}

This last experimental section investigates the effectiveness of using Evetac with the previously introduced slip detection and prediction models for reactive robotic grasping.
In particular, we equip a ROBOTIS \textrm{RH-P12-RN(A)} gripper with two Evetacs (cf. \fig{fig:evetac_parallel_grip}) and mount it onto a Franka Panda robot arm, as shown in \fig{fig:rr_exp_evolution}. 
Herein, we only consider the signal of one Evetac and use it in combination with the best model architecture from the previous section, i.e., the fast slow model that has been trained on predicting slip \SI{10}{\ms} ahead.
The slip prediction model is integrated into a real-time grasp control loop.
In the following, we first provide the experimental setup and describe the control strategy, followed by closed-loop pickup and grasping experiments using the previously unseen testing objects.
We end the section by investigating the robustness of the grasp controller w.r.t. changing from top-down to sideways grasps, adding additional disturbances by dropping weights onto the grasped objects, using a different Evetac sensor, and evaluating the pipeline using the original, closed gel.
Again, we provide videos on \href{https://sites.google.com/view/evetac}{our website}.

\subsection{Evaluation Procedure \& Control Strategy}

The experiments deal with the situation in which a robot has to pick up an a priori unknown object.
To solve this task, first, the robot has to position the gripper such that the fingers will make contact with the object upon closing.
Subsequently, sufficient grasping force has to be applied to lift the object stably.
Finally, if having to hold the object for longer, it will be beneficial to apply minimal grasping forces for improved efficiency.
Throughout the whole procedure, the grasping force should adapt to the object that is to be lifted.
The controller should apply less force when dealing with lighter objects, and vice versa.
Moreover, the grasp controller should be reactive w.r.t. any disturbances.
For control, we use the gripper's real-time position control interface.
Given the gripper's current opening width $x_\textrm{g}$, the desired reference opening width $x_{\textrm{ref}}$, and feedforward term $u_{\textrm{ff}}$, together with the control gain $K_p=50$, the control law yields $u_c = K_p (x_\textrm{e}+u_{\textrm{ff}})$ with $x_\textrm{e} = x_{\textrm{ref}} - x_g$.
The reason for using the position control interface is that it offers a higher resolution and, therefore, finer control compared to the current control interface.

To investigate the effectiveness of our trained models for online grasp control in the previously described scenario, we propose the following procedure.
We first move the robot to a suitable pre-grasp pose, assuming knowledge about the object's pose.
Next, we close the gripper to make light contact with the object to avoid any damage.
This is achieved by setting the feedforward signal to a small constant value $u_\textrm{ff}= -2$, which is just sufficient to make the gripper move, while keeping $x_{\textrm{e}}=0$.
The chosen, small control signal will make the fingers stop upon making very light contact.
When attempting to lift the object starting from this initial light grasp configuration, the object would just slip and remain on the table surface, as the applied forces are not sufficient.

\textbf{Lifting Phase.} To counteract this slippage during the lift, we use our slip detector in an online fashion, essentially controlling the width of the gripper.
Upon detecting slippage, we attempt to further close the gripper until the slip stops.
In particular, we now have a time-dependent feedforward term $u_{\textrm{ff}}(t) {=} u_{\textrm{ff}}(t{-}1) {+} u_i(t)$, with the increment $u_i(t){=}-1$, if slip was detected ($s(t){=}1$) between the last and the current call to the controller, and $u_i(t){=}0.01$, if no slip was detected ($s(t){=}0$).
$u_{\textrm{ff}}$ is initialized as zero.
The reason for the different increments is that whenever slip has been detected, we want to react fast, while we want to reduce the grasping force gently in case of no slip, as the reduction of grasping force might lead to new slippage.
Moreover, to limit the amount of force the gripper can exert onto the object, we clip $u_{\textrm{ff}}(t)$ to stay within $[-5,0]$.
Additionally, in the switching moments from slip to no slip (i.e. $s(t{-}1){=}1$ and $s(t){=}0$), we set the reference gripper position to the current gripper opening width $x_{\textrm{ref}}{=}x_g(t)$. 
This is the first gripper closing width realizing a stable grasp without slippage, and should thus be the desirable setpoint.
Before the first occurrence of this switching moment, we leave $x_e$ zero.
Note that this control law is reactive, i.e., if there are multiple occurrences where a switch between slip and no slip happens, the reference is adapted accordingly. Yet, the grasping force can only increase.
Since the slip detector is running at \SI{1000}{\hertz}, while the control loop is operating at half the frequency, we choose a shared memory to pass information between the slip detector and control loop asynchronously.
We create one integer variable that is incremented whenever slip occurs. In the controller, we access this variable and check whether the slip count increased (slip), or is unchanged (no slip) w.r.t. the previous function call.

\textbf{Balancing Phase.} After successful object lift, we aim to minimize grasping force while holding the object. 
We thus attempt to open the gripper, i.e. decrease the grasping force until slip is detected, and then counteract it, similar as in the previous maneuver. However, since now, gravity is also acting on the object, we use a bigger increment to counteract slippage.
Using the same time-varying feedforward term, the gripper is now opened through $u_\textrm{ff}(t){=}0.01$ as long as no slip is detected, and closed through $u_\textrm{ff}(t){=}-2$ upon detecting slip.
Moreover, the first moment of detecting slip (i.e., $s(t{-}1){=}0$ and $s(t){=}1$), activates the reference position of minimal force for holding the object ($x_{\textrm{ref}}{=}x_g(t{-}1)$). Before, $x_e$ has been set to zero.
Additionally, as initially, we want to open the gripper, we clip $u_{\textrm{ff}}(t)$ within $[-5,2]$. Upon first slip detection, we adapt it to $[-5,0]$, as we do not want to open the gripper beyond the reference position, which is the last stable position.

For evaluating our closed-loop grasp controller, we employ the following metrics.
First, we consider the success rates of the individual phases.
Success lift represents successful lifting of the object by around \SI{10}{\cm}, i.e., lifting the object without it slipping completely through the fingers.
Successful balance means successful completion of the second phase, i.e., opening the gripper until slippage is detected, catching the object, and subsequently holding the object applying as little force as possible. This phase is successful, if it still ends with the object in between the fingers.
The two phases take \SI{10}{} and \SI{20}{\s}, respectively.
Note that the balance success rate only counts the trials for which the lifting was successful.
Overall success is the amount of trajectories for which both phases were successful.
As shown in \fig{fig:rr_exp_evolution}, we also attached Optitrack markers to the objects to measure by how much the object moved relative to the gripper throughout the entire maneuver.
Lastly, we report the change in gripper opening width, which provides insight on the grasping strength applied to the objects.
As we do not have any force sensing capabilities available, similar as in \cite{james2020slipmulti}, we assume the grasp strength to be proportional to the gripper closing width.

\begin{figure*}[t]
	\centering
	\def\svgwidth{0.9\textwidth} %
        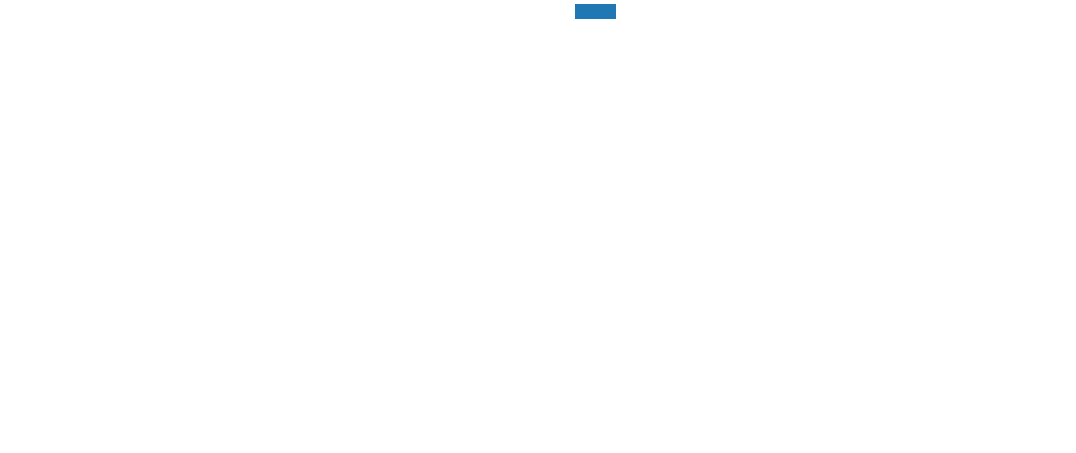 
	\caption{Closed-loop grasp control experiments using the previously unseen testing objects (cf. \tab{table:obj_overview}). For every object, we perform 10 trials. While the top row shows the success rates, the middle row shows the change in gripper opening applied by the controller for lifting the object and optimizing the grasp force, i.e., opening width, during the subsequent balancing phase. The bottom row shows how much the objects moved relative to the gripper during the whole procedure of pickup, lift, and balance. For an exemplary trajectory, see \figs{fig:rr_exp_evolution}.}%
	\label{fig:slip_catch}
 \vspace{-0.5cm}
\end{figure*}

\subsection{Evaluating the Closed-Loop Pickup and Grasp Controller}
\label{sec:eval_cl}

\fig{fig:slip_catch} shows the experimental results for the testing objects, i.e., the objects that have not been seen during model training (cf. \tab{table:obj_overview}).
We conduct 10 trials per object, use the same fast slow slip prediction model, predicting slip \SI{10}{\ms} ahead in time, and control strategy.
To ensure that the initial gripper closing for establishing object contact does not erroneously result in already establishing sufficient forces for successful lifting, for all objects, we also did 5 repetitions in closing the gripper, however, then attempting to lift the object without any further control.
This resulted in $0\%$ lifting successes. 
Therefore, slip detection and appropriate closed-loop control are crucial for successful completion.
As shown in the first row of \fig{fig:slip_catch}, across all objects, in total, we have a success rate of $93\%$.
Only for Objects 12 \& 15 we have one failure during lifting, and for Object 14, in the nominal configuration, we have 4 failures during balance control.
Regarding the two lifting failures, the initial contact forces might have been too light, and thus slip has not been detected.
Moreover, they occur with the two lightest objects that are lighter than all of the training objects (cf. \tab{table:obj_overview}).
These failed lifts are also in line with the results from the offline slip detection \& prediction experiments, where for both objects (12 \& 15), we had more than 10\% probability of not detecting slip at all.
For the cone-shaped Object 14, the 4 balancing failures might be due to the fact that we had to grasp the object upside down, which makes the task of stably grasping and balancing more difficult once the object has slipped and accelerated. 
As can be seen, we repeated the experiment for Object 14 and doubled the grasp control gain ($K_p$) during the balancing phase. 
This resulted in a reduced number of balancing failures and hints that the failures might not only be due to the slip detection but the interplay with control is also crucial.
Since the non-cylindrical Object 14 demanded faster reactions upon slippage, for Objects 19 \& 20, we also employed the higher gains, which resulted in successful execution without failure. This underlines that our pipeline can also handle more irregular contact configurations between sensor and object.
Yet, we believe that, in general, one should try to keep the control signals rather small to avoid exerting excessive forces, which is especially problematic when dealing with more delicate objects.
For all the other six objects, we did not observe any failures.
Looking at the second row, which shows the change in gripper opening width, we can actually see that the grasping force during lifting and the overall maneuver is really adaptive w.r.t. the object that is grasped.
Since object properties such as their surface might also play a role in the required grasping forces, the figure's first two columns provide a good comparison as they compare the grasping efforts for the same object (i.e., a bottle), that is once filled (Obj 18) and once empty (Obj 17).
We can clearly observe that for the heavier, filled object, the closed-loop grasping control pipeline applies more grasping strength, during lifting and also balance.
Comparing the filled and unfilled object, the overall change in gripper opening width for grasping and stabilizing is $-13.4$ and $-7.6$, on average, respectively.
This underlines the adaptiveness of the proposed grasp controller. 
The filled bottle is also roughly twice as heavy as the empty one.
Only for Object 20, even higher grasping forces are needed.
The reason for this is that we grasp the object at a significantly increased horizontal distance w.r.t. its center of mass. Therefore, we not only have to counteract the object's weight, but also prevent it from rotating, which explains the required increased grasping forces.
The videos underline that this object indeed shears Evetac's gel the most.
The figure also shows that the balancing phase is effective in that the grasping strength can be reduced across all the objects, on average by 43\%.
Lastly, when investigating the distance that the object traveled, we see that during lifing, in almost all attempts, the objects only move by a couple of millimeters.
Considering the overall maneuver, for most experiments, the objects move less than \SI{1}{\cm}.
As also shown in the videos, some objects move more than once in the balancing phase. This illustrates our controller's reactiveness and underlines that the gripper opening width has to be carefully chosen.
It can occur that the first desired setpoint is still too open, which might potentially be related with the finger's velocity during the opening movement.
The slip detector and grasp controller are run jointly, and in real-time on a PC with 128 GB RAM, \textit{NVIDIA RTX 3090}, and \textit{AMD Ryzen 9 5950X 16-Core}.
All components from reading the sensor, tracking the dots, and evaluating the neural network slip detector (mean inference time of \SI{100}{\SIUnitSymbolMicro s}) were run at \SI{1000}{\hertz}.

\begin{figure*}[t]
\centering
\hfill
\subfloat[\label{fig:slip_catch_ood}]{\centering
	\def\svgwidth{0.59\columnwidth} %
                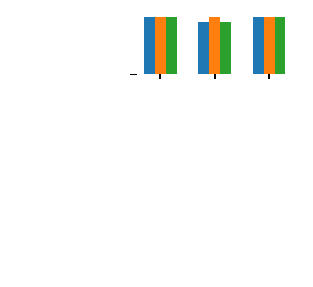 }
\hfill
\subfloat[\label{fig:slip_catch_diff_sensor}]{\centering
	\def\svgwidth{1.39\columnwidth} %
                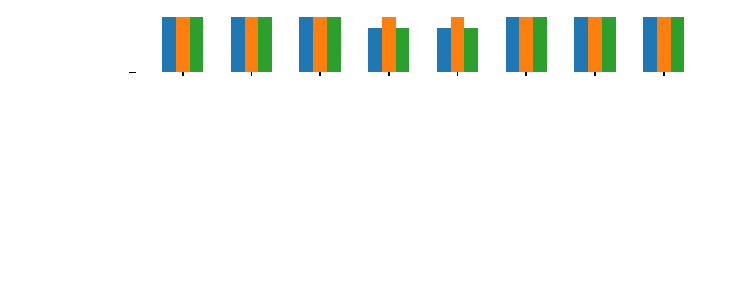 }
\hfill
\vspace{-0.4cm}
\begin{center}
\caption{Grasp controller robustness evaluation. \textbf{(a)} Contrary to data collection and the previous experiments, Object 16 is now grasped sideways. The left column presents the results for the standard maneuver as done in \figs{fig:rr_exp_evolution}. The other two columns show the results when perturbing the grasp during the balancing phase by dropping weights with \SI{20}{} and \SI{100}{\g} onto the object. We perform 10 repetitions for each configuration.
\textbf{(b)} Contrary to the previous experiments, we now use a different Evetac sensor. Additionally, we also investigate the performance when we equip the sensor with the original closed gel (cf.~\fig{fig:cut_gels} - left). We conduct 5 repetitions for each configuration.
As metrics, success rate and the distance that the object moved relative to the gripper are reported.
\label{fig:unlabelled}
} 
\end{center}
\vspace{-1.cm}
\end{figure*}

\subsection{Evaluating Controller Robustness w.r.t. Grasp Orientation \& External Disturbances}

We evaluate the model in scenarios in which objects might be grasped differently, i.e., from the side.
Such data is not included in the raw training data.
It is only covered through data augmentation.
Sideways grasps result in a different orientation of Evetac w.r.t. the object.
We also investigate the reactiveness of the proposed controller by dropping weights of either \SI{20}{\gram} or \SI{100}{\gram} onto the object, after successful initial stabilization in the balancing phase.
As dropping \SI{100}{\gram} onto the object is a substantial perturbation, in this configuration, we again double the control gain ($K_p$) to maintain high success rates. 
We perform 10 trials per configuration.

As shown in \fig{fig:slip_catch_ood}, grasping the object sideways still results in ten out of ten successful trials.
Also, the distance that the object moves is only slightly increased and still comparable to the previous experiment.
Considering the scenarios where additional weights are dropped onto the object upon grasp stabilization, we see that the distances that the object moves increase.
The results also show that the closed-loop control pipeline is reactive as the balancing success rates remain high at $90\%$ \& $100\%$. Only in one trial when dropping \SI{20}{\gram} there is a balancing failure.
The experiments with dropping the \SI{20}{\gram} weight object also indicate that the grip controller does not apply excessive forces.
The addition of \SI{20}{\gram} is already sufficient to destabilize the grasp and make the object move inside the gripper, as can also be seen in the videos.

\subsection{Evaluating Controller Robustness w.r.t. a different Evetac Sensor \& the closed Gel}

These experiments evaluate whether the proposed pipeline can transfer to a different, newly assembled Evetac sensor, and to using the different sensor and the closed, original gel~(cf.~\fig{fig:cut_gels} - left).
For this, we consider 4 of the previously unseen testing objects (Objects 16-20) and evaluate them as done in \sec{sec:eval_cl}, i.e., grasping from the top. We perform 5 repetitions per configuration.

The results in \fig{fig:slip_catch_diff_sensor} reveal that the proposed pipeline is indeed capable of transferring to a different Evetac sensor, as the success rate remains high at 95\%.
Out of the 40 trials, we only observe 2 failures.
Moreover, despite the different sensor, the distances that the objects move during the experiments remain comparable to the nominal configuration (cf.~\sec{sec:eval_cl}).
Most importantly, the table and the videos demonstrate that, as expected, the proposed pipeline of slip detector and control can handle the original, i.e., closed version of the gel.
This underlines that the slip detectors are independent of the existence of the window region and only pay attention to the tactile features in the closed gel region.
Thereby, we confirm our design choice of using the window region for data labeling, however, training the models solely using the tactile data from the remaining uncut gel region.

\section{Discussion \& Limitations}

The experiments investigated Evetac's properties, demonstrated different models for offline slip detection and prediction, and Evetac's effectiveness in grasping a wide range of previously unseen objects with different surfaces, materials, and weights.
While Evetac -- in combination with the corresponding touch processing and control algorithms -- yields good performance across all the tasks and satisfies the desiderata of high-frequency sensing, processing, and control, we also discovered some limitations.
In the grasp control experiments, our approach had the most failures with the cone-shaped Object 14.
We hypothesize that one reason for these difficulties is that Evetac's gel surface is planar and cannot adapt well to the geometry of the grasped object.
One way to overcome this limitation might be to employ a different gel with more curvature, as used in the BioTacs~\cite{fishel2008robust} or TacTip~\cite{ward2018tactip}.
Another limitation are Evetac's dimensions.
While Evetac is the first event-based optical tactile sensor for which all design files are openly available, its current size is unsuitable for integration with dexterous robotic hands (cf. \tab{table:hw_comparison}).
We hope that soon smaller event-based cameras will be released, as the camera's size is currently the major limiting factor.
Alternatively, fiber optic bundles could be exploited for mitigating Evetac's spatial requirements at the sensing location \cite{di2024using}.
Moreover, the price of Evetac's event-based camera results in a substantially increased cost compared with RGB optical tactile sensors.
Regarding the slip detection and prediction experiments, it would be interesting to consider additional network types, such as spiking neural networks, in the future.
They hold the potential to further decrease the overall pipeline's latency as they directly process the sensor's asynchronous output.
Lastly, the slip timing experiments indicated that it is object-dependent.
Future work should try to identify the causes of these findings. They could provide essential information regarding feature selection or sensor material choice.

\begin{table}[t]
\begin{center}
\caption{Comparing Evetac with other optical tactile sensors. Due to the asynchronous sensing principle of event-based (EB) cameras, we only report a value for the event-based sensor's frequency $f$, if the respective sensor is read out using a fixed time interval as done in this work. ( $^{*}$ indicates cost when manufacturing 1000 pieces).}
\label{table:hw_comparison}
\scriptsize
\scalebox{0.85}{
\begin{tabular}{c|c|c|c|c|c}
Sensor & Size [\SI{}{\mm}] & \makecell{Camera\\Type} & Resolution [px] & f [\SI{}{\hertz}] & Cost  \\
\hline
\hline
DIGIT \cite{lambeta2020digit} & 22x27x18 & RGB & 640x480 & 60 & 15\$$^{*}$ \\
\hline
GelSight Mini \cite{gelsightGelSightMiniStore} & 31x28x28 & RGB & 3840x2160 & 25 & 500\$ \\
\hline
TacTip \cite{ward2018tactip} & 48dia. x 55 & RGB & 1920x1080 & 120 & $<100$ \pounds \\  %
\hline
\makecell{Event-based \\ Sensor \cite{kumagai2019event}} & 80x80x125 & EB & 128x128 & 2000 & NA\\
\hline
NeuroTac \cite{ward2020neurotac} & NA & EB & 240x180 & NA & NA \\
\hline
\makecell{Miniaturised \\ NeuroTac \cite{ward2020miniaturised}} & 20x25x30 & EB & 128x128 & NA & NA \\
\hline
\makecell{Event-camera based \\ finger prototype \cite{muthusamy2020neuromorphic}} & NA & EB & 240x180 & 2000 & NA \\
\hline
\textbf{Evetac} & 32x33x65 & EB & 640x480 & 1000 & $\approx 3000$ \texteuro \\
\hline
\end{tabular}
}
\end{center}
\vspace{-0.5cm}
\end{table}

\section{Conclusion} 
\label{sec:conclusion}

This work introduced a new event-based optical tactile sensor called Evetac.
The sensor design aims to maximize re-use of existing components and solely requires 3D printing of a housing that connects together the event-based camera, the soft silicone gel, and the lighting.
In addition to the sensor design, this work also presented the necessary software to read out the sensor in real-time at \SI{1000}{\hertz}, as well as suitable touch processing algorithms running at the same frequency.
In particular, we devised a novel algorithm for tracking the dots imprinted in the gel and proposed a set of tactile features that were exploited for learning efficient neural network-based slip detectors from collected data.
The thorough experimental section first demonstrated the sensor's natural properties of being able to detect tactile vibrations of up to \SI{498}{\hertz}, providing significantly reduced data rates compared to RGB optical tactile sensors despite Evetac's high temporal resolution, and exploiting the dot displacements for shear force reconstruction.
To showcase Evetac's practical relevance, we also evaluated and compared different neural network architectures for the task of slip detection and prediction from Evetac's data on a wide range of objects.
The models formed the basis for designing an effective real-time grasp control loop, achieving high success rates of 93\%, robustness to perturbations, and adaptiveness w.r.t. object mass.
We were also able to showcase the transfer of the models to a different Evetac sensor.
We hope that our proposed open-source Evetac sensor, together with the touch processing algorithms and closed-loop grasping controller, will encourage further research in the field of event-based tactile sensing, and contribute to the efforts for achieving the still unparalleled human manipulation capabilities.%

\section*{Acknowledgments}

\small{
We thank Luca Dziarski for helping with the force reconstruction experiments.
This research received funding from the European Union’s Horizon Europe programme under Grant Agreement No. 101135959 (project ARISE) and from the AICO grant by the Nexplore/Hochtief Collaboration with TU Darmstadt.
It was also supported by the German Research Foundation (DFG, Deutsche Forschungsgemeinschaft) as part of Germany’s Excellence Strategy – EXC 2050/1 – Project ID 390696704 – Cluster of Excellence “Centre for Tactile Internet with Human-in-the-Loop” (CeTI) of Technische Universität Dresden, by Bundesministerium für Bildung und Forschung (BMBF), German Academic Exchange Service (DAAD) in project 57616814 (\href{https://secai.org/}{SECAI}, \href{https://secai.org/}{School of Embedded and Composite AI}),
and the German Research Foundation Emmy Noether Programme (CH 2676/1-1).
Calculations were conducted on the Lichtenberg high performance computer of TU Darmstadt.
}

\bibliographystyle{IEEEtran}
\bibliography{references}

\section*{Biography Section}

\begin{IEEEbiography}
[{\includegraphics[width=1in,height=1.25in,clip,keepaspectratio]{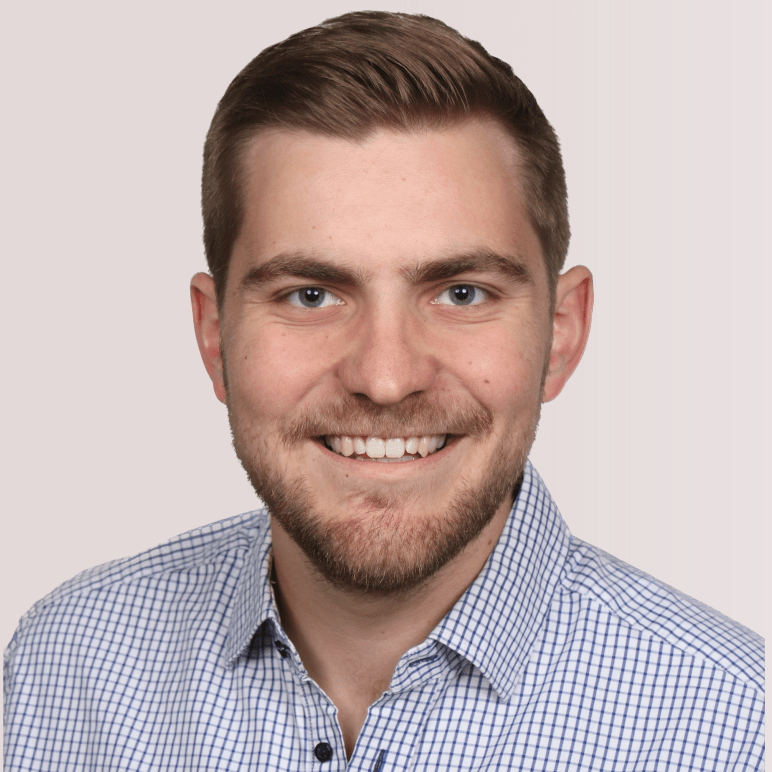}}]{Niklas Funk} received the B.Sc. in Electrical Engineering and Information Technology and M.Sc. in Robotics, Systems \& Control both from ETH Zurich, Zurich, Switzerland.
Currently, he is working toward the Ph.D. degree at the Intelligent Autonomous Systems Group at the Technical University of Darmstadt, Darmstadt, Germany.
His research focuses on the intersection between machine learning \& robotics (Robot Learning), with a particular focus on robotic manipulation, including tactile sensing, grasp \& motion planning, and long-horizon reasoning.  
\end{IEEEbiography}

\begin{IEEEbiography}
[{\includegraphics[width=1in,height=1.25in,clip,keepaspectratio]{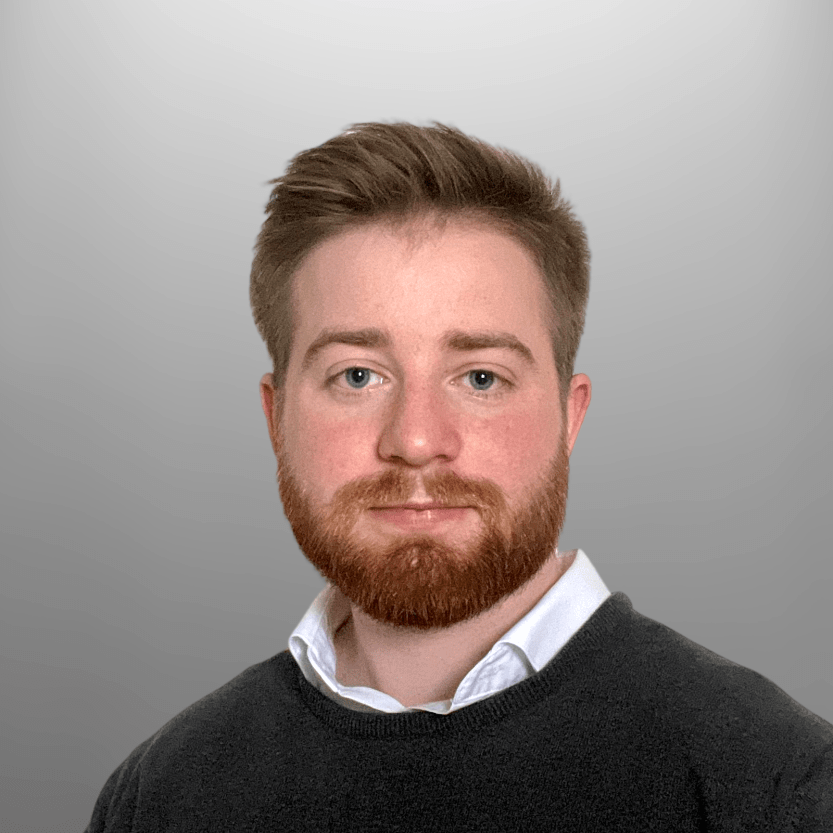}}]{Erik Helmut} received the B.Eng. degree in mechanical engineering from DHBW Mosbach, Mosbach, Germany, in 2021, with a focus on virtual engineering. He is currently working toward an M.Sc. degree in computational engineering at TU Darmstadt, Darmstadt, Germany. His main research interests include tactile sensing and robotics.
\end{IEEEbiography}

\begin{IEEEbiography}
[{\includegraphics[width=1in,height=1.25in,clip,keepaspectratio]{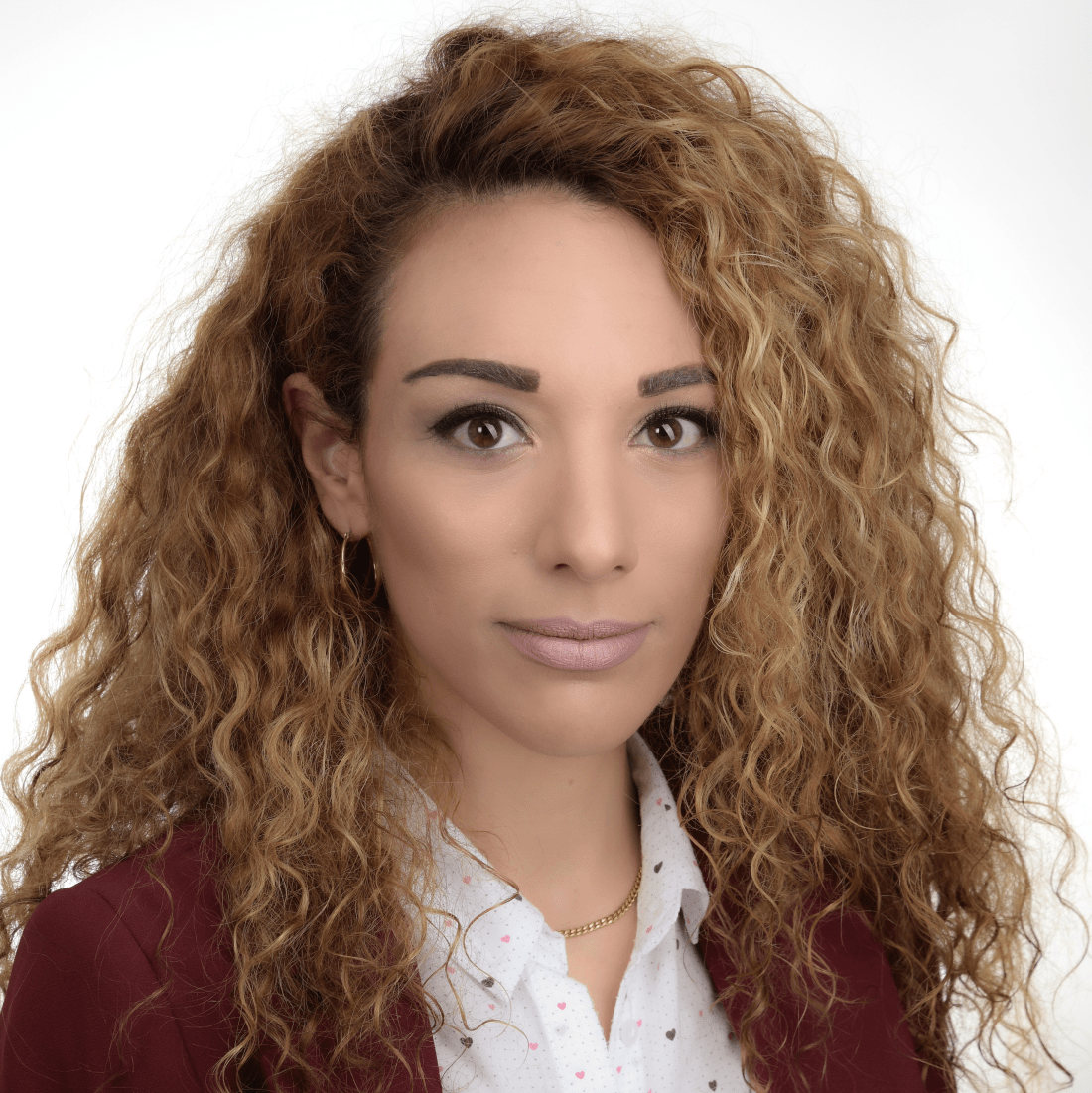}}]{Georgia Chalvatzaki} is a Full Professor (W3) for Interactive Robot Perception and Learning (PEARL) at the Computer Science Department of the Technical University of Darmstadt, Germany. She was awarded an AI Emmy Noether DFG research grant in 2021 and received several awards (IROS 2022 Best Paper Award in Mobile Manipulation,  Outstanding Associate Editor RA-L 2023, Daimler and Benz Foundation Scholarship 2022, 2021 AI Newcomer German Informatics Society, Robotics Science and Systems Pioneer 2020, etc.). She completed her Ph.D. in 2019 at the Electrical and Computer Engineering School of the National Technical University of Athens, Greece.
\end{IEEEbiography}

\begin{IEEEbiography}[{\includegraphics[width=1in,height=1.25in,clip,keepaspectratio]{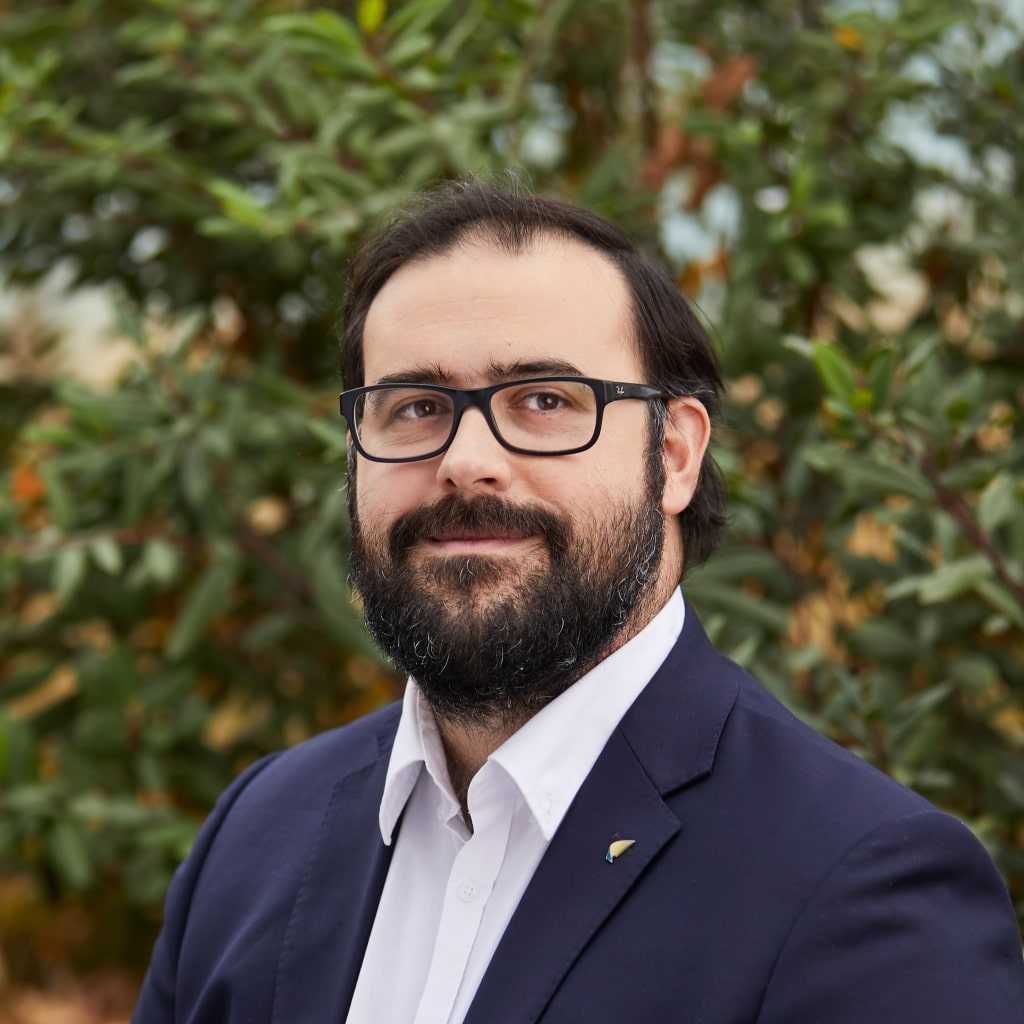}}]{Roberto Calandra}
is a Full (W3) Professor at the Technische Universit\"at Dresden. Previously, he founded at Meta AI (formerly Facebook AI Research) the Robotic Lab in Menlo Park. Prior to that, he was a Postdoctoral Scholar at the University of California, Berkeley (US). His education includes a Ph.D. from TU Darmstadt (Germany), a M.Sc. in Machine Learning and Data Mining from the Aalto university (Finland), and a B.Sc. in Computer Science from the Università degli studi di Palermo (Italy). His scientific interests are broadly at the conjunction of Robotics, Touch Sensing, and Machine Learning. In 2024, he received the IEEE Early Academic Career Award in Robotics and Automation.
\end{IEEEbiography}

\begin{IEEEbiography}
[{\includegraphics[width=1in,height=1.25in,clip,keepaspectratio]{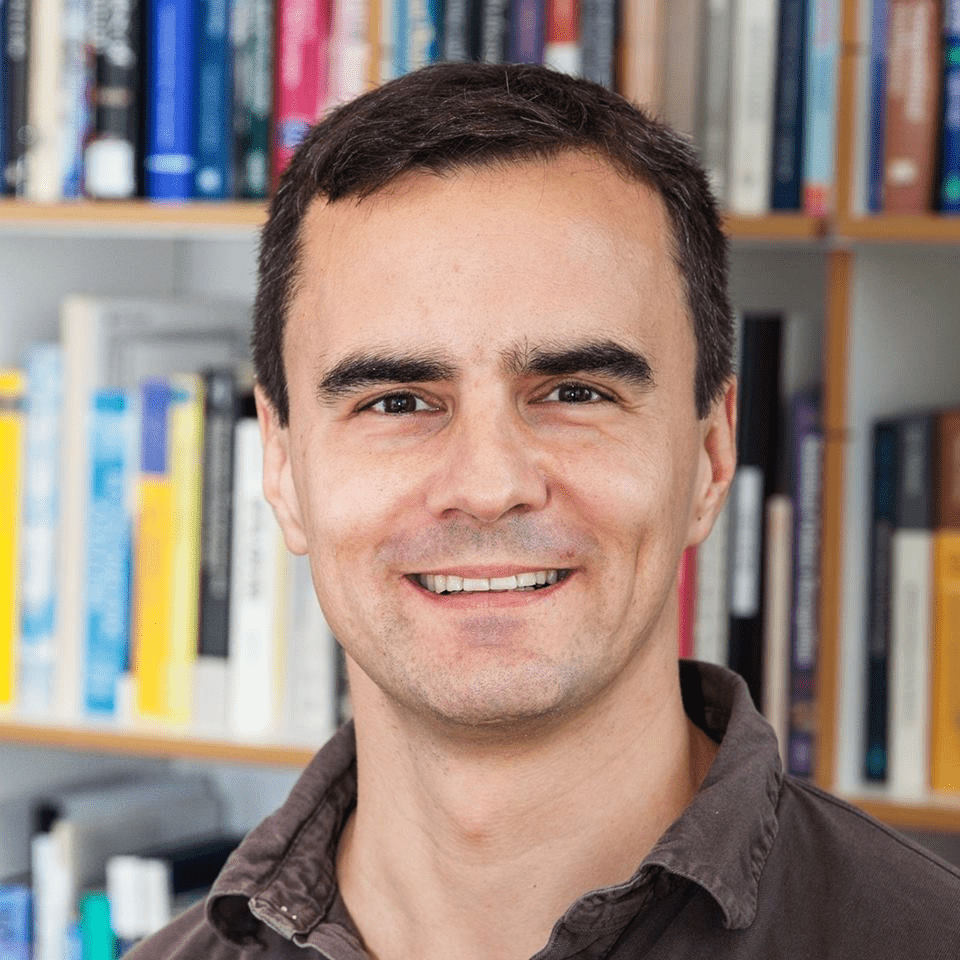}}]{Jan Peters} is a full professor (W3) for Intelligent Autonomous Systems at the Computer Science Department of the Technical University of Darmstadt, dept head of the research department on Systems AI for Robot Learning (SAIROL) at the German Research Center for Artificial Intelligence (Deutsches Forschungszentrum für Künstliche Intelligenz, DFKI), and
a founding research faculty member of The Hessian Center for Artificial Intelligence. He has received the Dick Volz Best 2007 US Ph.D. Thesis Runner-Up Award, RSS - Early Career Spotlight, INNS Young Investigator Award, and IEEE Robotics \& Automation Society’s Early Career Award, as well as numerous best paper awards. He received an ERC Starting Grant and was appointed an IEEE fellow, AIAA fellow and ELLIS fellow.
\end{IEEEbiography}

\vfill

\end{document}

%% file: misc/math_commands.tex
\def\1{\bm{1}}

\def\vc{{\bm{c}}}
\def\vct{{\bm{\tilde{c}}}}

\def\vx{{\bm{x}}}

\def\mA{{\bm{A}}}

\newcommand\norm[1]{\lVert#1\rVert}

%% file: pictures/dot_track/dot_track_small.pdf_tex
\begingroup%
  \makeatletter%
  \providecommand\color[2][]{%
    \errmessage{(Inkscape) Color is used for the text in Inkscape, but the package 'color.sty' is not loaded}%
    \renewcommand\color[2][]{}%
  }%
  \providecommand\transparent[1]{%
    \errmessage{(Inkscape) Transparency is used (non-zero) for the text in Inkscape, but the package 'transparent.sty' is not loaded}%
    \renewcommand\transparent[1]{}%
  }%
  \providecommand\rotatebox[2]{#2}%
  \newcommand*\fsize{\dimexpr\f@size pt\relax}%
  \newcommand*\lineheight[1]{\fontsize{\fsize}{#1\fsize}\selectfont}%
  \ifx\svgwidth\undefined%
    \setlength{\unitlength}{335.17575073bp}%
    \ifx\svgscale\undefined%
      \relax%
    \else%
      \setlength{\unitlength}{\unitlength * \real{\svgscale}}%
    \fi%
  \else%
    \setlength{\unitlength}{\svgwidth}%
  \fi%
  \global\let\svgwidth\undefined%
  \global\let\svgscale\undefined%
  \makeatother%
  \begin{picture}(1,0.24357598)%
    \lineheight{1}%
    \setlength\tabcolsep{0pt}%
    \put(0,0){\includegraphics[width=\unitlength,page=1]{dot_track_small.pdf}}%
    \put(0.139852,-0.00287156){\color[rgb]{0,0,0}\makebox(0,0)[t]{\lineheight{1.25}\smash{\begin{tabular}[t]{c}\small{$t=t_i$}\end{tabular}}}}%
    \put(0.49787313,-0.00287156){\color[rgb]{0,0,0}\makebox(0,0)[t]{\lineheight{1.25}\smash{\begin{tabular}[t]{c}\small{$t=t_{i+1}$}\end{tabular}}}}%
    \put(0.85589426,-0.00287156){\color[rgb]{0,0,0}\makebox(0,0)[t]{\lineheight{1.25}\smash{\begin{tabular}[t]{c}\small{$\mathcal{S}_\mathrm{E}(t_{i+1})$}\end{tabular}}}}%
  \end{picture}%
\endgroup%

%% file: pictures/results/frequency_analysis/freq_spec.pdf_tex
\begingroup%
  \makeatletter%
  \providecommand\color[2][]{%
    \errmessage{(Inkscape) Color is used for the text in Inkscape, but the package 'color.sty' is not loaded}%
    \renewcommand\color[2][]{}%
  }%
  \providecommand\transparent[1]{%
    \errmessage{(Inkscape) Transparency is used (non-zero) for the text in Inkscape, but the package 'transparent.sty' is not loaded}%
    \renewcommand\transparent[1]{}%
  }%
  \providecommand\rotatebox[2]{#2}%
  \newcommand*\fsize{\dimexpr\f@size pt\relax}%
  \newcommand*\lineheight[1]{\fontsize{\fsize}{#1\fsize}\selectfont}%
  \ifx\svgwidth\undefined%
    \setlength{\unitlength}{252bp}%
    \ifx\svgscale\undefined%
      \relax%
    \else%
      \setlength{\unitlength}{\unitlength * \real{\svgscale}}%
    \fi%
  \else%
    \setlength{\unitlength}{\svgwidth}%
  \fi%
  \global\let\svgwidth\undefined%
  \global\let\svgscale\undefined%
  \makeatother%
  \begin{picture}(1,0.51333332)%
    \lineheight{1}%
    \setlength\tabcolsep{0pt}%
    \put(0,0){\includegraphics[width=\unitlength,page=1]{freq_spec.pdf}}%
    \put(0.18960441,0.04346213){\color[rgb]{0,0,0}\makebox(0,0)[lt]{\lineheight{1.25}\smash{\begin{tabular}[t]{l}0\end{tabular}}}}%
    \put(0,0){\includegraphics[width=\unitlength,page=2]{freq_spec.pdf}}%
    \put(0.27112692,0.04346213){\color[rgb]{0,0,0}\makebox(0,0)[lt]{\lineheight{1.25}\smash{\begin{tabular}[t]{l}100\end{tabular}}}}%
    \put(0,0){\includegraphics[width=\unitlength,page=3]{freq_spec.pdf}}%
    \put(0.37788504,0.04346213){\color[rgb]{0,0,0}\makebox(0,0)[lt]{\lineheight{1.25}\smash{\begin{tabular}[t]{l}200\end{tabular}}}}%
    \put(0,0){\includegraphics[width=\unitlength,page=4]{freq_spec.pdf}}%
    \put(0.48464316,0.04346213){\color[rgb]{0,0,0}\makebox(0,0)[lt]{\lineheight{1.25}\smash{\begin{tabular}[t]{l}300\end{tabular}}}}%
    \put(0,0){\includegraphics[width=\unitlength,page=5]{freq_spec.pdf}}%
    \put(0.59140129,0.04346213){\color[rgb]{0,0,0}\makebox(0,0)[lt]{\lineheight{1.25}\smash{\begin{tabular}[t]{l}400\end{tabular}}}}%
    \put(0,0){\includegraphics[width=\unitlength,page=6]{freq_spec.pdf}}%
    \put(0.69602423,0.04346213){\color[rgb]{0,0,0}\makebox(0,0)[lt]{\lineheight{1.25}\smash{\begin{tabular}[t]{l}498\end{tabular}}}}%
    \put(0,0){\includegraphics[width=\unitlength,page=7]{freq_spec.pdf}}%
    \put(0.14920883,0.08630687){\color[rgb]{0,0,0}\makebox(0,0)[lt]{\lineheight{1.25}\smash{\begin{tabular}[t]{l}0\end{tabular}}}}%
    \put(0,0){\includegraphics[width=\unitlength,page=8]{freq_spec.pdf}}%
    \put(0.12397321,0.12461088){\color[rgb]{0,0,0}\makebox(0,0)[lt]{\lineheight{1.25}\smash{\begin{tabular}[t]{l}50\end{tabular}}}}%
    \put(0,0){\includegraphics[width=\unitlength,page=9]{freq_spec.pdf}}%
    \put(0.0987376,0.16291489){\color[rgb]{0,0,0}\makebox(0,0)[lt]{\lineheight{1.25}\smash{\begin{tabular}[t]{l}100\end{tabular}}}}%
    \put(0,0){\includegraphics[width=\unitlength,page=10]{freq_spec.pdf}}%
    \put(0.0987376,0.2012189){\color[rgb]{0,0,0}\makebox(0,0)[lt]{\lineheight{1.25}\smash{\begin{tabular}[t]{l}150\end{tabular}}}}%
    \put(0,0){\includegraphics[width=\unitlength,page=11]{freq_spec.pdf}}%
    \put(0.0987376,0.23952291){\color[rgb]{0,0,0}\makebox(0,0)[lt]{\lineheight{1.25}\smash{\begin{tabular}[t]{l}200\end{tabular}}}}%
    \put(0,0){\includegraphics[width=\unitlength,page=12]{freq_spec.pdf}}%
    \put(0.0987376,0.27782691){\color[rgb]{0,0,0}\makebox(0,0)[lt]{\lineheight{1.25}\smash{\begin{tabular}[t]{l}250\end{tabular}}}}%
    \put(0,0){\includegraphics[width=\unitlength,page=13]{freq_spec.pdf}}%
    \put(0.0987376,0.31613093){\color[rgb]{0,0,0}\makebox(0,0)[lt]{\lineheight{1.25}\smash{\begin{tabular}[t]{l}300\end{tabular}}}}%
    \put(0,0){\includegraphics[width=\unitlength,page=14]{freq_spec.pdf}}%
    \put(0.0987376,0.35443493){\color[rgb]{0,0,0}\makebox(0,0)[lt]{\lineheight{1.25}\smash{\begin{tabular}[t]{l}350\end{tabular}}}}%
    \put(0,0){\includegraphics[width=\unitlength,page=15]{freq_spec.pdf}}%
    \put(0.0987376,0.39273896){\color[rgb]{0,0,0}\makebox(0,0)[lt]{\lineheight{1.25}\smash{\begin{tabular}[t]{l}400\end{tabular}}}}%
    \put(0,0){\includegraphics[width=\unitlength,page=16]{freq_spec.pdf}}%
    \put(0.0987376,0.43104296){\color[rgb]{0,0,0}\makebox(0,0)[lt]{\lineheight{1.25}\smash{\begin{tabular}[t]{l}450\end{tabular}}}}%
    \put(0,0){\includegraphics[width=\unitlength,page=17]{freq_spec.pdf}}%
    \put(0.0987376,0.46781479){\color[rgb]{0,0,0}\makebox(0,0)[lt]{\lineheight{1.25}\smash{\begin{tabular}[t]{l}498\end{tabular}}}}%
    \put(0,0){\includegraphics[width=\unitlength,page=18]{freq_spec.pdf}}%
    \put(0.84059236,0.15029197){\color[rgb]{0,0,0}\makebox(0,0)[lt]{\lineheight{1.25}\smash{\begin{tabular}[t]{l}0.0\end{tabular}}}}%
    \put(0,0){\includegraphics[width=\unitlength,page=19]{freq_spec.pdf}}%
    \put(0.84059236,0.28740292){\color[rgb]{0,0,0}\makebox(0,0)[lt]{\lineheight{1.25}\smash{\begin{tabular}[t]{l}0.5\end{tabular}}}}%
    \put(0,0){\includegraphics[width=\unitlength,page=20]{freq_spec.pdf}}%
    \put(0.84059236,0.42451388){\color[rgb]{0,0,0}\makebox(0,0)[lt]{\lineheight{1.25}\smash{\begin{tabular}[t]{l}1.0\end{tabular}}}}%
    \put(0,0){\includegraphics[width=\unitlength,page=21]{freq_spec.pdf}}%
    \put(0.93781458,0.19743519){\color[rgb]{0,0,0}\rotatebox{90}{\makebox(0,0)[lt]{\lineheight{1.25}\smash{\begin{tabular}[t]{l}norm amplitude\end{tabular}}}}}%
    \put(0,0){\includegraphics[width=\unitlength,page=22]{freq_spec.pdf}}%
    \put(0.07667778,0.2324494){\color[rgb]{0,0,0}\rotatebox{90}{\makebox(0,0)[lt]{\lineheight{1.25}\smash{\begin{tabular}[t]{l}$f_d [\SI{}{\hertz}]$\end{tabular}}}}}%
    \put(0.48214038,0.0091531){\color[rgb]{0,0,0}\makebox(0,0)[t]{\lineheight{1.25}\smash{\begin{tabular}[t]{c}frequency [$\SI{}{\hertz}$]\end{tabular}}}}%
  \end{picture}%
\endgroup%

%% file: pictures/results/bandwidth/rgb_eb_bandwidth.pdf_tex
\begingroup%
  \makeatletter%
  \providecommand\color[2][]{%
    \errmessage{(Inkscape) Color is used for the text in Inkscape, but the package 'color.sty' is not loaded}%
    \renewcommand\color[2][]{}%
  }%
  \providecommand\transparent[1]{%
    \errmessage{(Inkscape) Transparency is used (non-zero) for the text in Inkscape, but the package 'transparent.sty' is not loaded}%
    \renewcommand\transparent[1]{}%
  }%
  \providecommand\rotatebox[2]{#2}%
  \newcommand*\fsize{\dimexpr\f@size pt\relax}%
  \newcommand*\lineheight[1]{\fontsize{\fsize}{#1\fsize}\selectfont}%
  \ifx\svgwidth\undefined%
    \setlength{\unitlength}{263.62204724bp}%
    \ifx\svgscale\undefined%
      \relax%
    \else%
      \setlength{\unitlength}{\unitlength * \real{\svgscale}}%
    \fi%
  \else%
    \setlength{\unitlength}{\svgwidth}%
  \fi%
  \global\let\svgwidth\undefined%
  \global\let\svgscale\undefined%
  \makeatother%
  \begin{picture}(1,0.523957)%
    \lineheight{1}%
    \setlength\tabcolsep{0pt}%
    \put(0,0){\includegraphics[width=\unitlength,page=1]{rgb_eb_bandwidth.pdf}}%
    \put(0.13146809,0.40477797){\color[rgb]{0,0,0}\makebox(0,0)[lt]{\lineheight{1.25}\smash{\begin{tabular}[t]{l}{\footnotesize 750}\end{tabular}}}}%
    \put(0,0){\includegraphics[width=\unitlength,page=2]{rgb_eb_bandwidth.pdf}}%
    \put(0.13146809,0.50508795){\color[rgb]{0,0,0}\makebox(0,0)[lt]{\lineheight{1.25}\smash{\begin{tabular}[t]{l}{\footnotesize 800}\end{tabular}}}}%
    \put(0.08385044,0.16830299){\color[rgb]{0,0,0}\rotatebox{90}{\makebox(0,0)[lt]{\lineheight{1.25}\smash{\begin{tabular}[t]{l}sensor output [kbytes]\end{tabular}}}}}%
    \put(0,0){\includegraphics[width=\unitlength,page=3]{rgb_eb_bandwidth.pdf}}%
    \put(0.50297651,0.48360595){\color[rgb]{0,0,0}\makebox(0,0)[lt]{\lineheight{1.25}\smash{\begin{tabular}[t]{l}{\footnotesize Evetac}\end{tabular}}}}%
    \put(0,0){\includegraphics[width=\unitlength,page=4]{rgb_eb_bandwidth.pdf}}%
    \put(0.64015081,0.48299863){\color[rgb]{0,0,0}\makebox(0,0)[lt]{\lineheight{1.25}\smash{\begin{tabular}[t]{l}{\footnotesize RGB Optical Tactile}\end{tabular}}}}%
    \put(0,0){\includegraphics[width=\unitlength,page=5]{rgb_eb_bandwidth.pdf}}%
    \put(0.18343838,0.16297678){\color[rgb]{0,0,0}\makebox(0,0)[lt]{\lineheight{1.25}\smash{\begin{tabular}[t]{l}{\footnotesize 0}\end{tabular}}}}%
    \put(0,0){\includegraphics[width=\unitlength,page=6]{rgb_eb_bandwidth.pdf}}%
    \put(0.35913147,0.16297678){\color[rgb]{0,0,0}\makebox(0,0)[lt]{\lineheight{1.25}\smash{\begin{tabular}[t]{l}{\footnotesize 5}\end{tabular}}}}%
    \put(0,0){\includegraphics[width=\unitlength,page=7]{rgb_eb_bandwidth.pdf}}%
    \put(0.52900408,0.16297678){\color[rgb]{0,0,0}\makebox(0,0)[lt]{\lineheight{1.25}\smash{\begin{tabular}[t]{l}{\footnotesize 10}\end{tabular}}}}%
    \put(0,0){\includegraphics[width=\unitlength,page=8]{rgb_eb_bandwidth.pdf}}%
    \put(0.70469721,0.16297678){\color[rgb]{0,0,0}\makebox(0,0)[lt]{\lineheight{1.25}\smash{\begin{tabular}[t]{l}{\footnotesize 15}\end{tabular}}}}%
    \put(0,0){\includegraphics[width=\unitlength,page=9]{rgb_eb_bandwidth.pdf}}%
    \put(0.16456129,0.19822635){\color[rgb]{0,0,0}\makebox(0,0)[lt]{\lineheight{1.25}\smash{\begin{tabular}[t]{l}{\footnotesize 0}\end{tabular}}}}%
    \put(0,0){\includegraphics[width=\unitlength,page=10]{rgb_eb_bandwidth.pdf}}%
    \put(0.14036057,0.25676376){\color[rgb]{0,0,0}\makebox(0,0)[lt]{\lineheight{1.25}\smash{\begin{tabular}[t]{l}{\footnotesize 10}\end{tabular}}}}%
    \put(0,0){\includegraphics[width=\unitlength,page=11]{rgb_eb_bandwidth.pdf}}%
    \put(0.14036057,0.31530112){\color[rgb]{0,0,0}\makebox(0,0)[lt]{\lineheight{1.25}\smash{\begin{tabular}[t]{l}{\footnotesize 20}\end{tabular}}}}%
    \put(0.8801831,0.16297678){\color[rgb]{0,0,0}\makebox(0,0)[lt]{\lineheight{1.25}\smash{\begin{tabular}[t]{l}{\footnotesize 20}\end{tabular}}}}%
    \put(-0.14416613,0.64317199){\color[rgb]{0,1,0}\makebox(0,0)[lt]{\begin{minipage}{2.52627101\unitlength}\raggedright \end{minipage}}}%
    \put(-0.60401069,0.68939904){\color[rgb]{0,1,0}\makebox(0,0)[lt]{\begin{minipage}{0.42535621\unitlength}\raggedright \end{minipage}}}%
    \put(0,0){\includegraphics[width=\unitlength,page=12]{rgb_eb_bandwidth.pdf}}%
    \put(0.5157809,0.13998954){\color[rgb]{0,0,0}\makebox(0,0)[lt]{\lineheight{1.25}\smash{\begin{tabular}[t]{l}{\footnotesize time [s]}\end{tabular}}}}%
    \put(0,0){\includegraphics[width=\unitlength,page=13]{rgb_eb_bandwidth.pdf}}%
    \put(0.18559246,0.0295661){\color[rgb]{0,0,0}\makebox(0,0)[lt]{\lineheight{1.25}\smash{\begin{tabular}[t]{l}\scriptsize{No Contact}\\\end{tabular}}}}%
    \put(0.40731549,0.02956607){\color[rgb]{0,0,0}\makebox(0,0)[lt]{\lineheight{1.25}\smash{\begin{tabular}[t]{l}\scriptsize{Making Contact}\\\end{tabular}}}}%
    \put(0,0){\includegraphics[width=\unitlength,page=14]{rgb_eb_bandwidth.pdf}}%
    \put(0.62357058,0.02939259){\color[rgb]{0,0,0}\makebox(0,0)[lt]{\lineheight{1.25}\smash{\begin{tabular}[t]{l}\scriptsize{Perturbation}\end{tabular}}}}%
    \put(0,0){\includegraphics[width=\unitlength,page=15]{rgb_eb_bandwidth.pdf}}%
    \put(0.78576304,0.02983627){\color[rgb]{0,0,0}\makebox(0,0)[lt]{\lineheight{1.25}\smash{\begin{tabular}[t]{l}\scriptsize{Slippage}\end{tabular}}}}%
    \put(0,0){\includegraphics[width=\unitlength,page=16]{rgb_eb_bandwidth.pdf}}%
    \put(0.18559246,0.00658419){\color[rgb]{0,0,0}\makebox(0,0)[lt]{\lineheight{1.25}\smash{\begin{tabular}[t]{l}\scriptsize{Gripper Open}\end{tabular}}}}%
    \put(0.40808422,0.00658401){\color[rgb]{0,0,0}\makebox(0,0)[lt]{\lineheight{1.25}\smash{\begin{tabular}[t]{l}\scriptsize{Gripper Closing}\end{tabular}}}}%
    \put(0.78545957,0.00658401){\color[rgb]{0,0,0}\makebox(0,0)[lt]{\lineheight{1.25}\smash{\begin{tabular}[t]{l}\scriptsize{Gripper Opening}\end{tabular}}}}%
  \end{picture}%
\endgroup%

%% file: pictures/results/force/res_force_recon.pdf_tex
\begingroup%
  \makeatletter%
  \providecommand\color[2][]{%
    \errmessage{(Inkscape) Color is used for the text in Inkscape, but the package 'color.sty' is not loaded}%
    \renewcommand\color[2][]{}%
  }%
  \providecommand\transparent[1]{%
    \errmessage{(Inkscape) Transparency is used (non-zero) for the text in Inkscape, but the package 'transparent.sty' is not loaded}%
    \renewcommand\transparent[1]{}%
  }%
  \providecommand\rotatebox[2]{#2}%
  \newcommand*\fsize{\dimexpr\f@size pt\relax}%
  \newcommand*\lineheight[1]{\fontsize{\fsize}{#1\fsize}\selectfont}%
  \ifx\svgwidth\undefined%
    \setlength{\unitlength}{232.15748464bp}%
    \ifx\svgscale\undefined%
      \relax%
    \else%
      \setlength{\unitlength}{\unitlength * \real{\svgscale}}%
    \fi%
  \else%
    \setlength{\unitlength}{\svgwidth}%
  \fi%
  \global\let\svgwidth\undefined%
  \global\let\svgscale\undefined%
  \makeatother%
  \begin{picture}(1,0.87234067)%
    \lineheight{1}%
    \setlength\tabcolsep{0pt}%
    \put(0,0){\includegraphics[width=\unitlength,page=1]{res_force_recon.pdf}}%
    \put(0.0617953,0.49786868){\color[rgb]{0,0,0}\makebox(0,0)[lt]{\lineheight{1.25}\smash{\begin{tabular}[t]{l}−\end{tabular}}}}%
    \put(0.09784841,0.49786868){\color[rgb]{0,0,0}\makebox(0,0)[lt]{\lineheight{1.25}\smash{\begin{tabular}[t]{l}10\end{tabular}}}}%
    \put(0,0){\includegraphics[width=\unitlength,page=2]{res_force_recon.pdf}}%
    \put(0.07626554,0.57442885){\color[rgb]{0,0,0}\makebox(0,0)[lt]{\lineheight{1.25}\smash{\begin{tabular}[t]{l}−\end{tabular}}}}%
    \put(0.11231865,0.57442885){\color[rgb]{0,0,0}\makebox(0,0)[lt]{\lineheight{1.25}\smash{\begin{tabular}[t]{l}5\end{tabular}}}}%
    \put(0,0){\includegraphics[width=\unitlength,page=3]{res_force_recon.pdf}}%
    \put(0.11234019,0.65098906){\color[rgb]{0,0,0}\makebox(0,0)[lt]{\lineheight{1.25}\smash{\begin{tabular}[t]{l}0\end{tabular}}}}%
    \put(0,0){\includegraphics[width=\unitlength,page=4]{res_force_recon.pdf}}%
    \put(0.11234019,0.72754926){\color[rgb]{0,0,0}\makebox(0,0)[lt]{\lineheight{1.25}\smash{\begin{tabular}[t]{l}5\end{tabular}}}}%
    \put(0,0){\includegraphics[width=\unitlength,page=5]{res_force_recon.pdf}}%
    \put(0.09786995,0.80410947){\color[rgb]{0,0,0}\makebox(0,0)[lt]{\lineheight{1.25}\smash{\begin{tabular}[t]{l}10\end{tabular}}}}%
    \put(0.04207541,0.57288728){\color[rgb]{0,0,0}\rotatebox{90}{\makebox(0,0)[lt]{\lineheight{1.25}\smash{\begin{tabular}[t]{l}\small{$F[\SI{}{\N}]$ (Evetac)}\end{tabular}}}}}%
    \put(0,0){\includegraphics[width=\unitlength,page=6]{res_force_recon.pdf}}%
    \put(0.27326275,0.51354027){\color[rgb]{0,0,0}\makebox(0,0)[lt]{\lineheight{1.25}\smash{\begin{tabular}[t]{l}\small{$F_x$ GT}\end{tabular}}}}%
    \put(0,0){\includegraphics[width=\unitlength,page=7]{res_force_recon.pdf}}%
    \put(0.27326275,0.47954905){\color[rgb]{0,0,0}\makebox(0,0)[lt]{\lineheight{1.25}\smash{\begin{tabular}[t]{l}\small{$F_x$ LR}\end{tabular}}}}%
    \put(0,0){\includegraphics[width=\unitlength,page=8]{res_force_recon.pdf}}%
    \put(0.50323871,0.51354027){\color[rgb]{0,0,0}\makebox(0,0)[lt]{\lineheight{1.25}\smash{\begin{tabular}[t]{l}\small{$F_x$ NN}\end{tabular}}}}%
    \put(0,0){\includegraphics[width=\unitlength,page=9]{res_force_recon.pdf}}%
    \put(0.50323871,0.47954906){\color[rgb]{0,0,0}\makebox(0,0)[lt]{\lineheight{1.25}\smash{\begin{tabular}[t]{l}\small{$F_y$ GT}\end{tabular}}}}%
    \put(0,0){\includegraphics[width=\unitlength,page=10]{res_force_recon.pdf}}%
    \put(0.7832209,0.51354027){\color[rgb]{0,0,0}\makebox(0,0)[lt]{\lineheight{1.25}\smash{\begin{tabular}[t]{l}\small{$F_y$ LR}\end{tabular}}}}%
    \put(0,0){\includegraphics[width=\unitlength,page=11]{res_force_recon.pdf}}%
    \put(0.7832209,0.47954905){\color[rgb]{0,0,0}\makebox(0,0)[lt]{\lineheight{1.25}\smash{\begin{tabular}[t]{l}\small{$F_y$ NN}\end{tabular}}}}%
    \put(0,0){\includegraphics[width=\unitlength,page=12]{res_force_recon.pdf}}%
    \put(0.19172854,0.03323018){\color[rgb]{0,0,0}\makebox(0,0)[lt]{\lineheight{1.25}\smash{\begin{tabular}[t]{l}0\end{tabular}}}}%
    \put(0,0){\includegraphics[width=\unitlength,page=13]{res_force_recon.pdf}}%
    \put(0.3240318,0.03323018){\color[rgb]{0,0,0}\makebox(0,0)[lt]{\lineheight{1.25}\smash{\begin{tabular}[t]{l}20\end{tabular}}}}%
    \put(0,0){\includegraphics[width=\unitlength,page=14]{res_force_recon.pdf}}%
    \put(0.47003135,0.03323018){\color[rgb]{0,0,0}\makebox(0,0)[lt]{\lineheight{1.25}\smash{\begin{tabular}[t]{l}40\end{tabular}}}}%
    \put(0,0){\includegraphics[width=\unitlength,page=15]{res_force_recon.pdf}}%
    \put(0.61603088,0.03323018){\color[rgb]{0,0,0}\makebox(0,0)[lt]{\lineheight{1.25}\smash{\begin{tabular}[t]{l}60\end{tabular}}}}%
    \put(0,0){\includegraphics[width=\unitlength,page=16]{res_force_recon.pdf}}%
    \put(0.76203037,0.03323018){\color[rgb]{0,0,0}\makebox(0,0)[lt]{\lineheight{1.25}\smash{\begin{tabular}[t]{l}80\end{tabular}}}}%
    \put(0,0){\includegraphics[width=\unitlength,page=17]{res_force_recon.pdf}}%
    \put(0.06179532,0.10073205){\color[rgb]{0,0,0}\makebox(0,0)[lt]{\lineheight{1.25}\smash{\begin{tabular}[t]{l}−\end{tabular}}}}%
    \put(0.09784843,0.10073205){\color[rgb]{0,0,0}\makebox(0,0)[lt]{\lineheight{1.25}\smash{\begin{tabular}[t]{l}10\end{tabular}}}}%
    \put(0,0){\includegraphics[width=\unitlength,page=18]{res_force_recon.pdf}}%
    \put(0.07626556,0.17729231){\color[rgb]{0,0,0}\makebox(0,0)[lt]{\lineheight{1.25}\smash{\begin{tabular}[t]{l}−\end{tabular}}}}%
    \put(0.11231867,0.17729231){\color[rgb]{0,0,0}\makebox(0,0)[lt]{\lineheight{1.25}\smash{\begin{tabular}[t]{l}5\end{tabular}}}}%
    \put(0,0){\includegraphics[width=\unitlength,page=19]{res_force_recon.pdf}}%
    \put(0.11234022,0.25385248){\color[rgb]{0,0,0}\makebox(0,0)[lt]{\lineheight{1.25}\smash{\begin{tabular}[t]{l}0\end{tabular}}}}%
    \put(0,0){\includegraphics[width=\unitlength,page=20]{res_force_recon.pdf}}%
    \put(0.11234022,0.33041265){\color[rgb]{0,0,0}\makebox(0,0)[lt]{\lineheight{1.25}\smash{\begin{tabular}[t]{l}5\end{tabular}}}}%
    \put(0,0){\includegraphics[width=\unitlength,page=21]{res_force_recon.pdf}}%
    \put(0.09786997,0.40697283){\color[rgb]{0,0,0}\makebox(0,0)[lt]{\lineheight{1.25}\smash{\begin{tabular}[t]{l}10\end{tabular}}}}%
    \put(0.04207544,0.11555225){\color[rgb]{0,0,0}\rotatebox{90}{\makebox(0,0)[lt]{\lineheight{1.25}\smash{\begin{tabular}[t]{l}\small{$F[\SI{}{\N}]$ (RGB Optical)}\end{tabular}}}}}%
    \put(0,0){\includegraphics[width=\unitlength,page=22]{res_force_recon.pdf}}%
    \put(0.27326272,0.11640365){\color[rgb]{0,0,0}\makebox(0,0)[lt]{\lineheight{1.25}\smash{\begin{tabular}[t]{l}\small{$F_x$ GT}\end{tabular}}}}%
    \put(0,0){\includegraphics[width=\unitlength,page=23]{res_force_recon.pdf}}%
    \put(0.27326272,0.08241236){\color[rgb]{0,0,0}\makebox(0,0)[lt]{\lineheight{1.25}\smash{\begin{tabular}[t]{l}\small{$F_x$ LR}\end{tabular}}}}%
    \put(0,0){\includegraphics[width=\unitlength,page=24]{res_force_recon.pdf}}%
    \put(0.50969977,0.11640365){\color[rgb]{0,0,0}\makebox(0,0)[lt]{\lineheight{1.25}\smash{\begin{tabular}[t]{l}\small{$F_x$ NN}\end{tabular}}}}%
    \put(0,0){\includegraphics[width=\unitlength,page=25]{res_force_recon.pdf}}%
    \put(0.50969977,0.08241234){\color[rgb]{0,0,0}\makebox(0,0)[lt]{\lineheight{1.25}\smash{\begin{tabular}[t]{l}\small{$F_y$ GT}\end{tabular}}}}%
    \put(0,0){\includegraphics[width=\unitlength,page=26]{res_force_recon.pdf}}%
    \put(0.78322086,0.11640365){\color[rgb]{0,0,0}\makebox(0,0)[lt]{\lineheight{1.25}\smash{\begin{tabular}[t]{l}\small{$F_y$ LR}\end{tabular}}}}%
    \put(0,0){\includegraphics[width=\unitlength,page=27]{res_force_recon.pdf}}%
    \put(0.78322087,0.08241236){\color[rgb]{0,0,0}\makebox(0,0)[lt]{\lineheight{1.25}\smash{\begin{tabular}[t]{l}\small{$F_y$ NN}\end{tabular}}}}%
    \put(0.89433414,0.03323038){\color[rgb]{0,0,0}\makebox(0,0)[lt]{\lineheight{1.25}\smash{\begin{tabular}[t]{l}100\\\end{tabular}}}}%
    \put(0.57722721,0.00297804){\color[rgb]{0,0,0}\makebox(0,0)[t]{\lineheight{1.25}\smash{\begin{tabular}[t]{c}time [s]\end{tabular}}}}%
  \end{picture}%
\endgroup%

%% file: pictures/onset_slip/slip_label.pdf_tex
\begingroup%
  \makeatletter%
  \providecommand\color[2][]{%
    \errmessage{(Inkscape) Color is used for the text in Inkscape, but the package 'color.sty' is not loaded}%
    \renewcommand\color[2][]{}%
  }%
  \providecommand\transparent[1]{%
    \errmessage{(Inkscape) Transparency is used (non-zero) for the text in Inkscape, but the package 'transparent.sty' is not loaded}%
    \renewcommand\transparent[1]{}%
  }%
  \providecommand\rotatebox[2]{#2}%
  \newcommand*\fsize{\dimexpr\f@size pt\relax}%
  \newcommand*\lineheight[1]{\fontsize{\fsize}{#1\fsize}\selectfont}%
  \ifx\svgwidth\undefined%
    \setlength{\unitlength}{1499.39866434bp}%
    \ifx\svgscale\undefined%
      \relax%
    \else%
      \setlength{\unitlength}{\unitlength * \real{\svgscale}}%
    \fi%
  \else%
    \setlength{\unitlength}{\svgwidth}%
  \fi%
  \global\let\svgwidth\undefined%
  \global\let\svgscale\undefined%
  \makeatother%
  \begin{picture}(1,0.35057947)%
    \lineheight{1}%
    \setlength\tabcolsep{0pt}%
    \put(0,0){\includegraphics[width=\unitlength,page=1]{slip_label.pdf}}%
    \put(0.23910454,0.00157289){\color[rgb]{0,0,0}\makebox(0,0)[t]{\lineheight{1.25}\smash{\begin{tabular}[t]{c}{\footnotesize $t^c_{s0}{-} \SI{30}{\ms}$}\end{tabular}}}}%
    \put(0.61955224,0.00157289){\color[rgb]{0,0,0}\makebox(0,0)[t]{\lineheight{1.25}\smash{\begin{tabular}[t]{c}{\footnotesize $t^c_{s0}$}\end{tabular}}}}%
    \put(0.88044781,0.00157289){\color[rgb]{0,0,0}\makebox(0,0)[t]{\lineheight{1.25}\smash{\begin{tabular}[t]{c}{\footnotesize $t^c_{s0}{+} \SI{30}{\ms}$}\end{tabular}}}}%
  \end{picture}%
\endgroup%

%% file: pictures/netw_arch/netw_arch.pdf_tex
\begingroup%
  \makeatletter%
  \providecommand\color[2][]{%
    \errmessage{(Inkscape) Color is used for the text in Inkscape, but the package 'color.sty' is not loaded}%
    \renewcommand\color[2][]{}%
  }%
  \providecommand\transparent[1]{%
    \errmessage{(Inkscape) Transparency is used (non-zero) for the text in Inkscape, but the package 'transparent.sty' is not loaded}%
    \renewcommand\transparent[1]{}%
  }%
  \providecommand\rotatebox[2]{#2}%
  \newcommand*\fsize{\dimexpr\f@size pt\relax}%
  \newcommand*\lineheight[1]{\fontsize{\fsize}{#1\fsize}\selectfont}%
  \ifx\svgwidth\undefined%
    \setlength{\unitlength}{795bp}%
    \ifx\svgscale\undefined%
      \relax%
    \else%
      \setlength{\unitlength}{\unitlength * \real{\svgscale}}%
    \fi%
  \else%
    \setlength{\unitlength}{\svgwidth}%
  \fi%
  \global\let\svgwidth\undefined%
  \global\let\svgscale\undefined%
  \makeatother%
  \begin{picture}(1,0.28018868)%
    \lineheight{1}%
    \setlength\tabcolsep{0pt}%
    \put(0.26415094,0.01413396){\color[rgb]{0,0,0}\makebox(0,0)[lt]{\lineheight{1.25}\smash{\begin{tabular}[t]{l}{\footnotesize fully connected + ReLU}\end{tabular}}}}%
    \put(0,0){\includegraphics[width=\unitlength,page=1]{netw_arch.pdf}}%
    \put(0.35566167,0.20182739){\color[rgb]{0,0,0}\makebox(0,0)[t]{\lineheight{1.25}\smash{\begin{tabular}[t]{c}{\small $\mathcal{F}_E$}\end{tabular}}}}%
    \put(0.2711051,0.20182739){\color[rgb]{0,0,0}\makebox(0,0)[t]{\lineheight{1.25}\smash{\begin{tabular}[t]{c}{\small $\mathcal{F}_D$}\end{tabular}}}}%
    \put(0,0){\includegraphics[width=\unitlength,page=2]{netw_arch.pdf}}%
    \put(0.47469245,0.14622642){\color[rgb]{0,0,0}\makebox(0,0)[t]{\lineheight{1.25}\smash{\begin{tabular}[t]{c}{\small fc1}\end{tabular}}}}%
    \put(0,0){\includegraphics[width=\unitlength,page=3]{netw_arch.pdf}}%
    \put(0.54969245,0.14622642){\color[rgb]{0,0,0}\makebox(0,0)[t]{\lineheight{1.25}\smash{\begin{tabular}[t]{c}{\small fc2}\end{tabular}}}}%
    \put(0,0){\includegraphics[width=\unitlength,page=4]{netw_arch.pdf}}%
    \put(0.31384339,0.07169811){\color[rgb]{0,0,0}\makebox(0,0)[t]{\lineheight{1.25}\smash{\begin{tabular}[t]{c}{\small $1 \textrm{x} l_i$}\end{tabular}}}}%
    \put(0,0){\includegraphics[width=\unitlength,page=5]{netw_arch.pdf}}%
    \put(0.47563585,0.08018868){\color[rgb]{0,0,0}\makebox(0,0)[t]{\lineheight{1.25}\smash{\begin{tabular}[t]{c}{\small $1 \textrm{x} l_{\textrm{fc1}}$}\end{tabular}}}}%
    \put(0,0){\includegraphics[width=\unitlength,page=6]{netw_arch.pdf}}%
    \put(0.55063585,0.08018868){\color[rgb]{0,0,0}\makebox(0,0)[t]{\lineheight{1.25}\smash{\begin{tabular}[t]{c}{\small $1 \textrm{x} l_{\textrm{fc2}}$}\end{tabular}}}}%
    \put(0,0){\includegraphics[width=\unitlength,page=7]{netw_arch.pdf}}%
    \put(0.64028625,0.03836366){\color[rgb]{0,0,0}\makebox(0,0)[t]{\lineheight{1.25}\smash{\begin{tabular}[t]{c}{\small $h_1 \textrm{x} w_1 \textrm{x} l_{\textrm{c1}}$}\end{tabular}}}}%
    \put(0,0){\includegraphics[width=\unitlength,page=8]{netw_arch.pdf}}%
    \put(0.62396299,0.22735849){\color[rgb]{0,0,0}\makebox(0,0)[t]{\lineheight{1.25}\smash{\begin{tabular}[t]{c}{\small conv1}\end{tabular}}}}%
    \put(0,0){\includegraphics[width=\unitlength,page=9]{netw_arch.pdf}}%
    \put(0.69688585,0.1990566){\color[rgb]{0,0,0}\makebox(0,0)[t]{\lineheight{1.25}\smash{\begin{tabular}[t]{c}{\small conv2}\end{tabular}}}}%
    \put(0,0){\includegraphics[width=\unitlength,page=10]{netw_arch.pdf}}%
    \put(0.83396226,0.14245283){\color[rgb]{0,0,0}\makebox(0,0)[t]{\lineheight{1.25}\smash{\begin{tabular}[t]{c}{\small fc3}\end{tabular}}}}%
    \put(0,0){\includegraphics[width=\unitlength,page=11]{netw_arch.pdf}}%
    \put(0.91226415,0.14245283){\color[rgb]{0,0,0}\makebox(0,0)[t]{\lineheight{1.25}\smash{\begin{tabular}[t]{c}{\small fc4}\end{tabular}}}}%
    \put(0,0){\includegraphics[width=\unitlength,page=12]{netw_arch.pdf}}%
    \put(0.71132076,0.05943396){\color[rgb]{0,0,0}\makebox(0,0)[t]{\lineheight{1.25}\smash{\begin{tabular}[t]{c}{\small $h_2 \textrm{x} w_2 \textrm{x} l_{\textrm{c2}}$}\end{tabular}}}}%
    \put(0,0){\includegraphics[width=\unitlength,page=13]{netw_arch.pdf}}%
    \put(0.96623019,0.14250377){\color[rgb]{0,0,0}\makebox(0,0)[t]{\lineheight{1.25}\smash{\begin{tabular}[t]{c}{\small output}\end{tabular}}}}%
    \put(0,0){\includegraphics[width=\unitlength,page=14]{netw_arch.pdf}}%
    \put(0.0946981,0.25754717){\color[rgb]{0,0,0}\makebox(0,0)[t]{\lineheight{1.25}\smash{\begin{tabular}[t]{c}Evetac Gel\end{tabular}}}}%
    \put(0,0){\includegraphics[width=\unitlength,page=15]{netw_arch.pdf}}%
    \put(0.38906038,0.25871887){\color[rgb]{0,0,0}\makebox(0,0)[t]{\lineheight{1.25}\smash{\begin{tabular}[t]{c}Per Dot Encoding\end{tabular}}}}%
    \put(0,0){\includegraphics[width=\unitlength,page=16]{netw_arch.pdf}}%
    \put(0.78455283,0.25865094){\color[rgb]{0,0,0}\makebox(0,0)[t]{\lineheight{1.25}\smash{\begin{tabular}[t]{c}Spatial Combination \& Prediction\end{tabular}}}}%
    \put(0,0){\includegraphics[width=\unitlength,page=17]{netw_arch.pdf}}%
    \put(0.83207547,0.08301887){\color[rgb]{0,0,0}\makebox(0,0)[t]{\lineheight{1.25}\smash{\begin{tabular}[t]{c}{\small $1 \textrm{x} l_{\textrm{fc3}}$}\end{tabular}}}}%
    \put(0,0){\includegraphics[width=\unitlength,page=18]{netw_arch.pdf}}%
    \put(0.91344413,0.08301887){\color[rgb]{0,0,0}\makebox(0,0)[t]{\lineheight{1.25}\smash{\begin{tabular}[t]{c}{\small $1 \textrm{x} l_{\textrm{fc4}}$}\end{tabular}}}}%
    \put(0,0){\includegraphics[width=\unitlength,page=19]{netw_arch.pdf}}%
    \put(0.96623019,0.10377358){\color[rgb]{0,0,0}\makebox(0,0)[t]{\lineheight{1.25}\smash{\begin{tabular}[t]{c}{\small $1$}\end{tabular}}}}%
    \put(0,0){\includegraphics[width=\unitlength,page=20]{netw_arch.pdf}}%
    \put(0.67735849,0.01415094){\color[rgb]{0,0,0}\makebox(0,0)[lt]{\lineheight{1.25}\smash{\begin{tabular}[t]{l}{\footnotesize max pooling}\end{tabular}}}}%
    \put(0,0){\includegraphics[width=\unitlength,page=21]{netw_arch.pdf}}%
    \put(0.4754717,0.01415094){\color[rgb]{0,0,0}\makebox(0,0)[lt]{\lineheight{1.25}\smash{\begin{tabular}[t]{l}{\footnotesize convolutional + ReLU}\end{tabular}}}}%
    \put(0,0){\includegraphics[width=\unitlength,page=22]{netw_arch.pdf}}%
    \put(0.82358491,0.01413396){\color[rgb]{0,0,0}\makebox(0,0)[lt]{\lineheight{1.25}\smash{\begin{tabular}[t]{l}{\footnotesize fully connected + sigmoid}\end{tabular}}}}%
    \put(0,0){\includegraphics[width=\unitlength,page=23]{netw_arch.pdf}}%
  \end{picture}%
\endgroup%

%% file: pictures/results/slip_off_train/slip_off_train.eps_tex
\begingroup%
  \makeatletter%
  \providecommand\color[2][]{%
    \errmessage{(Inkscape) Color is used for the text in Inkscape, but the package 'color.sty' is not loaded}%
    \renewcommand\color[2][]{}%
  }%
  \providecommand\transparent[1]{%
    \errmessage{(Inkscape) Transparency is used (non-zero) for the text in Inkscape, but the package 'transparent.sty' is not loaded}%
    \renewcommand\transparent[1]{}%
  }%
  \providecommand\rotatebox[2]{#2}%
  \newcommand*\fsize{\dimexpr\f@size pt\relax}%
  \newcommand*\lineheight[1]{\fontsize{\fsize}{#1\fsize}\selectfont}%
  \ifx\svgwidth\undefined%
    \setlength{\unitlength}{521.18928684bp}%
    \ifx\svgscale\undefined%
      \relax%
    \else%
      \setlength{\unitlength}{\unitlength * \real{\svgscale}}%
    \fi%
  \else%
    \setlength{\unitlength}{\svgwidth}%
  \fi%
  \global\let\svgwidth\undefined%
  \global\let\svgscale\undefined%
  \makeatother%
  \begin{picture}(1,0.2581582)%
    \lineheight{1}%
    \setlength\tabcolsep{0pt}%
    \put(0,0){\includegraphics[width=\unitlength]{slip_off_train.eps}}%
    \put(0.13928045,0.02549217){\color[rgb]{0,0,0}\makebox(0,0)[lt]{\lineheight{1.25}\smash{\begin{tabular}[t]{l}slip too early\end{tabular}}}}%
    \put(0.3476364,0.02549217){\color[rgb]{0,0,0}\makebox(0,0)[lt]{\lineheight{1.25}\smash{\begin{tabular}[t]{l}slip corr\end{tabular}}}}%
    \put(0.51439579,0.02549217){\color[rgb]{0,0,0}\makebox(0,0)[lt]{\lineheight{1.25}\smash{\begin{tabular}[t]{l}slip too late\end{tabular}}}}%
    \put(0.04760986,0.04620802){\color[rgb]{0,0,0}\makebox(0,0)[lt]{\lineheight{1.25}\smash{\begin{tabular}[t]{l}0.0\end{tabular}}}}%
    \put(0.04760986,0.06300677){\color[rgb]{0,0,0}\makebox(0,0)[lt]{\lineheight{1.25}\smash{\begin{tabular}[t]{l}0.1\end{tabular}}}}%
    \put(0.04760986,0.07980552){\color[rgb]{0,0,0}\makebox(0,0)[lt]{\lineheight{1.25}\smash{\begin{tabular}[t]{l}0.2\end{tabular}}}}%
    \put(0.04760986,0.09660427){\color[rgb]{0,0,0}\makebox(0,0)[lt]{\lineheight{1.25}\smash{\begin{tabular}[t]{l}0.3\end{tabular}}}}%
    \put(0.04760986,0.11340303){\color[rgb]{0,0,0}\makebox(0,0)[lt]{\lineheight{1.25}\smash{\begin{tabular}[t]{l}0.4\end{tabular}}}}%
    \put(0.04760986,0.13020178){\color[rgb]{0,0,0}\makebox(0,0)[lt]{\lineheight{1.25}\smash{\begin{tabular}[t]{l}0.5\end{tabular}}}}%
    \put(0.04760986,0.14700053){\color[rgb]{0,0,0}\makebox(0,0)[lt]{\lineheight{1.25}\smash{\begin{tabular}[t]{l}0.6\end{tabular}}}}%
    \put(0.04760986,0.16379928){\color[rgb]{0,0,0}\makebox(0,0)[lt]{\lineheight{1.25}\smash{\begin{tabular}[t]{l}0.7\end{tabular}}}}%
    \put(0.04760986,0.18059804){\color[rgb]{0,0,0}\makebox(0,0)[lt]{\lineheight{1.25}\smash{\begin{tabular}[t]{l}0.8\end{tabular}}}}%
    \put(0.04760986,0.19739678){\color[rgb]{0,0,0}\makebox(0,0)[lt]{\lineheight{1.25}\smash{\begin{tabular}[t]{l}0.9\end{tabular}}}}%
    \put(0.03594783,0.04675304){\color[rgb]{0,0,0}\rotatebox{90}{\makebox(0,0)[lt]{\lineheight{1.25}\smash{\begin{tabular}[t]{l}Ratio of Trajectories\end{tabular}}}}}%
    \put(0.72813036,0.04620802){\color[rgb]{0,0,0}\makebox(0,0)[lt]{\lineheight{1.25}\smash{\begin{tabular}[t]{l}0.2\end{tabular}}}}%
    \put(0.72813036,0.06984626){\color[rgb]{0,0,0}\makebox(0,0)[lt]{\lineheight{1.25}\smash{\begin{tabular}[t]{l}0.3\end{tabular}}}}%
    \put(0.72813036,0.0934845){\color[rgb]{0,0,0}\makebox(0,0)[lt]{\lineheight{1.25}\smash{\begin{tabular}[t]{l}0.4\end{tabular}}}}%
    \put(0.72813036,0.11712275){\color[rgb]{0,0,0}\makebox(0,0)[lt]{\lineheight{1.25}\smash{\begin{tabular}[t]{l}0.5\end{tabular}}}}%
    \put(0.72813036,0.14076099){\color[rgb]{0,0,0}\makebox(0,0)[lt]{\lineheight{1.25}\smash{\begin{tabular}[t]{l}0.6\end{tabular}}}}%
    \put(0.72813036,0.16439924){\color[rgb]{0,0,0}\makebox(0,0)[lt]{\lineheight{1.25}\smash{\begin{tabular}[t]{l}0.7\end{tabular}}}}%
    \put(0.72813036,0.18803748){\color[rgb]{0,0,0}\makebox(0,0)[lt]{\lineheight{1.25}\smash{\begin{tabular}[t]{l}0.8\end{tabular}}}}%
    \put(0.72813036,0.21167574){\color[rgb]{0,0,0}\makebox(0,0)[lt]{\lineheight{1.25}\smash{\begin{tabular}[t]{l}0.9\end{tabular}}}}%
    \put(0.71646832,0.09470528){\color[rgb]{0,0,0}\rotatebox{90}{\makebox(0,0)[lt]{\lineheight{1.25}\smash{\begin{tabular}[t]{l}F1 Score\end{tabular}}}}}%
    \put(0.38650082,0.00759077){\color[rgb]{0,0,0}\makebox(0,0)[t]{\lineheight{1.25}\smash{\begin{tabular}[t]{c}{\footnotesize $t^c_{s0} {-} \SI{50}{\ms} {\leq} t^m_{s0} \leq  t^c_{s0} {+} \SI{20}{\ms}$}\end{tabular}}}}%
    \put(0.20161108,0.00759078){\color[rgb]{0,0,0}\makebox(0,0)[t]{\lineheight{1.25}\smash{\begin{tabular}[t]{c}{\footnotesize $t^m_{s0} {<} t^c_{s0}{-} \SI{50}{\ms}$}\end{tabular}}}}%
    \put(0.57064417,0.00759078){\color[rgb]{0,0,0}\makebox(0,0)[t]{\lineheight{1.25}\smash{\begin{tabular}[t]{c}{\footnotesize $t^m_{s0} {>} t^c_{s0} {+} \SI{20}{\ms}$}\end{tabular}}}}%
    \put(0.06445703,0.24478488){\color[rgb]{0,0,0}\makebox(0,0)[lt]{\lineheight{1.25}\smash{\begin{tabular}[t]{l}no hist\end{tabular}}}}%
    \put(0.06445703,0.22814625){\color[rgb]{0,0,0}\makebox(0,0)[lt]{\lineheight{1.25}\smash{\begin{tabular}[t]{l}no hist img\end{tabular}}}}%
    \put(0.23039363,0.24478488){\color[rgb]{0,0,0}\makebox(0,0)[lt]{\lineheight{1.25}\smash{\begin{tabular}[t]{l}hist 10\end{tabular}}}}%
    \put(0.23039363,0.22814625){\color[rgb]{0,0,0}\makebox(0,0)[lt]{\lineheight{1.25}\smash{\begin{tabular}[t]{l}hist 10 img\end{tabular}}}}%
    \put(0.39680988,0.24478488){\color[rgb]{0,0,0}\makebox(0,0)[lt]{\lineheight{1.25}\smash{\begin{tabular}[t]{l}events only hist 10\end{tabular}}}}%
    \put(0.39680988,0.22814625){\color[rgb]{0,0,0}\makebox(0,0)[lt]{\lineheight{1.25}\smash{\begin{tabular}[t]{l}disp only hist 10\end{tabular}}}}%
    \put(0.64168253,0.24478488){\color[rgb]{0,0,0}\makebox(0,0)[lt]{\lineheight{1.25}\smash{\begin{tabular}[t]{l}hist 20\end{tabular}}}}%
    \put(0.64168253,0.22814625){\color[rgb]{0,0,0}\makebox(0,0)[lt]{\lineheight{1.25}\smash{\begin{tabular}[t]{l}down 5 (hist 50)\end{tabular}}}}%
    \put(0.83624955,0.24478488){\color[rgb]{0,0,0}\makebox(0,0)[lt]{\lineheight{1.25}\smash{\begin{tabular}[t]{l}fast slow (hist 50)\end{tabular}}}}%
  \end{picture}%
\endgroup%

%% file: pictures/results/traj_pred/traj_pred.pdf_tex
\begingroup%
  \makeatletter%
  \providecommand\color[2][]{%
    \errmessage{(Inkscape) Color is used for the text in Inkscape, but the package 'color.sty' is not loaded}%
    \renewcommand\color[2][]{}%
  }%
  \providecommand\transparent[1]{%
    \errmessage{(Inkscape) Transparency is used (non-zero) for the text in Inkscape, but the package 'transparent.sty' is not loaded}%
    \renewcommand\transparent[1]{}%
  }%
  \providecommand\rotatebox[2]{#2}%
  \newcommand*\fsize{\dimexpr\f@size pt\relax}%
  \newcommand*\lineheight[1]{\fontsize{\fsize}{#1\fsize}\selectfont}%
  \ifx\svgwidth\undefined%
    \setlength{\unitlength}{241.59117211bp}%
    \ifx\svgscale\undefined%
      \relax%
    \else%
      \setlength{\unitlength}{\unitlength * \real{\svgscale}}%
    \fi%
  \else%
    \setlength{\unitlength}{\svgwidth}%
  \fi%
  \global\let\svgwidth\undefined%
  \global\let\svgscale\undefined%
  \makeatother%
  \begin{picture}(1,0.60027362)%
    \lineheight{1}%
    \setlength\tabcolsep{0pt}%
    \put(0,0){\includegraphics[width=\unitlength,page=1]{traj_pred.pdf}}%
    \put(0.32151852,0.49582117){\color[rgb]{0,0,0}\makebox(0,0)[rt]{\lineheight{1.25}\smash{\begin{tabular}[t]{r}{\footnotesize ns}\end{tabular}}}}%
    \put(0,0){\includegraphics[width=\unitlength,page=2]{traj_pred.pdf}}%
    \put(0.32151852,0.54786028){\color[rgb]{0,0,0}\makebox(0,0)[rt]{\lineheight{1.25}\smash{\begin{tabular}[t]{r}{\footnotesize s}\end{tabular}}}}%
    \put(0.29612027,0.5208107){\color[rgb]{0,0,0}\makebox(0,0)[rt]{\lineheight{1.25}\smash{\begin{tabular}[t]{r}{\small no hist}\end{tabular}}}}%
    \put(0,0){\includegraphics[width=\unitlength,page=3]{traj_pred.pdf}}%
    \put(0.57991547,0.56728661){\color[rgb]{0,0,0}\makebox(0,0)[lt]{\lineheight{1.25}\smash{\begin{tabular}[t]{l}\small{GT}\end{tabular}}}}%
    \put(0,0){\includegraphics[width=\unitlength,page=4]{traj_pred.pdf}}%
    \put(0.83441309,0.56728661){\color[rgb]{0,0,0}\makebox(0,0)[lt]{\lineheight{1.25}\smash{\begin{tabular}[t]{l}\small{Model}\end{tabular}}}}%
    \put(0,0){\includegraphics[width=\unitlength,page=5]{traj_pred.pdf}}%
    \put(0.32151852,0.42522703){\color[rgb]{0,0,0}\makebox(0,0)[rt]{\lineheight{1.25}\smash{\begin{tabular}[t]{r}{\footnotesize ns}\end{tabular}}}}%
    \put(0,0){\includegraphics[width=\unitlength,page=6]{traj_pred.pdf}}%
    \put(0.32151852,0.47726614){\color[rgb]{0,0,0}\makebox(0,0)[rt]{\lineheight{1.25}\smash{\begin{tabular}[t]{r}{\footnotesize s}\end{tabular}}}}%
    \put(0.29723188,0.45021656){\color[rgb]{0,0,0}\makebox(0,0)[rt]{\lineheight{1.25}\smash{\begin{tabular}[t]{r}{\small hist 10}\end{tabular}}}}%
    \put(0,0){\includegraphics[width=\unitlength,page=7]{traj_pred.pdf}}%
    \put(0.32151852,0.35463288){\color[rgb]{0,0,0}\makebox(0,0)[rt]{\lineheight{1.25}\smash{\begin{tabular}[t]{r}{\footnotesize ns}\end{tabular}}}}%
    \put(0,0){\includegraphics[width=\unitlength,page=8]{traj_pred.pdf}}%
    \put(0.32151852,0.40667202){\color[rgb]{0,0,0}\makebox(0,0)[rt]{\lineheight{1.25}\smash{\begin{tabular}[t]{r}{\footnotesize s}\end{tabular}}}}%
    \put(0.29599587,0.38194426){\color[rgb]{0,0,0}\makebox(0,0)[rt]{\lineheight{1.25}\smash{\begin{tabular}[t]{r}{\small events only hist 10}\end{tabular}}}}%
    \put(0,0){\includegraphics[width=\unitlength,page=9]{traj_pred.pdf}}%
    \put(0.32151852,0.28403874){\color[rgb]{0,0,0}\makebox(0,0)[rt]{\lineheight{1.25}\smash{\begin{tabular}[t]{r}{\footnotesize ns}\end{tabular}}}}%
    \put(0,0){\includegraphics[width=\unitlength,page=10]{traj_pred.pdf}}%
    \put(0.32151852,0.33607788){\color[rgb]{0,0,0}\makebox(0,0)[rt]{\lineheight{1.25}\smash{\begin{tabular}[t]{r}{\footnotesize s}\end{tabular}}}}%
    \put(0.29803299,0.31326265){\color[rgb]{0,0,0}\makebox(0,0)[rt]{\lineheight{1.25}\smash{\begin{tabular}[t]{r}{\small disp only hist 10}\end{tabular}}}}%
    \put(0,0){\includegraphics[width=\unitlength,page=11]{traj_pred.pdf}}%
    \put(0.32151852,0.21344462){\color[rgb]{0,0,0}\makebox(0,0)[rt]{\lineheight{1.25}\smash{\begin{tabular}[t]{r}{\footnotesize ns}\end{tabular}}}}%
    \put(0,0){\includegraphics[width=\unitlength,page=12]{traj_pred.pdf}}%
    \put(0.32151852,0.26548374){\color[rgb]{0,0,0}\makebox(0,0)[rt]{\lineheight{1.25}\smash{\begin{tabular}[t]{r}{\footnotesize s}\end{tabular}}}}%
    \put(0.29723188,0.23843416){\color[rgb]{0,0,0}\makebox(0,0)[rt]{\lineheight{1.25}\smash{\begin{tabular}[t]{r}{\small hist 20}\end{tabular}}}}%
    \put(0,0){\includegraphics[width=\unitlength,page=13]{traj_pred.pdf}}%
    \put(0.32151852,0.14285048){\color[rgb]{0,0,0}\makebox(0,0)[rt]{\lineheight{1.25}\smash{\begin{tabular}[t]{r}{\footnotesize ns}\end{tabular}}}}%
    \put(0,0){\includegraphics[width=\unitlength,page=14]{traj_pred.pdf}}%
    \put(0.32151852,0.19488962){\color[rgb]{0,0,0}\makebox(0,0)[rt]{\lineheight{1.25}\smash{\begin{tabular}[t]{r}{\footnotesize s}\end{tabular}}}}%
    \put(0.30084561,0.16784002){\color[rgb]{0,0,0}\makebox(0,0)[rt]{\lineheight{1.25}\smash{\begin{tabular}[t]{r}{\small down 5}\end{tabular}}}}%
    \put(0,0){\includegraphics[width=\unitlength,page=15]{traj_pred.pdf}}%
    \put(0.36123502,0.03822485){\color[rgb]{0,0,0}\makebox(0,0)[lt]{\lineheight{1.25}\smash{\begin{tabular}[t]{l}0\end{tabular}}}}%
    \put(0,0){\includegraphics[width=\unitlength,page=16]{traj_pred.pdf}}%
    \put(0.55495256,0.03822485){\color[rgb]{0,0,0}\makebox(0,0)[lt]{\lineheight{1.25}\smash{\begin{tabular}[t]{l}20\end{tabular}}}}%
    \put(0,0){\includegraphics[width=\unitlength,page=17]{traj_pred.pdf}}%
    \put(0.75562267,0.03822485){\color[rgb]{0,0,0}\makebox(0,0)[lt]{\lineheight{1.25}\smash{\begin{tabular}[t]{l}40\end{tabular}}}}%
    \put(0,0){\includegraphics[width=\unitlength,page=18]{traj_pred.pdf}}%
    \put(0.95629279,0.03822485){\color[rgb]{0,0,0}\makebox(0,0)[lt]{\lineheight{1.25}\smash{\begin{tabular}[t]{l}60\end{tabular}}}}%
    \put(0.59688132,0.00336108){\color[rgb]{0,0,0}\makebox(0,0)[lt]{\lineheight{1.25}\smash{\begin{tabular}[t]{l}\small{Time [ms]}\end{tabular}}}}%
    \put(0,0){\includegraphics[width=\unitlength,page=19]{traj_pred.pdf}}%
    \put(0.32151852,0.07225635){\color[rgb]{0,0,0}\makebox(0,0)[rt]{\lineheight{1.25}\smash{\begin{tabular}[t]{r}{\footnotesize ns}\end{tabular}}}}%
    \put(0,0){\includegraphics[width=\unitlength,page=20]{traj_pred.pdf}}%
    \put(0.32151852,0.12429548){\color[rgb]{0,0,0}\makebox(0,0)[rt]{\lineheight{1.25}\smash{\begin{tabular}[t]{r}{\footnotesize s}\end{tabular}}}}%
    \put(0.30175511,0.09724589){\color[rgb]{0,0,0}\makebox(0,0)[rt]{\lineheight{1.25}\smash{\begin{tabular}[t]{r}{\small fast slow}\end{tabular}}}}%
    \put(0,0){\includegraphics[width=\unitlength,page=21]{traj_pred.pdf}}%
  \end{picture}%
\endgroup%

%% file: pictures/results/slip_off_unseen/unseen_objs.eps_tex
\begingroup%
  \makeatletter%
  \providecommand\color[2][]{%
    \errmessage{(Inkscape) Color is used for the text in Inkscape, but the package 'color.sty' is not loaded}%
    \renewcommand\color[2][]{}%
  }%
  \providecommand\transparent[1]{%
    \errmessage{(Inkscape) Transparency is used (non-zero) for the text in Inkscape, but the package 'transparent.sty' is not loaded}%
    \renewcommand\transparent[1]{}%
  }%
  \providecommand\rotatebox[2]{#2}%
  \newcommand*\fsize{\dimexpr\f@size pt\relax}%
  \newcommand*\lineheight[1]{\fontsize{\fsize}{#1\fsize}\selectfont}%
  \ifx\svgwidth\undefined%
    \setlength{\unitlength}{515.5199955bp}%
    \ifx\svgscale\undefined%
      \relax%
    \else%
      \setlength{\unitlength}{\unitlength * \real{\svgscale}}%
    \fi%
  \else%
    \setlength{\unitlength}{\svgwidth}%
  \fi%
  \global\let\svgwidth\undefined%
  \global\let\svgscale\undefined%
  \makeatother%
  \begin{picture}(1,0.28299169)%
    \lineheight{1}%
    \setlength\tabcolsep{0pt}%
    \put(0,0){\includegraphics[width=\unitlength]{unseen_objs.eps}}%
    \put(0.13496135,0.02577252){\color[rgb]{0,0,0}\makebox(0,0)[lt]{\lineheight{1.25}\smash{\begin{tabular}[t]{l}slip too early\end{tabular}}}}%
    \put(0.33577537,0.02577252){\color[rgb]{0,0,0}\makebox(0,0)[lt]{\lineheight{1.25}\smash{\begin{tabular}[t]{l}slip corr\end{tabular}}}}%
    \put(0.49453536,0.02577252){\color[rgb]{0,0,0}\makebox(0,0)[lt]{\lineheight{1.25}\smash{\begin{tabular}[t]{l}slip too late\end{tabular}}}}%
    \put(0.04813344,0.04671618){\color[rgb]{0,0,0}\makebox(0,0)[lt]{\lineheight{1.25}\smash{\begin{tabular}[t]{l}0.0\end{tabular}}}}%
    \put(0.04813344,0.06432538){\color[rgb]{0,0,0}\makebox(0,0)[lt]{\lineheight{1.25}\smash{\begin{tabular}[t]{l}0.1\end{tabular}}}}%
    \put(0.04813344,0.08193458){\color[rgb]{0,0,0}\makebox(0,0)[lt]{\lineheight{1.25}\smash{\begin{tabular}[t]{l}0.2\end{tabular}}}}%
    \put(0.04813344,0.09954378){\color[rgb]{0,0,0}\makebox(0,0)[lt]{\lineheight{1.25}\smash{\begin{tabular}[t]{l}0.3\end{tabular}}}}%
    \put(0.04813344,0.11715298){\color[rgb]{0,0,0}\makebox(0,0)[lt]{\lineheight{1.25}\smash{\begin{tabular}[t]{l}0.4\end{tabular}}}}%
    \put(0.04813344,0.13476218){\color[rgb]{0,0,0}\makebox(0,0)[lt]{\lineheight{1.25}\smash{\begin{tabular}[t]{l}0.5\end{tabular}}}}%
    \put(0.04813344,0.15237137){\color[rgb]{0,0,0}\makebox(0,0)[lt]{\lineheight{1.25}\smash{\begin{tabular}[t]{l}0.6\end{tabular}}}}%
    \put(0.04813344,0.16998057){\color[rgb]{0,0,0}\makebox(0,0)[lt]{\lineheight{1.25}\smash{\begin{tabular}[t]{l}0.7\end{tabular}}}}%
    \put(0.04813344,0.18758977){\color[rgb]{0,0,0}\makebox(0,0)[lt]{\lineheight{1.25}\smash{\begin{tabular}[t]{l}0.8\end{tabular}}}}%
    \put(0.04813344,0.20519898){\color[rgb]{0,0,0}\makebox(0,0)[lt]{\lineheight{1.25}\smash{\begin{tabular}[t]{l}0.9\end{tabular}}}}%
    \put(0.03634316,0.04144783){\color[rgb]{0,0,0}\rotatebox{90}{\makebox(0,0)[lt]{\lineheight{1.25}\smash{\begin{tabular}[t]{l}Ratio of Trajectories\end{tabular}}}}}%
    \put(0.7047696,0.04671618){\color[rgb]{0,0,0}\makebox(0,0)[lt]{\lineheight{1.25}\smash{\begin{tabular}[t]{l}0.5\end{tabular}}}}%
    \put(0.7047696,0.10247864){\color[rgb]{0,0,0}\makebox(0,0)[lt]{\lineheight{1.25}\smash{\begin{tabular}[t]{l}0.6\end{tabular}}}}%
    \put(0.7047696,0.15824111){\color[rgb]{0,0,0}\makebox(0,0)[lt]{\lineheight{1.25}\smash{\begin{tabular}[t]{l}0.7\end{tabular}}}}%
    \put(0.7047696,0.21400358){\color[rgb]{0,0,0}\makebox(0,0)[lt]{\lineheight{1.25}\smash{\begin{tabular}[t]{l}0.8\end{tabular}}}}%
    \put(0.69297933,0.09574676){\color[rgb]{0,0,0}\rotatebox{90}{\makebox(0,0)[lt]{\lineheight{1.25}\smash{\begin{tabular}[t]{l}F1 Score\end{tabular}}}}}%
    \put(0.90537089,0.02454535){\color[rgb]{0,0,0}\makebox(0,0)[t]{\lineheight{1.25}\smash{\begin{tabular}[t]{c}{\footnotesize $\Delta T_{\textrm{pred}}$}\end{tabular}}}}%
    \put(0.90537089,0.00378457){\color[rgb]{0,0,0}\makebox(0,0)[t]{\lineheight{1.25}\smash{\begin{tabular}[t]{c}{\footnotesize $\SI{20}{\ms}$}\end{tabular}}}}%
    \put(0.84321447,0.02454535){\color[rgb]{0,0,0}\makebox(0,0)[t]{\lineheight{1.25}\smash{\begin{tabular}[t]{c}{\footnotesize $\Delta T_{\textrm{pred}}$}\end{tabular}}}}%
    \put(0.84321447,0.00378457){\color[rgb]{0,0,0}\makebox(0,0)[t]{\lineheight{1.25}\smash{\begin{tabular}[t]{c}{\footnotesize $\SI{10}{\ms}$}\end{tabular}}}}%
    \put(0.78105807,0.02454535){\color[rgb]{0,0,0}\makebox(0,0)[t]{\lineheight{1.25}\smash{\begin{tabular}[t]{c}{\footnotesize $\Delta T_{\textrm{pred}}$}\end{tabular}}}}%
    \put(0.78105807,0.00378457){\color[rgb]{0,0,0}\makebox(0,0)[t]{\lineheight{1.25}\smash{\begin{tabular}[t]{c}{\footnotesize $\SI{0}{\ms}$}\end{tabular}}}}%
    \put(0.06983246,0.2677915){\color[rgb]{0,0,0}\makebox(0,0)[lt]{\lineheight{1.25}\smash{\begin{tabular}[t]{l}{\small hist 20}\end{tabular}}}}%
    \put(0.06983246,0.25096989){\color[rgb]{0,0,0}\makebox(0,0)[lt]{\lineheight{1.25}\smash{\begin{tabular}[t]{l}{\small hist 20 $\Delta T_{\textrm{pred}}{=}\SI{10}{\ms}$}\end{tabular}}}}%
    \put(0.06983246,0.22996561){\color[rgb]{0,0,0}\makebox(0,0)[lt]{\lineheight{1.25}\smash{\begin{tabular}[t]{l}{\small hist 20 $\Delta T_{\textrm{pred}}{=}\SI{20}{\ms}$}\end{tabular}}}}%
    \put(0.33373469,0.26822719){\color[rgb]{0,0,0}\makebox(0,0)[lt]{\lineheight{1.25}\smash{\begin{tabular}[t]{l}{\small hist 10 img}\end{tabular}}}}%
    \put(0.33373469,0.25096989){\color[rgb]{0,0,0}\makebox(0,0)[lt]{\lineheight{1.25}\smash{\begin{tabular}[t]{l}{\small hist 10 img $\Delta T_{\textrm{pred}}{=}\SI{10}{\ms}$}\end{tabular}}}}%
    \put(0.33373469,0.22996561){\color[rgb]{0,0,0}\makebox(0,0)[lt]{\lineheight{1.25}\smash{\begin{tabular}[t]{l}{\small hist 10 img $\Delta T_{\textrm{pred}}{=}\SI{20}{\ms}$}\end{tabular}}}}%
    \put(0.59228888,0.2677915){\color[rgb]{0,0,0}\makebox(0,0)[lt]{\lineheight{1.25}\smash{\begin{tabular}[t]{l}{\small fast slow}\end{tabular}}}}%
    \put(0.59228888,0.25096989){\color[rgb]{0,0,0}\makebox(0,0)[lt]{\lineheight{1.25}\smash{\begin{tabular}[t]{l}{\small fast slow $\Delta T_{\textrm{pred}}{=}\SI{10}{\ms}$}\end{tabular}}}}%
    \put(0.5935646,0.22996561){\color[rgb]{0,0,0}\makebox(0,0)[lt]{\lineheight{1.25}\smash{\begin{tabular}[t]{l}{\small fast slow $\Delta T_{\textrm{pred}}{=}\SI{20}{\ms}$}\end{tabular}}}}%
    \put(0.3750672,0.00634941){\color[rgb]{0,0,0}\makebox(0,0)[t]{\lineheight{1.25}\smash{\begin{tabular}[t]{c}{\footnotesize $t^c_{s0} {-} \SI{50}{\ms} {\leq} t^m_{s0} \leq  t^c_{s0} {+} \SI{20}{\ms}$}\end{tabular}}}}%
    \put(0.19797744,0.00634943){\color[rgb]{0,0,0}\makebox(0,0)[t]{\lineheight{1.25}\smash{\begin{tabular}[t]{c}{\footnotesize $t^m_{s0} {<} t^c_{s0}{-} \SI{50}{\ms}$}\end{tabular}}}}%
    \put(0.5514023,0.00634943){\color[rgb]{0,0,0}\makebox(0,0)[t]{\lineheight{1.25}\smash{\begin{tabular}[t]{c}{\footnotesize $t^m_{s0} {>} t^c_{s0} {+} \SI{20}{\ms}$}\end{tabular}}}}%
  \end{picture}%
\endgroup%

%% file: pictures/results/slip_catch/slip_times/slip_timing_cdf.pdf_tex
\begingroup%
  \makeatletter%
  \providecommand\color[2][]{%
    \errmessage{(Inkscape) Color is used for the text in Inkscape, but the package 'color.sty' is not loaded}%
    \renewcommand\color[2][]{}%
  }%
  \providecommand\transparent[1]{%
    \errmessage{(Inkscape) Transparency is used (non-zero) for the text in Inkscape, but the package 'transparent.sty' is not loaded}%
    \renewcommand\transparent[1]{}%
  }%
  \providecommand\rotatebox[2]{#2}%
  \newcommand*\fsize{\dimexpr\f@size pt\relax}%
  \newcommand*\lineheight[1]{\fontsize{\fsize}{#1\fsize}\selectfont}%
  \ifx\svgwidth\undefined%
    \setlength{\unitlength}{255.19830178bp}%
    \ifx\svgscale\undefined%
      \relax%
    \else%
      \setlength{\unitlength}{\unitlength * \real{\svgscale}}%
    \fi%
  \else%
    \setlength{\unitlength}{\svgwidth}%
  \fi%
  \global\let\svgwidth\undefined%
  \global\let\svgscale\undefined%
  \makeatother%
  \begin{picture}(1,0.68440943)%
    \lineheight{1}%
    \setlength\tabcolsep{0pt}%
    \put(0,0){\includegraphics[width=\unitlength,page=1]{slip_timing_cdf.pdf}}%
    \put(0.15555784,0.60594946){\color[rgb]{0,0,0}\makebox(0,0)[lt]{\lineheight{1.25}\smash{\begin{tabular}[t]{l}{\scriptsize0}\end{tabular}}}}%
    \put(0,0){\includegraphics[width=\unitlength,page=2]{slip_timing_cdf.pdf}}%
    \put(0.15555784,0.64984427){\color[rgb]{0,0,0}\makebox(0,0)[lt]{\lineheight{1.25}\smash{\begin{tabular}[t]{l}{\scriptsize1}\end{tabular}}}}%
    \put(0.00487845,0.62910161){\color[rgb]{0,0,0}\makebox(0,0)[lt]{\lineheight{1.25}\smash{\begin{tabular}[t]{l}{\footnotesize Objs 10-17}\end{tabular}}}}%
    \put(0,0){\includegraphics[width=\unitlength,page=3]{slip_timing_cdf.pdf}}%
    \put(0.15555784,0.54323596){\color[rgb]{0,0,0}\makebox(0,0)[lt]{\lineheight{1.25}\smash{\begin{tabular}[t]{l}{\scriptsize0}\end{tabular}}}}%
    \put(0,0){\includegraphics[width=\unitlength,page=4]{slip_timing_cdf.pdf}}%
    \put(0.15555784,0.58125302){\color[rgb]{0,0,0}\makebox(0,0)[lt]{\lineheight{1.25}\smash{\begin{tabular}[t]{l}{\scriptsize1}\end{tabular}}}}%
    \put(0.00487845,0.56638815){\color[rgb]{0,0,0}\makebox(0,0)[lt]{\lineheight{1.25}\smash{\begin{tabular}[t]{l}{\footnotesize Obj 10}\end{tabular}}}}%
    \put(0,0){\includegraphics[width=\unitlength,page=5]{slip_timing_cdf.pdf}}%
    \put(0.15555784,0.48052245){\color[rgb]{0,0,0}\makebox(0,0)[lt]{\lineheight{1.25}\smash{\begin{tabular}[t]{l}{\scriptsize0}\end{tabular}}}}%
    \put(0,0){\includegraphics[width=\unitlength,page=6]{slip_timing_cdf.pdf}}%
    \put(0.15555784,0.51853951){\color[rgb]{0,0,0}\makebox(0,0)[lt]{\lineheight{1.25}\smash{\begin{tabular}[t]{l}{\scriptsize1}\end{tabular}}}}%
    \put(0.00487845,0.50367464){\color[rgb]{0,0,0}\makebox(0,0)[lt]{\lineheight{1.25}\smash{\begin{tabular}[t]{l}{\footnotesize Obj 11}\end{tabular}}}}%
    \put(0,0){\includegraphics[width=\unitlength,page=7]{slip_timing_cdf.pdf}}%
    \put(0.15555784,0.41780897){\color[rgb]{0,0,0}\makebox(0,0)[lt]{\lineheight{1.25}\smash{\begin{tabular}[t]{l}{\scriptsize0}\end{tabular}}}}%
    \put(0,0){\includegraphics[width=\unitlength,page=8]{slip_timing_cdf.pdf}}%
    \put(0.15555784,0.45582601){\color[rgb]{0,0,0}\makebox(0,0)[lt]{\lineheight{1.25}\smash{\begin{tabular}[t]{l}{\scriptsize1}\end{tabular}}}}%
    \put(0.00487845,0.44096114){\color[rgb]{0,0,0}\makebox(0,0)[lt]{\lineheight{1.25}\smash{\begin{tabular}[t]{l}{\footnotesize Obj 12}\end{tabular}}}}%
    \put(0,0){\includegraphics[width=\unitlength,page=9]{slip_timing_cdf.pdf}}%
    \put(0.15555784,0.35509548){\color[rgb]{0,0,0}\makebox(0,0)[lt]{\lineheight{1.25}\smash{\begin{tabular}[t]{l}{\scriptsize0}\end{tabular}}}}%
    \put(0,0){\includegraphics[width=\unitlength,page=10]{slip_timing_cdf.pdf}}%
    \put(0.15555784,0.39311253){\color[rgb]{0,0,0}\makebox(0,0)[lt]{\lineheight{1.25}\smash{\begin{tabular}[t]{l}{\scriptsize1}\end{tabular}}}}%
    \put(0.00487845,0.37824766){\color[rgb]{0,0,0}\makebox(0,0)[lt]{\lineheight{1.25}\smash{\begin{tabular}[t]{l}{\footnotesize Obj 13}\end{tabular}}}}%
    \put(0,0){\includegraphics[width=\unitlength,page=11]{slip_timing_cdf.pdf}}%
    \put(0.15555784,0.29238198){\color[rgb]{0,0,0}\makebox(0,0)[lt]{\lineheight{1.25}\smash{\begin{tabular}[t]{l}{\scriptsize0}\end{tabular}}}}%
    \put(0,0){\includegraphics[width=\unitlength,page=12]{slip_timing_cdf.pdf}}%
    \put(0.15555784,0.33039904){\color[rgb]{0,0,0}\makebox(0,0)[lt]{\lineheight{1.25}\smash{\begin{tabular}[t]{l}{\scriptsize1}\end{tabular}}}}%
    \put(0.00487845,0.31553416){\color[rgb]{0,0,0}\makebox(0,0)[lt]{\lineheight{1.25}\smash{\begin{tabular}[t]{l}{\footnotesize Obj 14}\end{tabular}}}}%
    \put(0,0){\includegraphics[width=\unitlength,page=13]{slip_timing_cdf.pdf}}%
    \put(0.15555784,0.22966849){\color[rgb]{0,0,0}\makebox(0,0)[lt]{\lineheight{1.25}\smash{\begin{tabular}[t]{l}{\scriptsize0}\end{tabular}}}}%
    \put(0,0){\includegraphics[width=\unitlength,page=14]{slip_timing_cdf.pdf}}%
    \put(0.15555784,0.26768554){\color[rgb]{0,0,0}\makebox(0,0)[lt]{\lineheight{1.25}\smash{\begin{tabular}[t]{l}{\scriptsize1}\end{tabular}}}}%
    \put(0.00487845,0.25282067){\color[rgb]{0,0,0}\makebox(0,0)[lt]{\lineheight{1.25}\smash{\begin{tabular}[t]{l}{\footnotesize Obj 15}\end{tabular}}}}%
    \put(0,0){\includegraphics[width=\unitlength,page=15]{slip_timing_cdf.pdf}}%
    \put(0.15555784,0.166955){\color[rgb]{0,0,0}\makebox(0,0)[lt]{\lineheight{1.25}\smash{\begin{tabular}[t]{l}{\scriptsize0}\end{tabular}}}}%
    \put(0,0){\includegraphics[width=\unitlength,page=16]{slip_timing_cdf.pdf}}%
    \put(0.15555784,0.20497205){\color[rgb]{0,0,0}\makebox(0,0)[lt]{\lineheight{1.25}\smash{\begin{tabular}[t]{l}{\scriptsize1}\end{tabular}}}}%
    \put(0.00487845,0.19010718){\color[rgb]{0,0,0}\makebox(0,0)[lt]{\lineheight{1.25}\smash{\begin{tabular}[t]{l}{\footnotesize Obj 16}\end{tabular}}}}%
    \put(0,0){\includegraphics[width=\unitlength,page=17]{slip_timing_cdf.pdf}}%
    \put(0.23615024,0.05356731){\color[rgb]{0,0,0}\makebox(0,0)[lt]{\lineheight{1.25}\smash{\begin{tabular}[t]{l}−60\end{tabular}}}}%
    \put(0,0){\includegraphics[width=\unitlength,page=18]{slip_timing_cdf.pdf}}%
    \put(0.37529337,0.05356731){\color[rgb]{0,0,0}\makebox(0,0)[lt]{\lineheight{1.25}\smash{\begin{tabular}[t]{l}−40\end{tabular}}}}%
    \put(0,0){\includegraphics[width=\unitlength,page=19]{slip_timing_cdf.pdf}}%
    \put(0.51443649,0.05356731){\color[rgb]{0,0,0}\makebox(0,0)[lt]{\lineheight{1.25}\smash{\begin{tabular}[t]{l}−20\end{tabular}}}}%
    \put(0,0){\includegraphics[width=\unitlength,page=20]{slip_timing_cdf.pdf}}%
    \put(0.68244811,0.05356731){\color[rgb]{0,0,0}\makebox(0,0)[lt]{\lineheight{1.25}\smash{\begin{tabular}[t]{l}0\end{tabular}}}}%
    \put(0,0){\includegraphics[width=\unitlength,page=21]{slip_timing_cdf.pdf}}%
    \put(0.80913155,0.05356731){\color[rgb]{0,0,0}\makebox(0,0)[lt]{\lineheight{1.25}\smash{\begin{tabular}[t]{l}20\end{tabular}}}}%
    \put(0,0){\includegraphics[width=\unitlength,page=22]{slip_timing_cdf.pdf}}%
    \put(0.94827467,0.05356731){\color[rgb]{0,0,0}\makebox(0,0)[lt]{\lineheight{1.25}\smash{\begin{tabular}[t]{l}40\end{tabular}}}}%
    \put(0,0){\includegraphics[width=\unitlength,page=23]{slip_timing_cdf.pdf}}%
    \put(0.15555784,0.09836372){\color[rgb]{0,0,0}\makebox(0,0)[lt]{\lineheight{1.25}\smash{\begin{tabular}[t]{l}{\scriptsize0}\end{tabular}}}}%
    \put(0,0){\includegraphics[width=\unitlength,page=24]{slip_timing_cdf.pdf}}%
    \put(0.15555784,0.14225856){\color[rgb]{0,0,0}\makebox(0,0)[lt]{\lineheight{1.25}\smash{\begin{tabular}[t]{l}{\scriptsize1}\end{tabular}}}}%
    \put(0.00487845,0.12739368){\color[rgb]{0,0,0}\makebox(0,0)[lt]{\lineheight{1.25}\smash{\begin{tabular}[t]{l}{\footnotesize Obj 17}\end{tabular}}}}%
    \put(0,0){\includegraphics[width=\unitlength,page=25]{slip_timing_cdf.pdf}}%
    \put(0.57418965,0.01511937){\color[rgb]{0,0,0}\makebox(0,0)[t]{\lineheight{1.25}\smash{\begin{tabular}[t]{c}{\footnotesize $t^m_{s0} {-} t^c_{s0} \left[\SI{}{\ms}\right]$}\end{tabular}}}}%
  \end{picture}%
\endgroup%

%% file: pictures/results/slip_catch/normal/timing.pdf_tex
\begingroup%
  \makeatletter%
  \providecommand\color[2][]{%
    \errmessage{(Inkscape) Color is used for the text in Inkscape, but the package 'color.sty' is not loaded}%
    \renewcommand\color[2][]{}%
  }%
  \providecommand\transparent[1]{%
    \errmessage{(Inkscape) Transparency is used (non-zero) for the text in Inkscape, but the package 'transparent.sty' is not loaded}%
    \renewcommand\transparent[1]{}%
  }%
  \providecommand\rotatebox[2]{#2}%
  \newcommand*\fsize{\dimexpr\f@size pt\relax}%
  \newcommand*\lineheight[1]{\fontsize{\fsize}{#1\fsize}\selectfont}%
  \ifx\svgwidth\undefined%
    \setlength{\unitlength}{516.01831631bp}%
    \ifx\svgscale\undefined%
      \relax%
    \else%
      \setlength{\unitlength}{\unitlength * \real{\svgscale}}%
    \fi%
  \else%
    \setlength{\unitlength}{\svgwidth}%
  \fi%
  \global\let\svgwidth\undefined%
  \global\let\svgscale\undefined%
  \makeatother%
  \begin{picture}(1,0.42071569)%
    \lineheight{1}%
    \setlength\tabcolsep{0pt}%
    \put(0,0){\includegraphics[width=\unitlength,page=1]{timing.pdf}}%
    \put(0.58884024,0.40324273){\color[rgb]{0,0,0}\makebox(0,0)[lt]{\lineheight{1.25}\smash{\begin{tabular}[t]{l}Overall\end{tabular}}}}%
    \put(0,0){\includegraphics[width=\unitlength,page=2]{timing.pdf}}%
    \put(0.75114068,0.40324273){\color[rgb]{0,0,0}\makebox(0,0)[lt]{\lineheight{1.25}\smash{\begin{tabular}[t]{l}Lift\end{tabular}}}}%
    \put(0,0){\includegraphics[width=\unitlength,page=3]{timing.pdf}}%
    \put(0.8745314,0.40324273){\color[rgb]{0,0,0}\makebox(0,0)[lt]{\lineheight{1.25}\smash{\begin{tabular}[t]{l}Balance\end{tabular}}}}%
    \put(0,0){\includegraphics[width=\unitlength,page=4]{timing.pdf}}%
    \put(0.0665843,0.34875088){\color[rgb]{0,0,0}\makebox(0,0)[lt]{\lineheight{1.25}\smash{\begin{tabular}[t]{l}0.2\end{tabular}}}}%
    \put(0,0){\includegraphics[width=\unitlength,page=5]{timing.pdf}}%
    \put(0.0665843,0.36928846){\color[rgb]{0,0,0}\makebox(0,0)[lt]{\lineheight{1.25}\smash{\begin{tabular}[t]{l}0.6\end{tabular}}}}%
    \put(0,0){\includegraphics[width=\unitlength,page=6]{timing.pdf}}%
    \put(0.0665843,0.38982605){\color[rgb]{0,0,0}\makebox(0,0)[lt]{\lineheight{1.25}\smash{\begin{tabular}[t]{l}1.0\end{tabular}}}}%
    \put(0.03311892,0.34324771){\color[rgb]{0,0,0}\rotatebox{90}{\makebox(0,0)[lt]{\lineheight{1.25}\smash{\begin{tabular}[t]{l}Success \end{tabular}}}}}%
    \put(0.0548054,0.36875917){\color[rgb]{0,0,0}\rotatebox{90}{\makebox(0,0)[lt]{\lineheight{1.25}\smash{\begin{tabular}[t]{l}rate\end{tabular}}}}}%
    \put(0,0){\includegraphics[width=\unitlength,page=7]{timing.pdf}}%
    \put(0.05697018,0.19719517){\color[rgb]{0,0,0}\makebox(0,0)[lt]{\lineheight{1.25}\smash{\begin{tabular}[t]{l}−30\end{tabular}}}}%
    \put(0,0){\includegraphics[width=\unitlength,page=8]{timing.pdf}}%
    \put(0.05697018,0.21246484){\color[rgb]{0,0,0}\makebox(0,0)[lt]{\lineheight{1.25}\smash{\begin{tabular}[t]{l}−25\end{tabular}}}}%
    \put(0,0){\includegraphics[width=\unitlength,page=9]{timing.pdf}}%
    \put(0.05697018,0.2277345){\color[rgb]{0,0,0}\makebox(0,0)[lt]{\lineheight{1.25}\smash{\begin{tabular}[t]{l}−20\end{tabular}}}}%
    \put(0,0){\includegraphics[width=\unitlength,page=10]{timing.pdf}}%
    \put(0.05697018,0.24300416){\color[rgb]{0,0,0}\makebox(0,0)[lt]{\lineheight{1.25}\smash{\begin{tabular}[t]{l}−15\end{tabular}}}}%
    \put(0,0){\includegraphics[width=\unitlength,page=11]{timing.pdf}}%
    \put(0.05697018,0.25827383){\color[rgb]{0,0,0}\makebox(0,0)[lt]{\lineheight{1.25}\smash{\begin{tabular}[t]{l}−10\end{tabular}}}}%
    \put(0,0){\includegraphics[width=\unitlength,page=12]{timing.pdf}}%
    \put(0.06929411,0.27354348){\color[rgb]{0,0,0}\makebox(0,0)[lt]{\lineheight{1.25}\smash{\begin{tabular}[t]{l}−5\end{tabular}}}}%
    \put(0,0){\includegraphics[width=\unitlength,page=13]{timing.pdf}}%
    \put(0.08552415,0.28881315){\color[rgb]{0,0,0}\makebox(0,0)[lt]{\lineheight{1.25}\smash{\begin{tabular}[t]{l}0\end{tabular}}}}%
    \put(0,0){\includegraphics[width=\unitlength,page=14]{timing.pdf}}%
    \put(0.08552415,0.30408282){\color[rgb]{0,0,0}\makebox(0,0)[lt]{\lineheight{1.25}\smash{\begin{tabular}[t]{l}5\end{tabular}}}}%
    \put(0,0){\includegraphics[width=\unitlength,page=15]{timing.pdf}}%
    \put(0.07320022,0.31935247){\color[rgb]{0,0,0}\makebox(0,0)[lt]{\lineheight{1.25}\smash{\begin{tabular}[t]{l}10\end{tabular}}}}%
    \put(0.0235048,0.23700474){\color[rgb]{0,0,0}\rotatebox{90}{\makebox(0,0)[lt]{\lineheight{1.25}\smash{\begin{tabular}[t]{l}Delta\end{tabular}}}}}%
    \put(0.04519128,0.18863315){\color[rgb]{0,0,0}\rotatebox{90}{\makebox(0,0)[lt]{\lineheight{1.25}\smash{\begin{tabular}[t]{l}Gripper Opening\end{tabular}}}}}%
    \put(0,0){\includegraphics[width=\unitlength,page=16]{timing.pdf}}%
    \put(0.048189,0.09988522){\color[rgb]{0,0,0}\makebox(0,0)[lt]{\lineheight{1.25}\smash{\begin{tabular}[t]{l}0.000\end{tabular}}}}%
    \put(0,0){\includegraphics[width=\unitlength,page=17]{timing.pdf}}%
    \put(0.048189,0.11753836){\color[rgb]{0,0,0}\makebox(0,0)[lt]{\lineheight{1.25}\smash{\begin{tabular}[t]{l}0.005\end{tabular}}}}%
    \put(0,0){\includegraphics[width=\unitlength,page=18]{timing.pdf}}%
    \put(0.048189,0.13519151){\color[rgb]{0,0,0}\makebox(0,0)[lt]{\lineheight{1.25}\smash{\begin{tabular}[t]{l}0.010\end{tabular}}}}%
    \put(0,0){\includegraphics[width=\unitlength,page=19]{timing.pdf}}%
    \put(0.048189,0.15284465){\color[rgb]{0,0,0}\makebox(0,0)[lt]{\lineheight{1.25}\smash{\begin{tabular}[t]{l}0.015\end{tabular}}}}%
    \put(0.01472362,0.09533565){\color[rgb]{0,0,0}\rotatebox{90}{\makebox(0,0)[lt]{\lineheight{1.25}\smash{\begin{tabular}[t]{l}Object\end{tabular}}}}}%
    \put(0.0364101,0.07978751){\color[rgb]{0,0,0}\rotatebox{90}{\makebox(0,0)[lt]{\lineheight{1.25}\smash{\begin{tabular}[t]{l}Moved [m]\end{tabular}}}}}%
    \put(0,0){\includegraphics[width=\unitlength,page=20]{timing.pdf}}%
    \put(0.30213796,0.03084241){\color[rgb]{0,0,0}\makebox(0,0)[t]{\lineheight{1.25}\smash{\begin{tabular}[t]{c}{\footnotesize Obj 10}\end{tabular}}}}%
    \put(0,0){\includegraphics[width=\unitlength,page=21]{timing.pdf}}%
    \put(0.36785944,0.03084241){\color[rgb]{0,0,0}\makebox(0,0)[t]{\lineheight{1.25}\smash{\begin{tabular}[t]{c}{\footnotesize Obj 11}\end{tabular}}}}%
    \put(0,0){\includegraphics[width=\unitlength,page=22]{timing.pdf}}%
    \put(0.6964668,0.03084239){\color[rgb]{0,0,0}\makebox(0,0)[t]{\lineheight{1.25}\smash{\begin{tabular}[t]{c}{\footnotesize Obj 14}\end{tabular}}}}%
    \put(0,0){\includegraphics[width=\unitlength,page=23]{timing.pdf}}%
    \put(0.76218817,0.03084239){\color[rgb]{0,0,0}\makebox(0,0)[t]{\lineheight{1.25}\smash{\begin{tabular}[t]{c}{\footnotesize Obj 14}\end{tabular}}}}%
    \put(0.76218819,0.01340104){\color[rgb]{0,0,0}\transparent{0.95999998}\makebox(0,0)[t]{\lineheight{1.25}\smash{\begin{tabular}[t]{c}{\scriptsize Higher}\end{tabular}}}}%
    \put(0.76218819,-0.00116711){\color[rgb]{0,0,0}\transparent{0.95999998}\makebox(0,0)[t]{\lineheight{1.25}\smash{\begin{tabular}[t]{c}{\scriptsize Gains}\end{tabular}}}}%
    \put(0.82646303,0.03084239){\color[rgb]{0,0,0}\makebox(0,0)[t]{\lineheight{1.25}\smash{\begin{tabular}[t]{c}{\footnotesize Obj 19}\end{tabular}}}}%
    \put(0.82646304,0.01340104){\color[rgb]{0,0,0}\transparent{0.95999998}\makebox(0,0)[t]{\lineheight{1.25}\smash{\begin{tabular}[t]{c}{\scriptsize Higher}\end{tabular}}}}%
    \put(0.82646304,-0.00116711){\color[rgb]{0,0,0}\transparent{0.95999998}\makebox(0,0)[t]{\lineheight{1.25}\smash{\begin{tabular}[t]{c}{\scriptsize Gains}\end{tabular}}}}%
    \put(0.8921845,0.03084239){\color[rgb]{0,0,0}\makebox(0,0)[t]{\lineheight{1.25}\smash{\begin{tabular}[t]{c}{\footnotesize Obj 20}\end{tabular}}}}%
    \put(0.89218451,0.01340104){\color[rgb]{0,0,0}\transparent{0.95999998}\makebox(0,0)[t]{\lineheight{1.25}\smash{\begin{tabular}[t]{c}{\scriptsize Higher}\end{tabular}}}}%
    \put(0.89218451,-0.00116711){\color[rgb]{0,0,0}\transparent{0.95999998}\makebox(0,0)[t]{\lineheight{1.25}\smash{\begin{tabular}[t]{c}{\scriptsize Gains}\end{tabular}}}}%
    \put(0.17069506,0.0308424){\color[rgb]{0,0,0}\makebox(0,0)[t]{\lineheight{1.25}\smash{\begin{tabular}[t]{c}{\footnotesize Obj 18}\end{tabular}}}}%
    \put(0.23641653,0.0308424){\color[rgb]{0,0,0}\transparent{0.95999998}\makebox(0,0)[t]{\lineheight{1.25}\smash{\begin{tabular}[t]{c}{\footnotesize Obj 17}\end{tabular}}}}%
    \put(0.43358091,0.0308424){\color[rgb]{0,0,0}\makebox(0,0)[t]{\lineheight{1.25}\smash{\begin{tabular}[t]{c}{\footnotesize Obj 12}\end{tabular}}}}%
    \put(0.49930239,0.03084241){\color[rgb]{0,0,0}\makebox(0,0)[t]{\lineheight{1.25}\smash{\begin{tabular}[t]{c}{\footnotesize Obj 13}\end{tabular}}}}%
    \put(0.56502384,0.03084239){\color[rgb]{0,0,0}\makebox(0,0)[t]{\lineheight{1.25}\smash{\begin{tabular}[t]{c}{\footnotesize Obj 15}\end{tabular}}}}%
    \put(0.63074533,0.03084238){\color[rgb]{0,0,0}\makebox(0,0)[t]{\lineheight{1.25}\smash{\begin{tabular}[t]{c}{\footnotesize Obj 16}\end{tabular}}}}%
    \put(0,0){\includegraphics[width=\unitlength,page=24]{timing.pdf}}%
  \end{picture}%
\endgroup%

%% file: pictures/results/slip_catch/ood/sideways.pdf_tex
\begingroup%
  \makeatletter%
  \providecommand\color[2][]{%
    \errmessage{(Inkscape) Color is used for the text in Inkscape, but the package 'color.sty' is not loaded}%
    \renewcommand\color[2][]{}%
  }%
  \providecommand\transparent[1]{%
    \errmessage{(Inkscape) Transparency is used (non-zero) for the text in Inkscape, but the package 'transparent.sty' is not loaded}%
    \renewcommand\transparent[1]{}%
  }%
  \providecommand\rotatebox[2]{#2}%
  \newcommand*\fsize{\dimexpr\f@size pt\relax}%
  \newcommand*\lineheight[1]{\fontsize{\fsize}{#1\fsize}\selectfont}%
  \ifx\svgwidth\undefined%
    \setlength{\unitlength}{150.02999496bp}%
    \ifx\svgscale\undefined%
      \relax%
    \else%
      \setlength{\unitlength}{\unitlength * \real{\svgscale}}%
    \fi%
  \else%
    \setlength{\unitlength}{\svgwidth}%
  \fi%
  \global\let\svgwidth\undefined%
  \global\let\svgscale\undefined%
  \makeatother%
  \begin{picture}(1,0.91516703)%
    \lineheight{1}%
    \setlength\tabcolsep{0pt}%
    \put(0,0){\includegraphics[width=\unitlength,page=1]{sideways.pdf}}%
    \put(0.26273662,0.67248883){\color[rgb]{0,0,0}\makebox(0,0)[lt]{\lineheight{1.25}\smash{\begin{tabular}[t]{l}0.00\end{tabular}}}}%
    \put(0,0){\includegraphics[width=\unitlength,page=2]{sideways.pdf}}%
    \put(0.26273662,0.74428354){\color[rgb]{0,0,0}\makebox(0,0)[lt]{\lineheight{1.25}\smash{\begin{tabular}[t]{l}0.50\end{tabular}}}}%
    \put(0,0){\includegraphics[width=\unitlength,page=3]{sideways.pdf}}%
    \put(0.26273662,0.83607427){\color[rgb]{0,0,0}\makebox(0,0)[lt]{\lineheight{1.25}\smash{\begin{tabular}[t]{l}1.00\end{tabular}}}}%
    \put(0.12763864,0.67125097){\color[rgb]{0,0,0}\rotatebox{90}{\makebox(0,0)[lt]{\lineheight{1.25}\smash{\begin{tabular}[t]{l}Success \end{tabular}}}}}%
    \put(0.20222789,0.74087647){\color[rgb]{0,0,0}\rotatebox{90}{\makebox(0,0)[lt]{\lineheight{1.25}\smash{\begin{tabular}[t]{l}rate\end{tabular}}}}}%
    \put(0,0){\includegraphics[width=\unitlength,page=4]{sideways.pdf}}%
    \put(0.26273662,0.40730619){\color[rgb]{0,0,0}\makebox(0,0)[lt]{\lineheight{1.25}\smash{\begin{tabular}[t]{l}0.00\end{tabular}}}}%
    \put(0,0){\includegraphics[width=\unitlength,page=5]{sideways.pdf}}%
    \put(0.26273662,0.48352948){\color[rgb]{0,0,0}\makebox(0,0)[lt]{\lineheight{1.25}\smash{\begin{tabular}[t]{l}0.01\end{tabular}}}}%
    \put(0,0){\includegraphics[width=\unitlength,page=6]{sideways.pdf}}%
    \put(0.26273662,0.5597528){\color[rgb]{0,0,0}\makebox(0,0)[lt]{\lineheight{1.25}\smash{\begin{tabular}[t]{l}0.02\end{tabular}}}}%
    \put(0.12763864,0.36997755){\color[rgb]{0,0,0}\rotatebox{90}{\makebox(0,0)[lt]{\lineheight{1.25}\smash{\begin{tabular}[t]{l}Object \end{tabular}}}}}%
    \put(0.20222789,0.31104844){\color[rgb]{0,0,0}\rotatebox{90}{\makebox(0,0)[lt]{\lineheight{1.25}\smash{\begin{tabular}[t]{l}Moved [m]\end{tabular}}}}}%
    \put(0,0){\includegraphics[width=\unitlength,page=7]{sideways.pdf}}%
    \put(0.51268094,0.15888884){\color[rgb]{0,0,0}\makebox(0,0)[t]{\lineheight{1.25}\smash{\begin{tabular}[t]{c}{\scriptsize Obj 16}\end{tabular}}}}%
    \put(0.68680934,0.15888884){\color[rgb]{0,0,0}\makebox(0,0)[t]{\lineheight{1.25}\smash{\begin{tabular}[t]{c}{\scriptsize Obj 16}\end{tabular}}}}%
    \put(0.68680934,0.09401472){\color[rgb]{0,0,0}\transparent{0.95999998}\makebox(0,0)[t]{\lineheight{1.25}\smash{\begin{tabular}[t]{c}{\scriptsize Drop}\end{tabular}}}}%
    \put(0.68680934,0.02886042){\color[rgb]{0,0,0}\transparent{0.95999998}\makebox(0,0)[t]{\lineheight{1.25}\smash{\begin{tabular}[t]{c}{\scriptsize \SI{20}{\g}}\end{tabular}}}}%
    \put(0.86093769,0.09401472){\color[rgb]{0,0,0}\transparent{0.95999998}\makebox(0,0)[t]{\lineheight{1.25}\smash{\begin{tabular}[t]{c}{\scriptsize Drop}\end{tabular}}}}%
    \put(0.86093769,0.02886042){\color[rgb]{0,0,0}\transparent{0.95999998}\makebox(0,0)[t]{\lineheight{1.25}\smash{\begin{tabular}[t]{c}{\scriptsize \SI{100}{\g}}\end{tabular}}}}%
    \put(0.86093766,0.15888884){\color[rgb]{0,0,0}\makebox(0,0)[t]{\lineheight{1.25}\smash{\begin{tabular}[t]{c}{\scriptsize Obj 16}\end{tabular}}}}%
    \put(0,0){\includegraphics[width=\unitlength,page=8]{sideways.pdf}}%
  \end{picture}%
\endgroup%

%% file: pictures/results/slip_catch/ood/different_evetac.pdf_tex
\begingroup%
  \makeatletter%
  \providecommand\color[2][]{%
    \errmessage{(Inkscape) Color is used for the text in Inkscape, but the package 'color.sty' is not loaded}%
    \renewcommand\color[2][]{}%
  }%
  \providecommand\transparent[1]{%
    \errmessage{(Inkscape) Transparency is used (non-zero) for the text in Inkscape, but the package 'transparent.sty' is not loaded}%
    \renewcommand\transparent[1]{}%
  }%
  \providecommand\rotatebox[2]{#2}%
  \newcommand*\fsize{\dimexpr\f@size pt\relax}%
  \newcommand*\lineheight[1]{\fontsize{\fsize}{#1\fsize}\selectfont}%
  \ifx\svgwidth\undefined%
    \setlength{\unitlength}{352.5bp}%
    \ifx\svgscale\undefined%
      \relax%
    \else%
      \setlength{\unitlength}{\unitlength * \real{\svgscale}}%
    \fi%
  \else%
    \setlength{\unitlength}{\svgwidth}%
  \fi%
  \global\let\svgwidth\undefined%
  \global\let\svgscale\undefined%
  \makeatother%
  \begin{picture}(1,0.38951065)%
    \lineheight{1}%
    \setlength\tabcolsep{0pt}%
    \put(0.24873629,0.06762579){\color[rgb]{0,0,0}\transparent{0.95999998}\makebox(0,0)[t]{\lineheight{1.25}\smash{\begin{tabular}[t]{c}{\scriptsize Obj 17}\end{tabular}}}}%
    \put(0.34230993,0.03685852){\color[rgb]{0,0,0}\transparent{0.95999998}\makebox(0,0)[t]{\lineheight{1.25}\smash{\begin{tabular}[t]{c}{\scriptsize Closed}\end{tabular}}}}%
    \put(0.34230993,0.00912773){\color[rgb]{0,0,0}\transparent{0.95999998}\makebox(0,0)[t]{\lineheight{1.25}\smash{\begin{tabular}[t]{c}{\scriptsize Gel}\end{tabular}}}}%
    \put(0.3423099,0.06762578){\color[rgb]{0,0,0}\makebox(0,0)[t]{\lineheight{1.25}\smash{\begin{tabular}[t]{c}{\scriptsize Obj 17}\end{tabular}}}}%
    \put(0.43588357,0.06762579){\color[rgb]{0,0,0}\transparent{0.95999998}\makebox(0,0)[t]{\lineheight{1.25}\smash{\begin{tabular}[t]{c}{\scriptsize Obj 16}\end{tabular}}}}%
    \put(0.52945714,0.03685852){\color[rgb]{0,0,0}\transparent{0.95999998}\makebox(0,0)[t]{\lineheight{1.25}\smash{\begin{tabular}[t]{c}{\scriptsize Closed}\end{tabular}}}}%
    \put(0.52945714,0.00912774){\color[rgb]{0,0,0}\transparent{0.95999998}\makebox(0,0)[t]{\lineheight{1.25}\smash{\begin{tabular}[t]{c}{\scriptsize Gel}\end{tabular}}}}%
    \put(0.52945711,0.06762579){\color[rgb]{0,0,0}\makebox(0,0)[t]{\lineheight{1.25}\smash{\begin{tabular}[t]{c}{\scriptsize Obj 16}\end{tabular}}}}%
    \put(0.71660438,0.03685852){\color[rgb]{0,0,0}\transparent{0.95999998}\makebox(0,0)[t]{\lineheight{1.25}\smash{\begin{tabular}[t]{c}{\scriptsize Closed}\end{tabular}}}}%
    \put(0.71660438,0.00912773){\color[rgb]{0,0,0}\transparent{0.95999998}\makebox(0,0)[t]{\lineheight{1.25}\smash{\begin{tabular}[t]{c}{\scriptsize Gel}\end{tabular}}}}%
    \put(0.71660436,0.06762578){\color[rgb]{0,0,0}\makebox(0,0)[t]{\lineheight{1.25}\smash{\begin{tabular}[t]{c}{\scriptsize Obj 19}\end{tabular}}}}%
    \put(0.6230308,0.06762578){\color[rgb]{0,0,0}\transparent{0.95999998}\makebox(0,0)[t]{\lineheight{1.25}\smash{\begin{tabular}[t]{c}{\scriptsize Obj 19}\end{tabular}}}}%
    \put(0.90375164,0.03685852){\color[rgb]{0,0,0}\transparent{0.95999998}\makebox(0,0)[t]{\lineheight{1.25}\smash{\begin{tabular}[t]{c}{\scriptsize Closed}\end{tabular}}}}%
    \put(0.90375164,0.00912774){\color[rgb]{0,0,0}\transparent{0.95999998}\makebox(0,0)[t]{\lineheight{1.25}\smash{\begin{tabular}[t]{c}{\scriptsize Gel}\end{tabular}}}}%
    \put(0.90375162,0.06762579){\color[rgb]{0,0,0}\makebox(0,0)[t]{\lineheight{1.25}\smash{\begin{tabular}[t]{c}{\scriptsize Obj 20}\end{tabular}}}}%
    \put(0.810178,0.06762579){\color[rgb]{0,0,0}\transparent{0.95999998}\makebox(0,0)[t]{\lineheight{1.25}\smash{\begin{tabular}[t]{c}{\scriptsize Obj 20}\end{tabular}}}}%
    \put(0,0){\includegraphics[width=\unitlength,page=1]{different_evetac.pdf}}%
    \put(0.11059383,0.28939715){\color[rgb]{0,0,0}\makebox(0,0)[lt]{\lineheight{1.25}\smash{\begin{tabular}[t]{l}0.00\end{tabular}}}}%
    \put(0,0){\includegraphics[width=\unitlength,page=2]{different_evetac.pdf}}%
    \put(0.11059383,0.31844256){\color[rgb]{0,0,0}\makebox(0,0)[lt]{\lineheight{1.25}\smash{\begin{tabular}[t]{l}0.50\end{tabular}}}}%
    \put(0,0){\includegraphics[width=\unitlength,page=3]{different_evetac.pdf}}%
    \put(0.11059383,0.35599861){\color[rgb]{0,0,0}\makebox(0,0)[lt]{\lineheight{1.25}\smash{\begin{tabular}[t]{l}1.00\end{tabular}}}}%
    \put(0.05309383,0.28728307){\color[rgb]{0,0,0}\rotatebox{90}{\makebox(0,0)[lt]{\lineheight{1.25}\smash{\begin{tabular}[t]{l}Success\end{tabular}}}}}%
    \put(0.08484028,0.31691687){\color[rgb]{0,0,0}\rotatebox{90}{\makebox(0,0)[lt]{\lineheight{1.25}\smash{\begin{tabular}[t]{l}rate\end{tabular}}}}}%
    \put(0,0){\includegraphics[width=\unitlength,page=4]{different_evetac.pdf}}%
    \put(0.35026879,0.37399489){\color[rgb]{0,0,0}\makebox(0,0)[lt]{\lineheight{1.25}\smash{\begin{tabular}[t]{l}Overall\end{tabular}}}}%
    \put(0,0){\includegraphics[width=\unitlength,page=5]{different_evetac.pdf}}%
    \put(0.58785743,0.37399489){\color[rgb]{0,0,0}\makebox(0,0)[lt]{\lineheight{1.25}\smash{\begin{tabular}[t]{l}Lift\end{tabular}}}}%
    \put(0,0){\includegraphics[width=\unitlength,page=6]{different_evetac.pdf}}%
    \put(0.76848685,0.37399489){\color[rgb]{0,0,0}\makebox(0,0)[lt]{\lineheight{1.25}\smash{\begin{tabular}[t]{l}Balance\end{tabular}}}}%
    \put(0,0){\includegraphics[width=\unitlength,page=7]{different_evetac.pdf}}%
    \put(0.11059383,0.17755702){\color[rgb]{0,0,0}\makebox(0,0)[lt]{\lineheight{1.25}\smash{\begin{tabular}[t]{l}0.00\end{tabular}}}}%
    \put(0,0){\includegraphics[width=\unitlength,page=8]{different_evetac.pdf}}%
    \put(0.11059383,0.21489058){\color[rgb]{0,0,0}\makebox(0,0)[lt]{\lineheight{1.25}\smash{\begin{tabular}[t]{l}0.01\end{tabular}}}}%
    \put(0,0){\includegraphics[width=\unitlength,page=9]{different_evetac.pdf}}%
    \put(0.11059383,0.25222415){\color[rgb]{0,0,0}\makebox(0,0)[lt]{\lineheight{1.25}\smash{\begin{tabular}[t]{l}0.02\end{tabular}}}}%
    \put(0.05309383,0.15588153){\color[rgb]{0,0,0}\rotatebox{90}{\makebox(0,0)[lt]{\lineheight{1.25}\smash{\begin{tabular}[t]{l}Object\end{tabular}}}}}%
    \put(0.08484028,0.13080029){\color[rgb]{0,0,0}\rotatebox{90}{\makebox(0,0)[lt]{\lineheight{1.25}\smash{\begin{tabular}[t]{l}Moved [m]\end{tabular}}}}}%
    \put(0,0){\includegraphics[width=\unitlength,page=10]{different_evetac.pdf}}%
  \end{picture}%
\endgroup%